%% file: realtabformer.tex
\documentclass[nohyperref,dvipsnames]{article}

\usepackage{microtype}
\usepackage{graphicx}
\usepackage{subfigure}
\usepackage{booktabs} 

\usepackage{hyperref}

\newcommand{\m}{\textup{\texttt{-}}}
\usepackage{multirow}
\usepackage{mdframed}

\usepackage[accepted]{icml2023}

\usepackage[frozencache=true,cachedir=minted-cache]{minted} 
\definecolor{LightGray}{gray}{0.9}

\usepackage{amsmath}
\usepackage{amssymb}
\usepackage{mathtools}
\usepackage{amsthm}

\usepackage[capitalize,noabbrev]{cleveref}

\theoremstyle{plain}

\theoremstyle{definition}

\theoremstyle{remark}

\usepackage[textsize=tiny]{todonotes}

\icmltitlerunning{REaLTabFormer: Generating Realistic Relational and Tabular Data using Transformers}

\begin{document}

\twocolumn[
\icmltitle{REaLTabFormer: Generating Realistic Relational \\ and Tabular Data using Transformers}



\icmlsetsymbol{equal}{*}

\begin{icmlauthorlist}
\icmlauthor{Aivin V.  Solatorio}{wb_org}
\icmlauthor{Olivier Dupriez}{wb_org}
\end{icmlauthorlist}


\icmlaffiliation{wb_org}{Development Economics Data Group, The World Bank, USA}
\icmlcorrespondingauthor{\\ Aivin V. Solatorio}{asolatorio@worldbank.org}{avsolatorio}

\icmlkeywords{REaLTabFormer, Tabular Data, Generative Models, Transformer, Deep Learning, Probabilistic Models, Machine Learning}

\vskip 0.3in
]



\printAffiliationsAndNotice{}  

\begin{abstract}

Tabular data is a common form of organizing data. Multiple models are available to generate synthetic tabular datasets where observations are independent, but few have the ability to produce relational datasets. Modeling relational data is challenging as it requires modeling both a “parent” table and its relationships across tables. We introduce REaLTabFormer (Realistic Relational and Tabular Transformer), a tabular and relational synthetic data generation model. It first creates a parent table using an autoregressive GPT-2 model, then generates the relational dataset conditioned on the parent table using a sequence-to-sequence (Seq2Seq) model. We implement target masking to prevent data copying and propose the $Q_{\delta}$ statistic and statistical bootstrapping to detect overfitting. Experiments using real-world datasets show that REaLTabFormer captures the relational structure better than a baseline model. REaLTabFormer also achieves state-of-the-art results on prediction tasks, “out-of-the-box”, for large non-relational datasets without needing fine-tuning.

\end{abstract}

\section{Introduction}

\begin{figure}[t]
\vskip 0.0in
\begin{center}

\centerline{\includegraphics[width=\columnwidth] {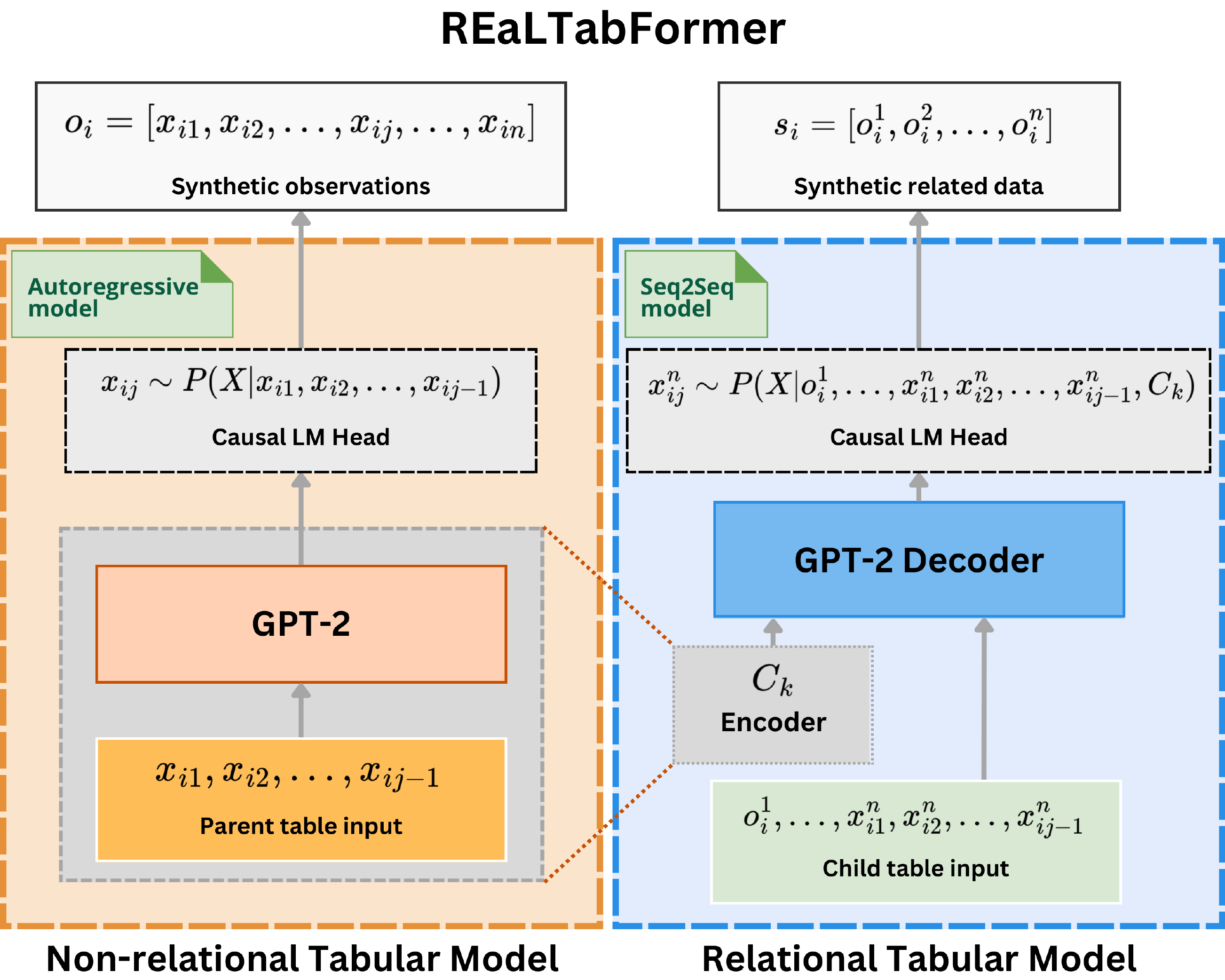}}
\caption{
Illustration of the REaLTabFormer model. The left block shows the non-relational tabular data model using GPT-2 with a causal LM head. In contrast, the right block shows how a relational dataset's child table is modeled using a sequence-to-sequence (Seq2Seq) model. The Seq2Seq model uses the observations in the parent table to condition the generation of the observations in the child table. The trained GPT-2 model on the parent table, with weights frozen, is also used as the encoder in the Seq2Seq model.}
\label{fig:realtabformer-model}
\end{center}
\vskip -0.3in
\end{figure}

Tabular data is one of the most common forms of data. Many datasets from surveys, censuses, and administrative sources are provided in this form. These datasets may contain sensitive information that cannot be shared openly \citep{abdelhameed2018privacy}. Even when statistical disclosure methods are applied, they may remain vulnerable to malicious attacks \citep{cheng2017enterprise}. As a result, their dissemination is restricted and the data have limited utility \citep{o2015individual}. Differential privacy methods \citep{ji2014differential}, homomorphic encryption approaches \citep{aslett2015review,wood2020homomorphic}, or federated machine learning \citep{yang2019federated,lin2021privacy} may be implemented, allowing insights from sensitive data to be accessible to researchers. Synthetic tabular data with similar statistical properties as the real data offer an alternative, offering more value especially in granular and segmentation analyses. To comply with data privacy requirements, the generative models that produce these synthetic data must provide guarantees that “data copying” does not happen \citep{meehan2020non,carlini2023extracting}.

Formally, tabular data is a collection of observations (rows) $o_i$ that may or may not be independent. A single observation in a tabular data $\mathbf{T}$ with $n$ columns is defined by $o_i = [x_{i1}, x_{i2}, ..., x_{ij}, ..., x_{in}]$, and $j$ indicating the $j^{th}$ column. We refer to tabular data having observations independent of each other as \textbf{non-relational tabular data}. Tabular data having observations related to each other are referred to as \textbf{relational tabular data}. Relational datasets have at least one pair of tabular data files with a one-to-many mapping of observations between the parent table and the child table, respectively, linked by a unique identifier. In the context of a relational dataset, a \textbf{parent table} is a non-relational tabular data, whereas the \textbf{child table} is a relational tabular data. Relational tabular databases model the logical partitioning of data and prevent unnecessary duplication of observations from the parent to child tables \citep{jatana2012survey}. Despite its ubiquity, limited work has been done in generating synthetic relational datasets. This may be due to the challenging nature of modeling the complex relationships within and across tables.

The field of synthetic data generation has seen significant development in recent years \citep{gupta2016synthetic,abufadda2021survey,hernandez2022synthetic,figueira2022survey}. Generative models have become mainstream with the advent of synthetic image generation models such as DALL-E \citep{ramesh2021zero}, and most recently, ChatGPT. While generative models for images and text are common, models for producing synthetic tabular data are comparatively limited despite their multiple possible applications. Synthetic tabular data can contribute to addressing data privacy issues and data sparseness \citep{appenzeller2022privacy}. They can help to make sensitive data accessible to researchers \citep{goncalves2020generation}, and to fill gaps in data availability for counterfactual research and agent-based simulations \citep{fagiolo2019validation}, and for synthetic control methods \citep{abadie2015comparative}. Further value can be derived from tabular data by building predictive models using machine learning \citep{shwartz2022tabular}. These predictive models can infer variables of interest in the data that may otherwise be expensive to collect or correspond to some success metrics that can guide business decisions. Synthetic data produced by deep learning models have been shown to perform well in predictive modeling tasks. This extends the utility of real-world data that may otherwise be unused due to privacy concerns.

This paper introduces the \textbf{REalTabFormer}, a transformer-based framework for generating non-relational tabular data and relational datasets. It makes the following contributions:

\vspace{-5pt}
\paragraph{Unified framework}

The REalTabFormer uses an autoregressive (GPT-2) transformer to model non-relational tabular data for modeling and generating parent tables. It then models and generates observations in the child table using the sequence-to-sequence (Seq2Seq) \citep{yun2019transformers} framework. The encoder network uses the pre-trained weights of the network for the parent table, contextualizing the input for generating arbitrary-length data corresponding to observations in a child table, via the decoder network.

\vspace{-5pt}
\paragraph{Strategies for privacy-preserving training}

Synthetic data generation models must not only be able to generate realistic data but also implement safeguards to prevent the model from ``memorizing" and copying observations in the training data during sampling. We use the \textit{distance to closest record} (DCR), a data-copying measure, and \textit{statistical bootstrapping} to detect overfitting during training robustly. We introduce \textit{target masking} for regularization to reduce the likelihood of training data being replicated by the model.

\vspace{-5pt}
\paragraph{Open-sourced models}

We publish the REaLTabFormer models as an open-sourced Python package. Install the package using: \hspace{0.1cm} \texttt{pip install realtabformer}.\footnote{\href{https://github.com/avsolatorio/REaLTabFormer}{\url{https://github.com/avsolatorio/REaLTabFormer}}}

\vspace{-5pt}
\paragraph{Comprehensive evaluation}

We evaluate the performance of our models on a variety of real-world datasets. We use open-sourced state-of-the-art models as baselines to assess the performance of REaLTabFormer in generating non-relational and relational tabular datasets.

Our experiments demonstrate the effectiveness of the REaLTabFormer model for non-relational tabular data, beating current state-of-the-art in machine learning tasks for large datasets. We further demonstrate that the synthesized observations for the child table generated by the REaLTabFormer capture relational statistics more accurately than the baseline models.

\vspace{-2pt}
\section{Related Work}
\label{relatedwork}

Recent advances in deep learning, such as generative adversarial networks \citep{park2018data,xu_modeling_2019,zhao2021ctab}, autoencoders \citep{li2019evaluating,xu_modeling_2019,darabi2021synthesising}, language models \citep{borisov_language_2022}, and diffusion models \citep{kotelnikov2022tabddpm} have been applied to synthetic non-relational tabular data generation.  These papers demonstrate deep learning models' capacity to produce more realistic data than traditional approaches such as Bayesian networks \citep{xu_modeling_2019}.

On the other hand, generative models for relational datasets are limited  \citep{patki_synthetic_2016,gueye2022row}. Existing models are based on Hierarchical Modeling Algorithms \citep{patki_synthetic_2016} where traditional statistical models, Gaussian Copulas, are used to learn the joint distributions for each and across tables. While these models can synthesize data, the quality of the generated data does not accurately capture the nuanced conditions within and across tables (Fig.~\ref{fig:rossmann-date-sales} and Fig.~\ref{fig:airbnb-device_type-age_group}).

\citet{padhi_tabular_2021} presented TabGPT for generating synthetic transactional data. They showed that autoregressive transformers, particularly GPT, can synthesize arbitrary-length data. One limitation of TabGPT is that one has to train independent models to produce transactions for each user. This becomes impractical for real-world applications. Our work generalizes the use of GPT by proposing a sequence-to-sequence framework for generating arbitrary-length synthetic data conditioned on an input.

\section{REaLTabFormer}
\label{realtabformer}

REaLTabFormer is a transformer-based framework for generating non-relational tabular data using an autoregressive model and relational tabular data using a sequence-to-sequence (Seq2Seq) architecture. The framework also consists of strategies for encoding tabular data (Section \ref{tabdataencoding}), a statistical method to detect overfitting (Section \ref{overfitting}), and a constrained sampling strategy during generation.

Details of the framework are described in this section. First, we present our proposed models to synthesize realistic relational datasets. Next, we outline and describe the data processing applied to the tabular data as input for training the models. We then discuss solutions to improve our model's training and sampling process.

\subsection{The REaLTabFormer Models}

\paragraph{Parent table model}
To generate synthetic observations for a non-relational tabular data $\mathbf{T}$, we model the conditional distribution of columnar values in each row of the data. Consider a single observation $o_i = [x_{i1}, x_{i2}, ..., x_{ij}, ..., x_{in}]$ in $\mathbf{T}$ as defined earlier. We treat $o_i$ as a sequence with potential dependencies across values $x_{ij}$, similar to a sentence in a text. This re-framing provides us with a framework to learn the conditional distribution $x_{ij} \sim P(X|x_{i1}, x_{i2}, ..., x_{ij-1})$ and sequentially generate the next values in the sequence, eventually generating the full observation \citep{jelinek1985markov,bengio2000neural}. We use an autoregressive model to learn this distribution, Fig.~\ref{fig:realtabformer-model}. In the context of relational datasets, we use this approach to generate synthetic observations for the parent table $\mathbf{T_0}$. We extend this formulation to generate the child table $\mathbf{T'}$---a relational tabular data---associated with the parent table $\mathbf{T_0}$.

\vspace{-5pt}
\paragraph{Child table model}

The extension is established by introducing a context learned by an encoder network from observations in $\mathbf{T_0}$. Instead of generating $o_i$ in $\mathbf{T'}$ independently, we concatenate the child table observations related to the same observation in $\mathbf{T_0}$. This forms an arbitrary-length sequence $s_i = [o_i^1, o_i^2, ..., o_i^n]$, where $n$ is the number of related observations in $\mathbf{T'}$.

We propose to model the generation of $s_i$ given an observation in $\mathbf{T_0}$ as $x_{ij}^n \sim P(X|o_i^1, ...,x_{i1}^n, x_{i2}^n, ..., x_{ij-1}^n, C_k)$, where $C_k$ is a context captured from a related observation in the parent tabular data $\mathbf{T_0}$. We also use the same network trained on the parent table, \textbf{with weights frozen}, as the Seq2Seq model's encoder. This choice is expected to speed up the training process since only the cross-attention layer and the decoder network are needed to be trained for the child table model. The encoder network is assumed to have learned the properties of the parent table and will transfer this information to the decoder without further fine-tuning its weights, Fig~\ref{fig:realtabformer-model}. 

\begin{figure}[t]
\begin{center}
\centerline{\includegraphics[width=\columnwidth]{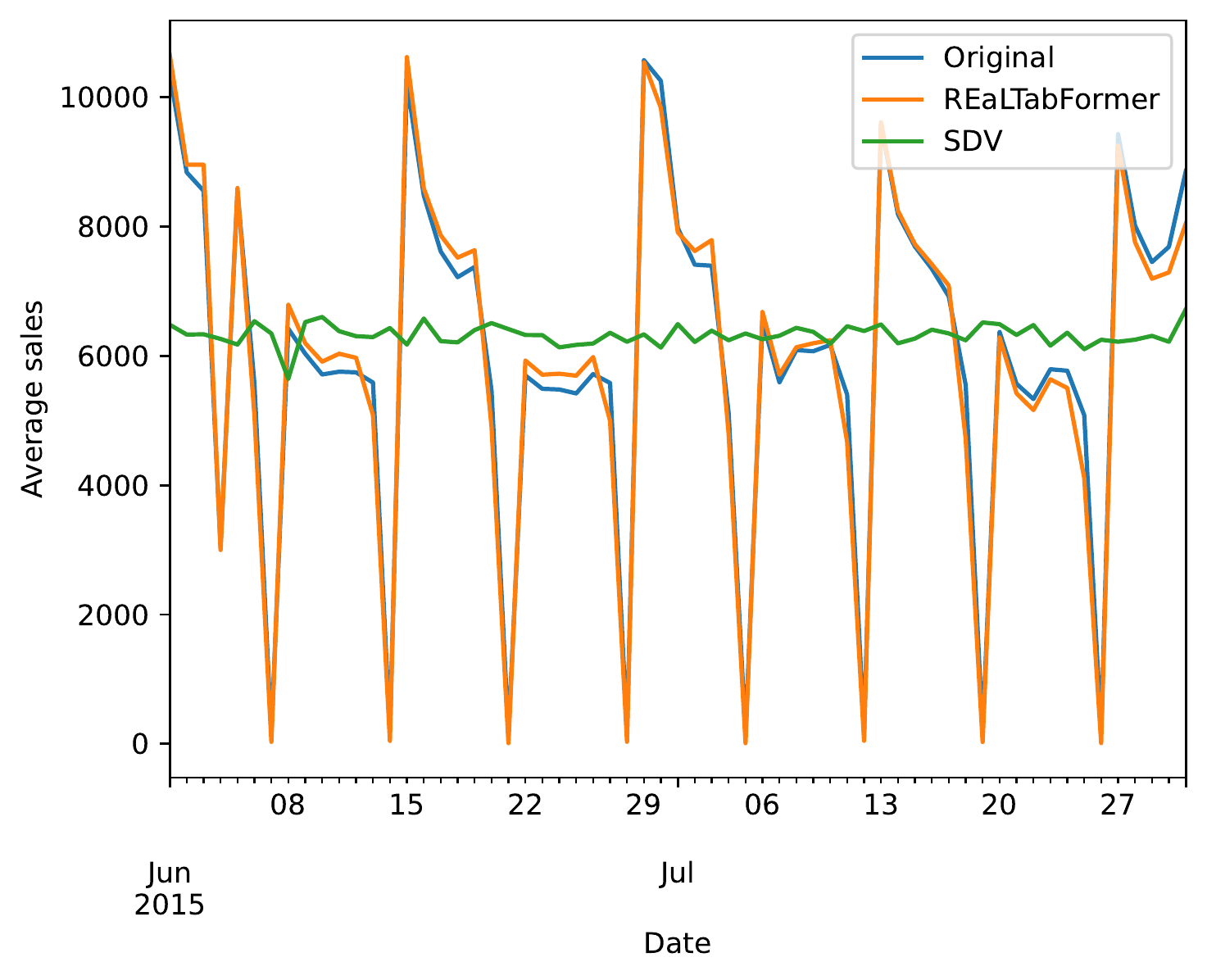}}
\caption{Graph of the daily mean of the \texttt{Sales} variable computed from the original Rossmann dataset (blue), synthetic data produced by REaLTabFormer (orange), and data generated by SDV (green). The REaLTabFormer closely captures the seasonality in the data compared with the HMA model from the SDV.}
\vskip -0.4in
\label{fig:rossmann-date-sales}
\end{center}
\end{figure}

\paragraph{GPT-2: an autoregressive transformer}

Previous works have shown that transformer-based autoregressive models can capture the conditional distribution of sequential data very well \citep{radford_language_2019,padhi_tabular_2021}. REaLTabFormer uses the GPT-2 architecture---a transformer-decoder architecture designed for autoregressive tasks---as its base model. We adopt the same architecture for all GPT-2 instances in the framework for simplicity. The GPT-2 architecture used in the REaLTabFormer has 768-dimensional embeddings, 6 decoder layers, and 12 attention heads---a set of parameters similar to DistilGPT2. We use the implementation from the HuggingFace transformers library \citep{wolf_transformers_2020}.

\newcommand{\rulesep}{\unskip\ \vrule\ }

\begin{figure*}
    \centering
    \includegraphics[width=0.38\textwidth]{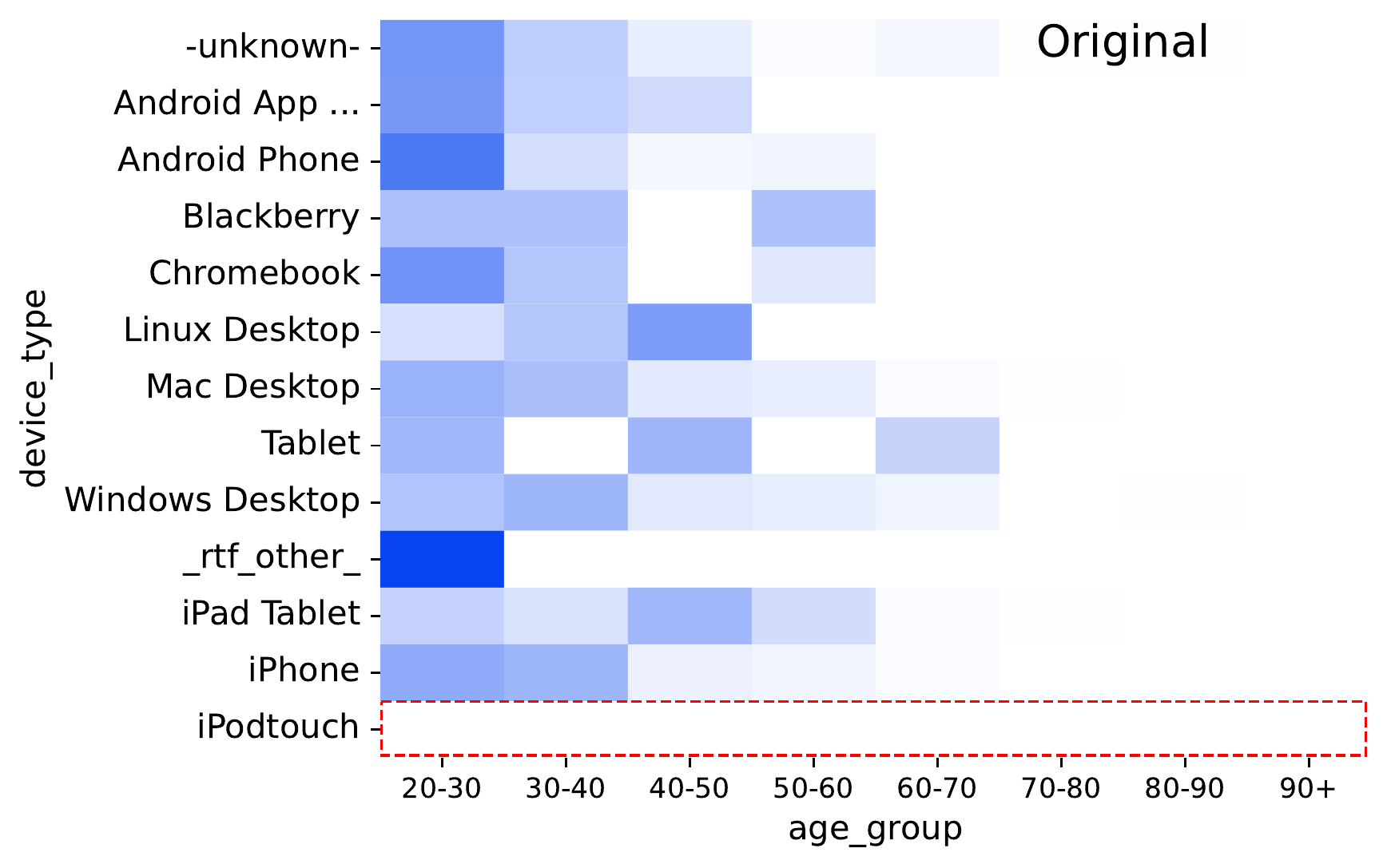}
    \includegraphics[width=0.29\textwidth]{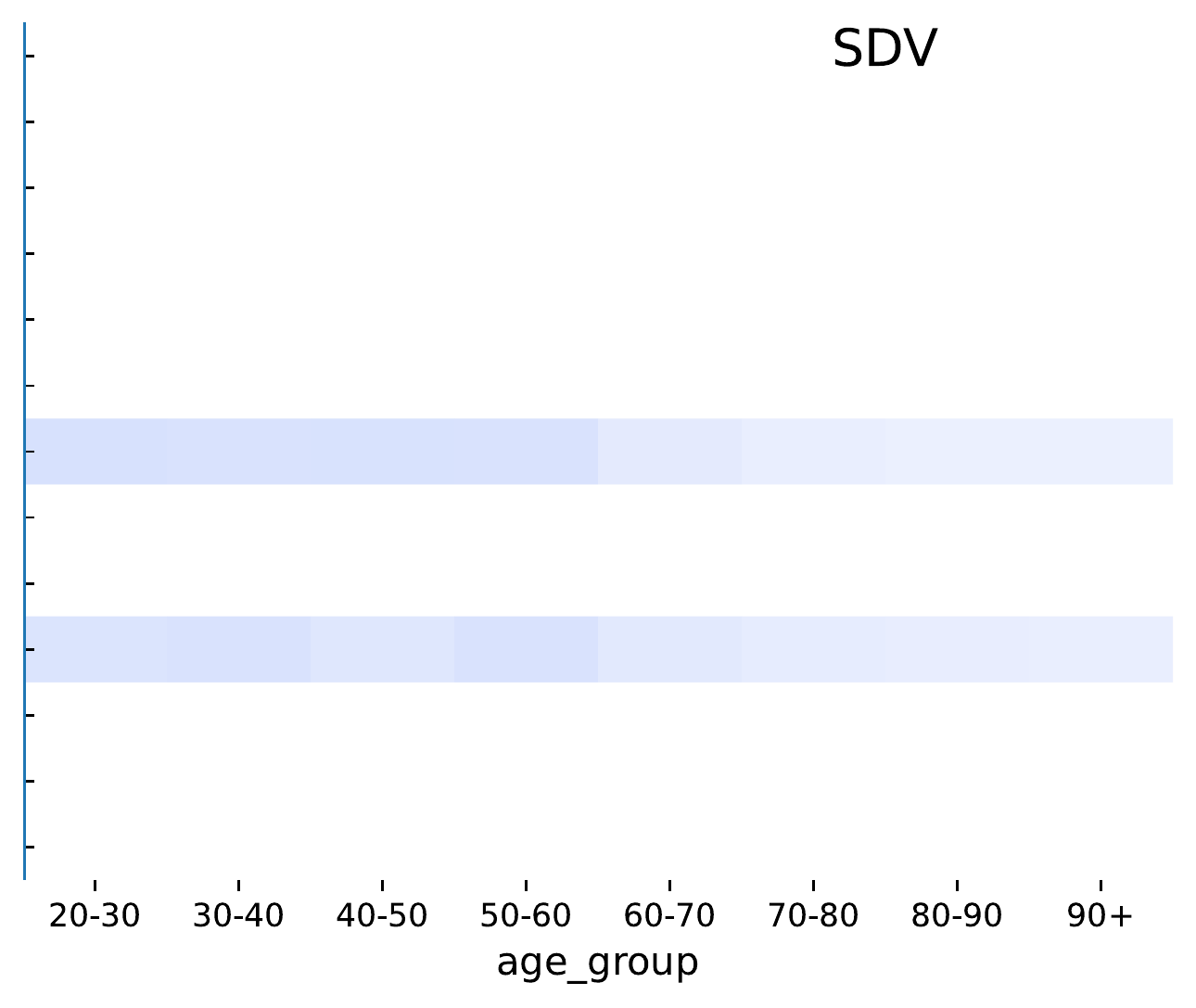}
    \includegraphics[width=0.27\textwidth]{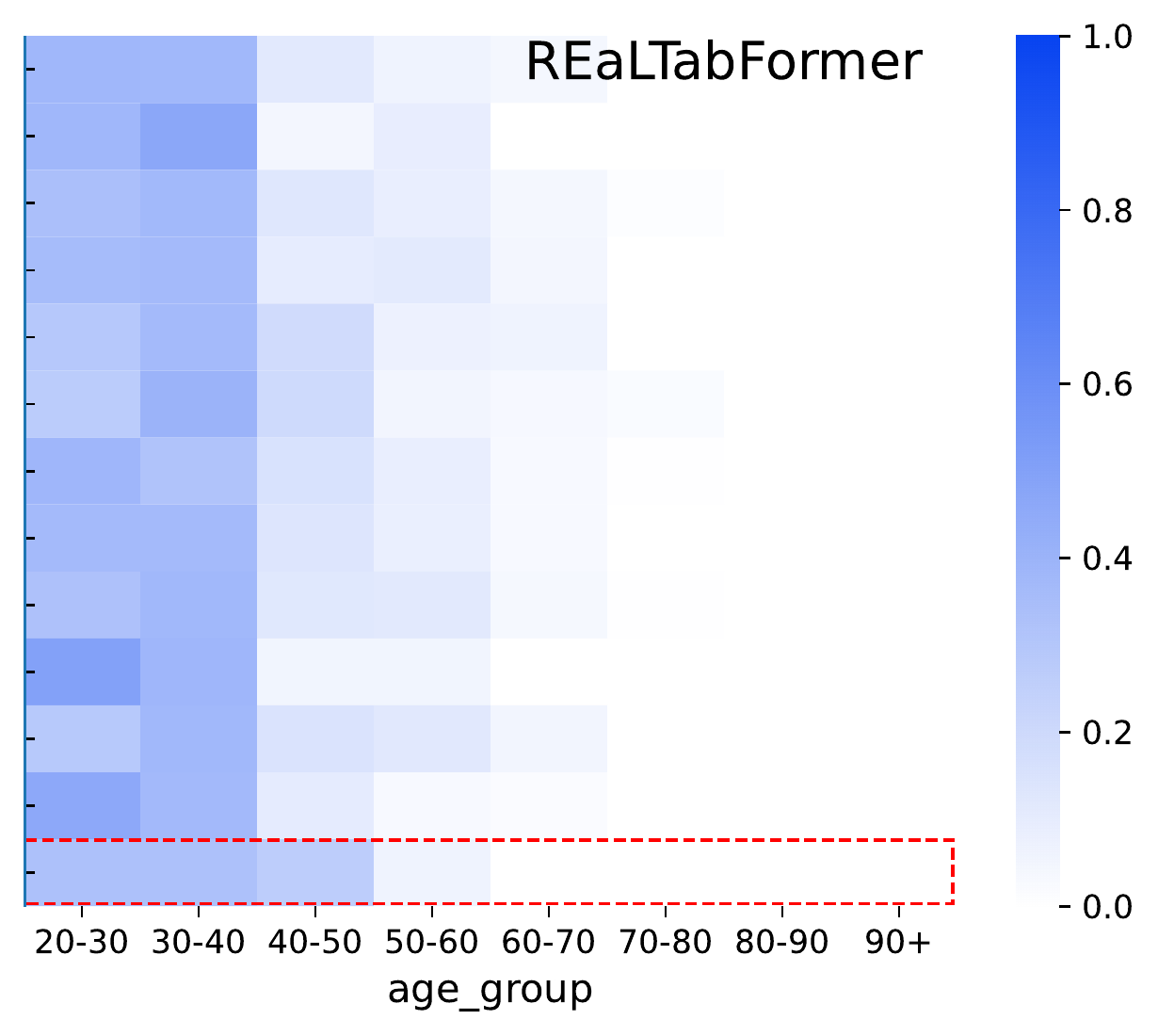}
    \vskip 0.05in
    \caption{Joint distributions of the \texttt{age\_group} variable in the parent table and the \texttt{device\_type} in the child table of the Airbnb test dataset (left), the SDV (middle), and the REalTabFormer (right). The plots show that the REaLTabFormer can synthesize values across the domain of the variables, while SDV learned only two device types out of thirteen. The REaLTabFormer also generalized and imputed age values for users with ``iPodtouch" device (red box). This device type group has missing values for age in the original data.}
    \label{fig:airbnb-device_type-age_group}
\end{figure*}

\vspace{-2pt}
\subsection{Tabular Data Encoding}
\label{tabdataencoding}

The GReaT model that uses pretrained large language models (LLMs) proposed by \citet{borisov_language_2022} offers insight into the minimal data processing requirements for language models in generating tabular data. There is, however, the potential for optimization in using autoregressive language models for this task, as the fine-tuning process of a large pretrained model incurs computational costs. Particularly, LLMs are trained on a large vocabulary where most of the tokens are not needed for generating the tabular data at hand. These unnecessary tokens increase the model's computational requirements and prolong training and sampling times. To improve the efficiency of our model, we adopt a fixed-set vocabulary as initially proposed by \citet{padhi_tabular_2021}. Generating a fixed vocabulary for each column in the tabular data offers various advantages in training performance and sampling. One of the main advantages is being able to filter irrelevant tokens when generating values for a specific column. This directly contributes to efficiency in sampling by reducing the chances of generating invalid samples. Our model performs minimal transformation of the raw data. First, we identify the various data types for each column in the data. We then perform a series of data processing specific to the column and data type. Notably, we adopt a fully text-based strategy in handling numerical values. These transformations produce a transformed tabular data used to train the model. \citet{borisov_language_2022} showed that variable order has an insignificant impact on language models, so we did not apply variable permutation. We discuss the processing for each data type in Appendix~\ref{appendix:raw-data-processing}.

\vspace{-4pt}
\paragraph{Training data for the parent table}

The GPT-2 model we use requires a set of token ids as input. To generate these sequences of token ids, we first create a vocabulary. This vocabulary maps the unique tokens in each column to a unique token id. Then, for each row in the modified data, we apply the mapping in the vocabulary to the tokens. This produces a list of token ids for each row of the data. The model is then trained on an autoregressive task wherein the target data corresponds to the right-shifted tokens of the input data.

\vspace{-4pt}
\paragraph{Training data for the child table}

We concatenate the transformed observations corresponding to related rows in the child table. A special token is added before and after the set of tokens representing an individual observation. In this form, the data we use to train the Seq2Seq model contains input-output pairs. An input value contains a fixed-length array of token ids representing the observation in the parent table. The input is similar to the input used in the parent table model. An arbitrary-length array with the concatenated token ids for each related observation in the child table represents the output value. The number of related observations that can be modeled is limited by computational resources.

\subsection{REaLTabFormer Training and Sampling}

Deep learning models for generative tasks face challenges of overfitting the data resulting in issues such as data-copying \citep{meehan2020non,carlini2023extracting}. Furthermore, observations generated by generative models for tabular data could face issues of validity and inconsistency. These issues in the generated samples impact the efficiency of the generative process. 
Our proposed framework addresses the aforementioned issues by, (i) introducing a robust statistical method to monitor overfitting, and (ii) target masking to further reduce the risk of data copying. To improve the rate of producing valid observations by the model, we also implement a constrained generation strategy during the sampling stage.

\subsubsection{Target masking}
\label{targetmasking}
Data copying is a critical issue for deep learning-based generative tabular models as it can expose and compromise sensitive information in the training data. To mitigate data-copying, we introduce \textbf{target masking}. Target masking is a form of regularization aimed at minimizing the likelihood of records in the training data being ``memorized" and copied by the generative model. Unlike the token masking introduced in BERT \citep{devlin2018bert}, where input tokens are masked and the model is expected to predict the correct token, target masking implements random replacement of the target or label tokens with a special mask token. This artificially introduces missing values in the data.

We intend for the model to learn the masks instead of the actual values. During the sampling stage, we then restrict the mask token generation, forcing the model to fill the value with a valid token probabilistically. Notably, even when the model learns to copy the input-output pair, the learned output corresponds to the masked version of the input. Therefore, when we process the output, the probabilistic nature of replacing the mask token reduces the likelihood of generating the training data. The mask rate parameter controls the proportion of the tokens that are masked. We use a mask rate of $10\%$ in our experiments. 

\subsubsection{Overfitting assessment}
\label{overfitting}

Applying deep learning models to small datasets may easily result in overfitting. This may cause  privacy-related issues when the model generates observations that are copied from the training data. Knowing when the model overfits the data is also crucial when the purpose is to generate diverse out-of-sample data. An overfitted model tends to generate samples generally closer to the training data, thereby limiting the generalization capacity of the model. While the former issue can be resolved by post-generation filtering, the latter must be detected during the model training.

Taking hold-out data to detect overfitting is a common strategy in machine learning. Unfortunately, this strategy could result in the premature termination of model training. It may also penalize a model based only on a small subset of the data \citep{blum1999beating}. The training procedures of existing state-of-the-art models do not explicitly check for overfitting \citep{xu_modeling_2019,borisov_language_2022,kotelnikov2022tabddpm}. We propose and describe below an empirical statistical method to inform the generative model when overfitting happens. The method allows for the full data to be used in the training without the need for a hold-out set. The design of the method is expected to also help prevent data copying and the production of data that is riskily close to the training data.

\vspace{-2pt}
\paragraph{Distance to closest record}

We use the distance to the closest record (DCR) \citep{park2018data} to measure the similarity of synthetic samples to the original data. The DCR is evaluated by taking a specified distance metric $L$ between the training data $T_r$ and the generated data $G$. We then find the smallest distance for each record. Consider the distance matrix between $T_r$ and $G$,

\vspace{-5pt}
\begin{equation}
    \mathbf{D} = L(T_r, G)
\vspace{-3pt}
\end{equation}

The minimum value in each row $i$ of $\mathbf{D}$ is the minimum distance of the $i^{th}$ record in the training data with respect to all records in the generated data. We denote this set of minimum values as $\vec{d_i}$. The minimum value in each column $j$ of $\mathbf{D}$ is the minimum distance of the $j^{th}$ record in the generated data with respect to all records in the training data. We denote this set of minimum values as $\vec{d_j}$. We then take $\vec{d_g} = [\vec{d_i}, \vec{d_j}]$ as the distribution of distances to closest records between the training data and the generated data. We define the quantity

\vspace{-5pt}
\begin{equation} \label{eq:dcr}
    \vec{d} = [\vec{d_i}, \vec{d_j}]
\vspace{-3pt}
\end{equation}

as the distance to closest record distribution for some $T_r$ and some arbitrary sample. We also derive the DCR between the train dataset and some hold-out data $T_h$. Let us denote this distribution of distances as $\vec{d_h}$. We use the distributions $\vec{d_g}$ and $\vec{d_h}$ in our proposed non-parametric method.

\input{tables/results_ml_efficiency.tex}

\vspace{-2pt}
\paragraph{Quantile difference ($\mathbf{Q_\delta}$) statistic}

Two samples from a similar distribution should, on average, have approximately the same values at each quantile of the distribution. To detect whether two samples come from different distributions, we define a set of quantiles over which we compare the two samples. For each quantile in the set, we find the value at the given quantile in one sample and measure the proportion of the values in the other sample that are below it. If the distributions are similar, the proportion should be close to the given quantile, for all quantiles being tested.

Formally, let $S_h$ and $S_g$ having $m$ and $n$ observations, respectively, be two samples being compared. Let $Q$ be a set of $N$ quantiles, and $q \in Q$ is a specific quantile in the set. Consider $v_q$ as the value in $S_h$ at quantile $q$. Then, we compute the value

\vspace{-5pt}
\begin{equation} \label{eq:pq}
    p_q = \frac{\sum_{x \in S_g} [x \leq v_q]}{n}
\vspace{-4pt}
\end{equation}

where $p_q$ is the proportion of values in $S_g$ that are less than or equal to $v_q$. We define a statistic

\vspace{-5pt}
\begin{equation}
Q_\delta = \frac{1}{N} \sum_{q} (p_q - q)
\vspace{-4pt}
\end{equation}

This formulation has similarities with the Cramer-von Mises $\omega^2$ criterion, but the $Q_\delta$ statistic has one key difference: the asymmetry of the statistic. This stems from the fact that the choice of which sample is considered as $S_h$---the distribution from which $v_q$ is identified---matters. Since we are averaging over the quantiles, this statistic may not yield conclusive guidance for distributions with cumulative distribution functions (CDFs) intersecting at some quantile. Nonetheless, this statistic works best in detecting the dissimilarity of the two samples at the left tail of the distribution which matters most for our purpose. This is because the distributions we are comparing are the DCRs. We want to detect when the distance between the sample and the training data is significantly closer to zero than expected. 

We use the $Q_\delta$ statistic as the basis for detecting overfitting. The threshold against which this statistic will be compared during training is produced through an empirical bootstrapping over random samples from the training data. The details of the bootstrapping method are explained next.

\vspace{-5pt}
\paragraph{$\mathbf{Q_\delta}$ statistic threshold via bootstrapping}

We use three hyperparameters in estimating the threshold that will signal when overfitting occurs during training. First, a sample proportion $\rho$ corresponds to a fraction of the training data. This fraction will be randomly sampled during the bootstrapping and evaluation phases of the generative model training. Second, the $\alpha$ value for choosing the critical threshold for the bootstrap statistic. Third, we specify a bootstrap round $\mathbf{B}$ corresponding to the number of times we compute the $Q_\delta$ statistic between three random samples---two, each of size $\rho$, and the rest having size $1 - 2\rho$ of the training data.

Formally, for a given training data $T_r$ with $N$ observations, we define a bootstrap method to generate a confidence interval for the $Q_\delta$ statistic specific to the tabular data at hand. For each bootstrap round $b \in B$, we take three random samples $S_{tr}$, $S_h$, and $S_g$, without replacement. $S_h$ and $S_g$ are each of size $\rho N$, while $S_{tr}$ contains $(1 - 2\rho) N$ samples. We compute the DCR distributions $\vec{d_g}$ and $\vec{d_h}$ for the two samples $S_h$ and $S_g$, respectively, relative to sample $S_{tr}$. We then compute the $Q_\delta$ statistic between $\vec{d_g}$ and $\vec{d_h}$, where we take $\vec{d_h}$ as the distribution from which we compute the value $v_q$ in Equation \ref{eq:pq}. We store the statistic computed across the bootstrap rounds. We use the specified $\alpha$ to get the cutoff value that will be used as the statistic threshold. We use this threshold $Q_\delta^\prime$ during training to compare the $Q_\delta$ statistic derived from the generated samples by the model. We set $\rho=0.165$, $\alpha=0.95$, and $\mathbf{B}=500$ in our experiments.

\vspace{-5pt}
\paragraph{Early stopping with $\mathbf{Q_\delta^\prime}$}

Our training procedure is paused at each epoch that is a multiple of $E$. We generate data from the model during these epochs. The generated data has size $S_g$. We then take two mutually exclusive random samples from the training data, without replacement, to represent $S_{tr}$ and $S_h$. We compute the $Q_\delta^e$ for this epoch based on the samples generated and drawn. Then, we compare this statistic to the previously computed threshold $Q_\delta^\prime$. We continue training the model if $Q_\delta^e < Q_\delta^\prime$. We save a checkpoint of this model. We terminate the model training when $Q_\delta^e > Q_\delta^\prime$ for $X$ consecutive epochs. We then load the checkpoint for the most recent model that satisfied the condition $Q_\delta^e < Q_\delta^\prime$. In our experiments, we set $E=5$ as the period of our overfitting evaluation and $X=2$ as our grace period before training termination.

\subsubsection{Sampling}

The models we use build each observation sequentially, one token at a time. We leverage the structure of our data processing to optimize the generation of samples from the trained models. Using a vocabulary specific to a column in the input data allows us to implement a constrained generation of tokens for each column.

We track the token ids that form the domain of each column during the generation of the vocabulary using a hash map. Based on this, the tokens that are invalid for the columns will not be considered for generation in the timestep representing the column. This strategy allows for efficient sampling wherein the likelihood of generating an invalid sample is close to zero. In our experiments, $\ll 1\%$ invalid samples are generated during the sampling phase.

\input{tables/relational_ld_results.tex}

\section{Experiments and Results}

This section outlines the evaluation process we conducted to quantify the performance of the proposed REaLTabFormer framework compared with baseline models. We first demonstrate that the performance of the model we use to generate the parent tables, and non-relational tabular data in general, compares with or exceeds the performance of state-of-the-art models in real-world tabular data generation tasks measured by the machine learning efficacy metric. We also use the discriminator measure to quantify how realistic the samples generated by each model are. We proceed to model real-world relational datasets and show, quantitatively using logistic detection, that the synthetic data produced by the REaLTabFormer are more realistic and accurate.

\subsection{Data}
\label{data}

We use a collection of real-world datasets, listed in Table~\ref{tab:data-summary}, commonly used in previous works for non-relational tabular data generation \citep{xu_modeling_2019,zhao2021ctab,gorishniy2021revisiting,borisov_language_2022,kotelnikov2022tabddpm}. These datasets differ with respect to the number of observations, ranging from 768 up to 197,080 observations. There is also variation in the number of variables they contain, ranging from 8 to 50 numerical variables and 0 up to 8 categorical variables. The datasets cover regression, binary, and multi-class classification prediction tasks.

\begin{figure}[ht]
\vskip 0.1in
\begin{center}
\centerline{\includegraphics[width=\columnwidth]{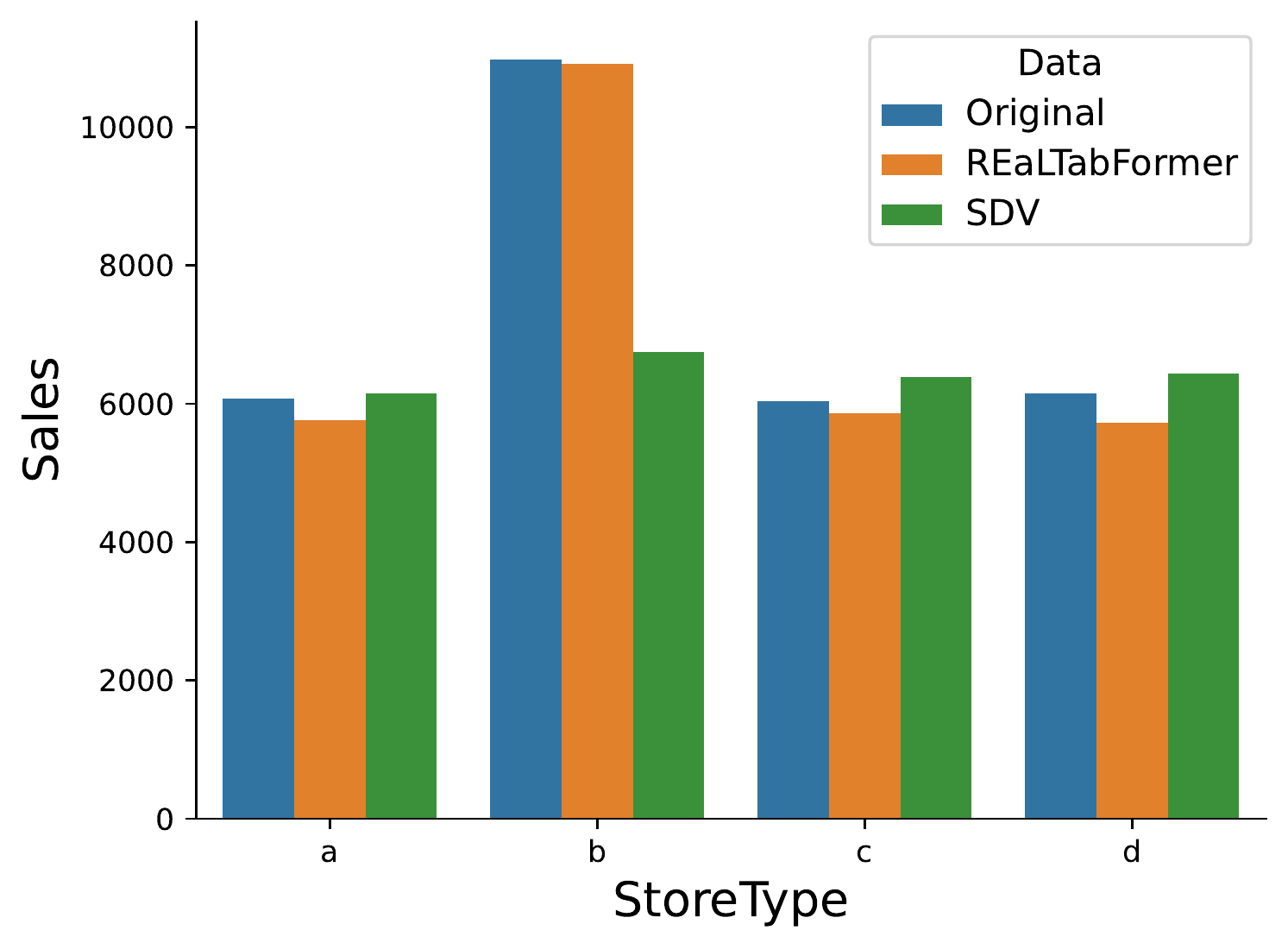}}
\caption{Summary of the average ``Sales" variable in the child table of the Rossmann dataset grouped by ``StoreType" variable in the parent table. The values shown are from the original data (blue), synthetic data produced by REaLTabFormer (orange), and data generated by SDV (green). This graph shows that REaLTabFormer captures the inter-table variations and relationships well.}
\label{fig:rossmann-storetype-sales}
\end{center}
\vskip -0.1in
\end{figure}

We use two real-world datasets to compare the performance of the REaLTabFormer on modeling relational tabular data compared with the baseline. These datasets are the Rossmann dataset and the Airbnb dataset used in prior work on synthetic relation data generation \citep{patki_synthetic_2016}.

\subsection{Baseline models}

\paragraph{Non-relational tabular data}

We use models that apply different deep learning architectures for generating non-relational tabular data as baselines to compare the REaLTabFormer model with. The TVAE is based on variational autoencoder \citep{xu_modeling_2019}, the CTABGAN+ on GAN architecture \citep{zhao2022ctab}, the Tab-DDPM on diffusion \citep{kotelnikov2022tabddpm}, and GReaT uses pretrained LLM \citep{borisov_language_2022}.

\paragraph{Relational datasets}

Models for generating relational datasets are limited. \citet{gueye2022row} published work on using GAN for relational datasets but no open-sourced implementation is available. We choose to limit our baselines to open-sourced models; hence, we only use the Hierarchical Modeling Algorithm (HMA) available in the Synthetic Data Vault (SDV) as our baseline \citep{patki_synthetic_2016}.

\subsection{Generative models training}

The GReaT model was trained for 100 epochs for each data. The parameters for the TVAE, CTABGAN+, and Tab-DDPM models had been tuned for the predictive task itself using the real validation data from \citet{kotelnikov2022tabddpm}. For the relational datasets, we trained the HMA model as prescribed in the SDV documentation. In contrast, the REaLTabFormer model was not tuned against any of the machine learning tasks. The model solely relied on the overfitting metric discussed in Section \ref{overfitting}. We used the same parameters for the different datasets to test how the REaLTabFormer performs ``out-of-the-box".

\subsection{Measures and Results}

We select the following measures to quantify the quality and utility of the generated samples by the generative models.

\paragraph{Machine Learning (ML) efficacy}

The machine learning (ML) efficacy \citep{xu_modeling_2019,kotelnikov2022tabddpm,borisov_language_2022} measures the potential utility of the synthetic data to supplant the real data for machine learning tasks, in particular, training a prediction model. The ML efficacy reported by \citet{borisov_language_2022} in their work used ML models that were not fine-tuned. \citet{kotelnikov2022tabddpm} showed that the ML efficacy computed from models that are not fine-tuned may show spurious results. They instead optimized the ML models---CatBoost \citep{prokhorenkova2018catboost}---they used in reporting the ML efficacy. This approach is closer to what researchers are expected to do in the real world, therefore, we adopt these tuned models in our experiments. We generate a validation set from the generative models to signal the early-stopping condition during the ML model training. This is in contrast with the method used by \citet{kotelnikov2022tabddpm} where they still relied on the real validation data for the early-stopping of the ML model.

We report the macro average F1 score \citep{opitz2019macro} for classification tasks and the $R^2$ metric for regression tasks. Our results presented in Table~\ref{tab:mle-dm-results} (MLE) show that REaLTabFormer, despite not being fine-tuned, produces ML efficacy scores that are the best or second-best compared with the baselines. This demonstrates that REaLTabFormer can be used, ``out-of-the-box", to generate synthetic data with state-of-the-art performance in machine learning tasks.

The FB comments dataset, where REaLTabFormer obtained the best performance, is the largest dataset tested and has the largest number of columns. Training the GReaT model on this dataset yielded impractical runtime so no result is reported. This supports our view that using LLM trained on a large vocabulary, containing a majority of irrelevant tokens, limits the efficiency of the model.

\vspace{-5pt}
\paragraph{Discriminator measure}

We adopt the discriminator measure \citep{borisov_language_2022} to quantify whether the data generated by a model is easily distinguishable from real data. A dataset is made by combining an equal number of real and synthetic data. Real observations in this dataset are labeled as “1” and synthetic observations are labeled as “0”. Similar to \citet{borisov_language_2022}, we train a random forest model to predict the labels given an observation. A held-out dataset containing a combination of synthetic samples and real test data is then used to report the final measure.

An accuracy that is closer to 50\% implies better synthetic data quality since the discriminative model is not able to distinguish the real from the synthetic observations. We report our results in Table~\ref{tab:mle-dm-results} (DM). The DM measure shows that the Tab-DDPM has the most indistinguishable synthetic data. Nonetheless, REaLTabFormer, without the need for tuning, has DM measures that are close to the Tab-DDPM. This suggests that the synthetic data produced by a diffusion-based model and REaLTabFormer are realistic compared with the other baselines.

\vspace{-2pt}
\paragraph{Logistic Detection} 

For relational datasets, we use logistic detection (LD) \citep{fisher2019machine,gueye2022row} to quantify the quality of the parent, child, and merged tables generated by REaLTabFormer compared with the HMA model. We evaluate ROC-AUC scores averaged over (N=3) cross-validation folds,

\vspace{-10pt}
\begin{equation}
\mu_{\mathrm{RA}} = \frac{1}{N}\sum_{i=1}^N \max (0.5, \mathrm{ROC-AUC}) \times 2 - 1    
\vspace{-5pt}
\end{equation}

The value reported is $LD=100 \times (1 - \mu_{\mathrm{RA}})$, where scores range from 0 to 100, and scores closer to 100 imply better synthetic data quality. We use random forest in measuring the logistic detection instead of the standard logistic regression model. The random forest model captures non-linearity in the data well than logistic regression \citep{couronne2018random}, reducing the likelihood of spurious results. We report the results in Table~\ref{tab:relational-ld-results}. Additional LD results using logistic regression are shown in Table~\ref{tab:relational-ld-logistic-results}.

The REaLTabFormer model produces significantly higher-quality synthetic data than the HMA model across the datasets tested. The high values of LD for the child and the merged tables highlight the ability of REaLTabFormer to accurately synthesize relational datasets in comparison with the leading baseline. The LD metric shows that data generated by SDV for the Airbnb child table is entirely distinguishable from the real data. The quantitative results are supported by relational statistics computed from synthetic datasets produced by REaLTabFormer and the HMA model: \cref{fig:rossmann-date-sales,fig:airbnb-device_type-age_group,fig:rossmann-storetype-sales}. 

\section{Conclusion}

We presented REaLTabFormer, a framework capable of generating high-quality non-relational tabular data and relational datasets. This work extends the application of sequence-to-sequence models to modeling and generating relational datasets. We introduced target masking as a component in the model to mitigate data-copying and safeguarding from potentially sensitive data leaking from the training data. We proposed a statistical method and the $Q_\delta$ statistic for detecting overfitting in model training. This statistical method may be adapted to other generative model training. We showed that our proposed model generates realistic synthetic tabular data that can be a proxy for real-world data in machine learning tasks. REaLTabFormer's ability to model relational datasets accurately compared with existing open-sourced alternative contributes to solving existing gaps in generative models for realistic relational datasets. Finally, this work can be extended and applied to data imputation, cross-survey imputation, and upsampling for machine learning with imbalanced data. A BERT-like encoder can be used instead of GPT-2 with the REaLTabFormer for modeling relational datasets. We also see opportunities to improve privacy protection strategies and the development of more components like target masking embedded into synthetic data generation models to prevent sensitive data exposure.

\pagebreak
\section{REaLTabFormer Python Package}

We publish the REaLTabFormer as a package on PyPi. We show below how the model can be easily trained on any tabular dataset, loaded as a Pandas DataFrame.

\vspace{-5pt}
\subsection{Non-relational tabular model}
\label{appendix:python-non-relational}
Use the following snippet to fit the REaLTabFormer on a non-relational tabular dataset. One can control the various hyper-parameters of the model and the fitting method, e.g., the number of bootstrap rounds \texttt{num\_bootstrap}, the fraction of training data \texttt{frac} used to generate the $\mathbf{Q}_\delta$ statistic, etc. Keyword arguments for the HuggingFace transformers \texttt{Trainer} class can also be passed as \texttt{**kwargs} when initializing the model.

\begin{minted}[
frame=lines,
framesep=2mm,
bgcolor=BrickRed!2,
baselinestretch=1.2,
fontsize=\scriptsize,
xleftmargin=10pt,
numbersep=5pt,
linenos
]{python}
# pip install realtabformer
import pandas as pd
from realtabformer import REaLTabFormer

# NOTE: Remove any unique identifiers in the 
# data that you don't want to be modeled.
df = pd.read_csv("foo.csv")

# Non-relational or parent table.
rtf_model = REaLTabFormer(
    model_type="tabular", 
    gradient_accumulation_steps=4)

# Fit the model on the dataset.
# Additional parameters can be 
# passed to the `.fit` method.
rtf_model.fit(df)

# Save the model to the current directory.
# A new directory `rtf_model/` will be created.
# In it, a directory with the model's 
# experiment id `idXXXX` will also be created 
# where the artefacts of the model will be stored.
rtf_model.save("rtf_model/")

# Generate synthetic data with the same 
# number of observations as the real dataset.
samples = rtf_model.sample(n_samples=len(df))

# Load the saved model. The directory to the
# experiment must be provided.
rtf_model2 = REaLTabFormer.load_from_dir(
    path="rtf_model/idXXXX")
\end{minted}

\vspace{-10pt}
\subsection{Non-relational tabular model}

REaLTabFormer for relational databases requires a two-phase training. First, the model for the parent table is trained as a non-relational tabular data, then saved. Second, we pass the path of the saved parent model when creating the REaLTabFormer instance for the child model to be used as its encoder, then train. Generate synthetic samples from the parent table and use as input to the trained child model to generate the synthetic relational observations. 

\begin{minted}[
frame=lines,
framesep=2mm,
bgcolor=BrickRed!2,
baselinestretch=1.2,
fontsize=\scriptsize,
xleftmargin=10pt,
numbersep=5pt,
linenos
]{python}
# pip install realtabformer
import os
import pandas as pd
from realtabformer import REaLTabFormer

pdir = Path("rtf_parent/")
parent_df = pd.read_csv("foo.csv")
child_df = pd.read_csv("bar.csv")
join_on = "unique_id" 

# Make sure that the key columns in both the 
# parent and the child table have the same name.
assert ((join_on in parent_df.columns) and 
        (join_on in child_df.columns))

# Non-relational or parent table. Don't include the 
# unique_id field.
parent_model = REaLTabFormer(model_type="tabular")
parent_model.fit(parent_df.drop(join_on, axis=1))
parent_model.save(pdir)

# # Get the most recently saved parent model,
# # or a specify some other saved model.
# parent_model_path = pdir / "idXXX"
parent_model_path = sorted([
    p for p in pdir.glob("id*") if p.is_dir()],
    key=os.path.getmtime)[-1]

child_model = REaLTabFormer(
    model_type="relational",
    parent_realtabformer_path=parent_model_path,
    output_max_length=None, train_size=0.8)

child_model.fit(
    df=child_df, in_df=parent_df, join_on=join_on)

# Generate parent samples.
parent_samples = parent_model.sample(len(parend_df))

# Create the unique ids based on the index.
parent_samples.index.name = join_on
parent_samples = parent_samples.reset_index()

# Generate the relational observations.
child_samples = child_model.sample(
    input_unique_ids=parent_samples[join_on],
    input_df=parent_samples.drop(join_on, axis=1),
    gen_batch=64)
\end{minted}

\vspace{-10pt}

\paragraph{Acknowledgments} 


This project was supported by the “Enhancing Responsible Microdata Access to Improve Policy and Response in Forced Displacement Situations” project funded by the World Bank-UNHCR Joint Data Center on Forced Displacement (JDC) – KP-P174174-GINP-TF0B5124. We also thank Patrick Brock for providing insightful comments. The findings, interpretations, and conclusions expressed in this paper are entirely those of the authors. They do not necessarily represent the views of the International Bank for Reconstruction and Development/World Bank and its affiliated organizations, or those of the Executive Directors of the World Bank or the governments they represent.





\pagebreak

\bibliography{realtabformer}
\bibliographystyle{icml2023}

\newpage
\appendix
\twocolumn
\section{Raw data processing}
\label{appendix:raw-data-processing}

\paragraph{Numerical data}

Various methods have been proposed for representing numerical data as input to generative models in the context of tabular data. The CTGAN and TVAE models suggest the use of gaussian mixture models to encode numerical values \citep{xu_modeling_2019}. On the other hand, the TabFormer model introduced quantization as a way to encode numeric data \citep{padhi_tabular_2021}. However, these approaches are lossy. As argued by \citet{borisov_language_2022}, these lossy transformations may not be optimal.

In our model, we adopt a fully text-based strategy in handling numerical values. We apply a sequence of transformations that converts a column of numeric value into, possibly, multi-columnar data. We use the following transformation of numerical columns. We also show the outcome of each transformation step on the sample numerical series below. 

For illustration, this example numerical-valued series $$[1032.325345, 10.291, \m3.0]$$ is converted into

$$
\begin{matrix}
    [``10", ``32", ``.3", ``3"] \\
    [``00", ``10", ``.2", ``9"] \\
    [``\m0", ``03", ``.0", ``0"]
\end{matrix}
$$




\begin{itemize}
	\item We set a rounding resolution to normalize the size of the numerical values. For example, round to at most 2 decimal places. 
\begin{itemize}
    \item $[1032.33, 10.29, \m3.0]$
\end{itemize} 
	\item We then cast the values to string. 
\begin{itemize}
    \item $[``1032.33", ``10.29", ``\m3.0"]$
\end{itemize}
	\item We identify the magnitude of the most significant digit of the largest value in the column by looking for the location of the decimal point of the largest value. The magnitude of the most significant digit for the largest value in the example is 4.
\begin{itemize}
    \item $[``\underline{1032}.33", ``10.29", ``\m3.0"]$
\end{itemize}
        \item We use the magnitude to left-align all the other values in the data by padding them with leading zeros.
\begin{itemize}
    \item $[``1032.33", ``\underline{00}10.29", ``\underline{00}\m3.0"]$
\end{itemize}
        \item We then take the length of the longest string after this transformation and left-justify the data by padding zeros to the right of the values that are shorter than the longest string. 
\begin{itemize}
    \item $[``1032.33", ``0010.29", ``00\m3.0\underline{0}"]$
\end{itemize}
        \item Then, the negative sign for negative values is transposed to the leftmost part of the string. 
\begin{itemize}
    \item $[``1032.33", ``0010.29", ``\underline{\m}0\underline{0}3.00"]$
\end{itemize}
    \item Note that for integral values, we only perform the left alignment by padding the values with leading zeros.
\end{itemize}

After this series of transformations, we tokenize the values into fixed-length partitions. For the same example values, say we choose the partition size to be 2, we get the following tokenized table.

$$
\begin{matrix}
    [``10", ``32", ``.3", ``3"] \\
    [``00", ``10", ``.2", ``9"] \\
    [``\m0", ``03", ``.0", ``0"]
\end{matrix}
$$

This transformation is done to mitigate the explosion of the vocabulary if the numeric values are all distinct. We found in our experiments that using single-character partitioning works best. We suppose that this effect is attributable to the inherent regularization of generating an entire sequence of numbers one digit at a time.



\paragraph{Datetime data}

For date or time data types, we first perform a transformation of the raw data into Unix timestamp representation. This representation is then treated as regular numeric data; hence, we apply the data processing discussed for numeric data types. 

\paragraph{Categorical data}

Unique values in categorical columns are treated as unique tokens in the vocabulary. No additional processing is done.

\paragraph{Missing values}

No transformation is done for missing values present in the data. We let the model learn the distribution of the missing values. This strategy gives us the flexibility to let the model impute or generate missing values during the sampling process.

\paragraph{Input data aggregation}

As illustrated above, the transformation of numerical data types expands the dataset by partitioning the string version of the values. As such, we combine the processed columns into modified tabular data. We use this modified tabular data as input for our models. Each unique value in the new columns in this data will be mapped to a unique token in the vocabulary that is independent of values in the other columns. This means that in the illustrated numerical transformation shown above, the “1” in the first column will have a different token id than the “1” present in the third column.

\onecolumn

\input{tables/data_summary.tex}

\section{Datasets}
\label{appendix:datasets}

\subsection{Non-relational tabular data}

We used six real-world datasets to assess the performance of our proposed model for generating realistic and useful synthetic tabular data. The datasets are diverse with respect to the types of variables---mix of numerical and categorical data types---as well as the number of variables in each dataset---ranging from 8 to 51 columns. The collection includes, Abalone (OpenML)\footnote{\href{https://www.openml.org/search?type=data&sort=runs&id=183&status=active}{Abalone (OpenML)}}, Adult (income estimation)\footnote{\href{https://archive.ics.uci.edu/ml/datasets/adult}{Adult (income estimation)}}, Buddy (Kaggle)\footnote{\href{https://www.kaggle.com/datasets/akash14/adopt-a-buddy}{Buddy (Kaggle)}}, California Housing (real estate data)\footnote{\href{https://www.dcc.fc.up.pt/~ltorgo/Regression/cal_housing.html}{California Housing (real estate data)}}, Diabetes (OpenML)\footnote{\href{https://www.openml.org/search?type=data&sort=runs&id=37&status=active}{Diabetes (OpenML)}}, and Facebook Comments\footnote{\href{https://archive.ics.uci.edu/ml/datasets/Facebook+Comment+Volume+Dataset}{Facebook Comments}}. Original source, copyright, and license information are available in the links in the footnote.

We used the data splits by \citet{kotelnikov2022tabddpm} published in \href{https://github.com/rotot0/tab-ddpm}{Tab-DDPM GitHub}. Based on their pickled numpy data dumps, we recreated the splits to create data frames that we can use for our experiments with REaLTabFormer and GReaT. The latter model expects contextual input from the column names.

We also used the open-sourced optimized model parameters published in the above GitHub repo after reviewing the code, and the correctness of the code relevant to producing the assets of interest has been confirmed. We trained the TVAE, CTABGAN+, and Tab-DDPM models from scratch using the parameters on each dataset.

\subsection{Relational tabular data}

To test the REaLTabFormer in modeling relational datasets, we used two real-world data: the Rossmann store sales\footnote{\href{https://www.kaggle.com/competitions/rossmann-store-sales/data}{Rossmann store sales}} dataset and the Airbnb new user bookings\footnote{\href{https://www.kaggle.com/competitions/airbnb-recruiting-new-user-bookings/data}{Airbnb new user bookings}} dataset.

We created train and test splits. For the Rossmann dataset, we used 80\% of the stores data and their associated sales records for our training data. We used the remaining stores as the test data. We also limit the data used in the experiments from 2015-06 onwards spanning 2 months of sales data per store. In the Airbnb dataset, we considered a random sample of 10,000 users for the experiment. We take 8,000 as part of our training data, and we assessed the metrics and plots using the 2,000 users in the test data. We also limit the users considered to those having at most 50 sessions in the data.

\section{Reproducibility}

We used \texttt{be\_great==0.0.3} for the GReaT model. We used the Tab-DDPM GitHub repo version with this permanent link \href{https://github.com/rotot0/tab-ddpm/tree/41f2415a378f1e8e8f4f5c3b8736521c0d47cf22}{https://github.com/rotot0/tab-ddpm/tree/41f2415a378f1e8e8f4f5c3b8736521c0d47cf22}. We used \texttt{sdv==0.17.2} and \texttt{sdmetrics==0.8.1}; however, we fixed a bug in the \texttt{HyperTransformer} implementation. We used \texttt{transformers==4.25.1} and \texttt{torch==1.13.1}. We will open-source the REaLTabFormer package and experiments repository. We used \texttt{Python version 3.9}.

We ran our experiments on a standalone workstation with the following specs: 2x AMD EPYC 7H12 64-Core Processor, 2x RTX 3090 GPU, and 1TB RAM running Ubuntu 20.04 LTS.

\section{Other measures and results}

We also computed the logistic detection measure with the standard approach of using a logistic regression model. We find that the logistic regression model appears to not provide reliable results Table~\ref{tab:relational-ld-logistic-results}. In particular, the scores returned by the model are too high which is suspicious given that qualitative observation of the synthetic data hints at inaccuracies by both models in producing perfect alignment with the original data. These spurious results may be due to the model's limited capacity of learning the structure of the data. While techniques can be applied to help the model detect non-linearities better, we opted to report the results using the random forest as the base detector since it naturally is able to learn non-linearities and appears to give reasonable results.

\input{tables/relational_ld_logistic_results.tex}

\subsection{Joint plots}

The joint plot provides a qualitative assessment of the quality of the synthetic data generated by each model. We show in the sequence of figures below the joint plots of two numerical variables in the datasets used.

\newcommand\imgjointwidth{0.28}

\begin{figure}[H]
    \centering
    \includegraphics[width=\imgjointwidth\textwidth]{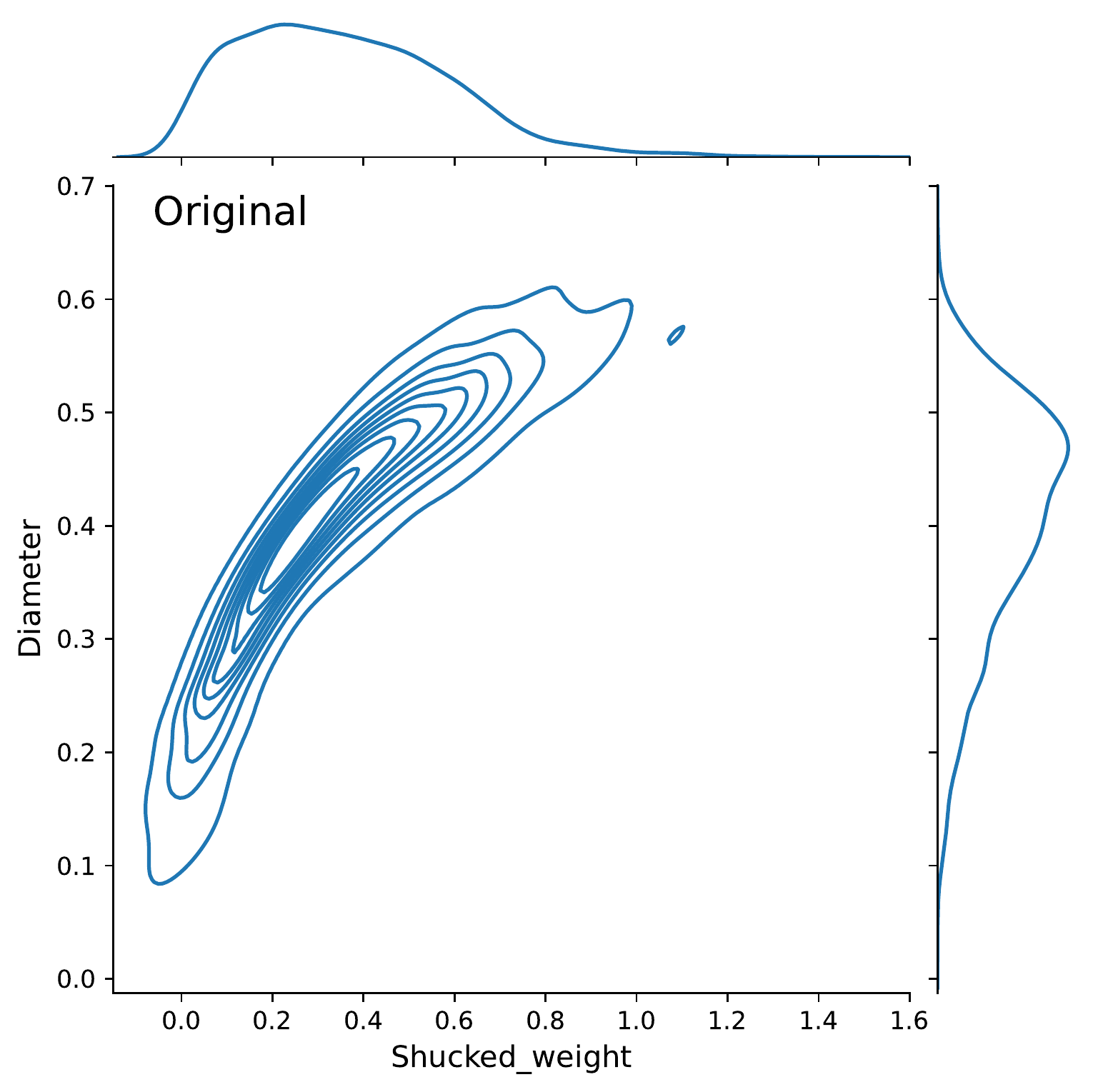}
    \includegraphics[width=\imgjointwidth\textwidth]{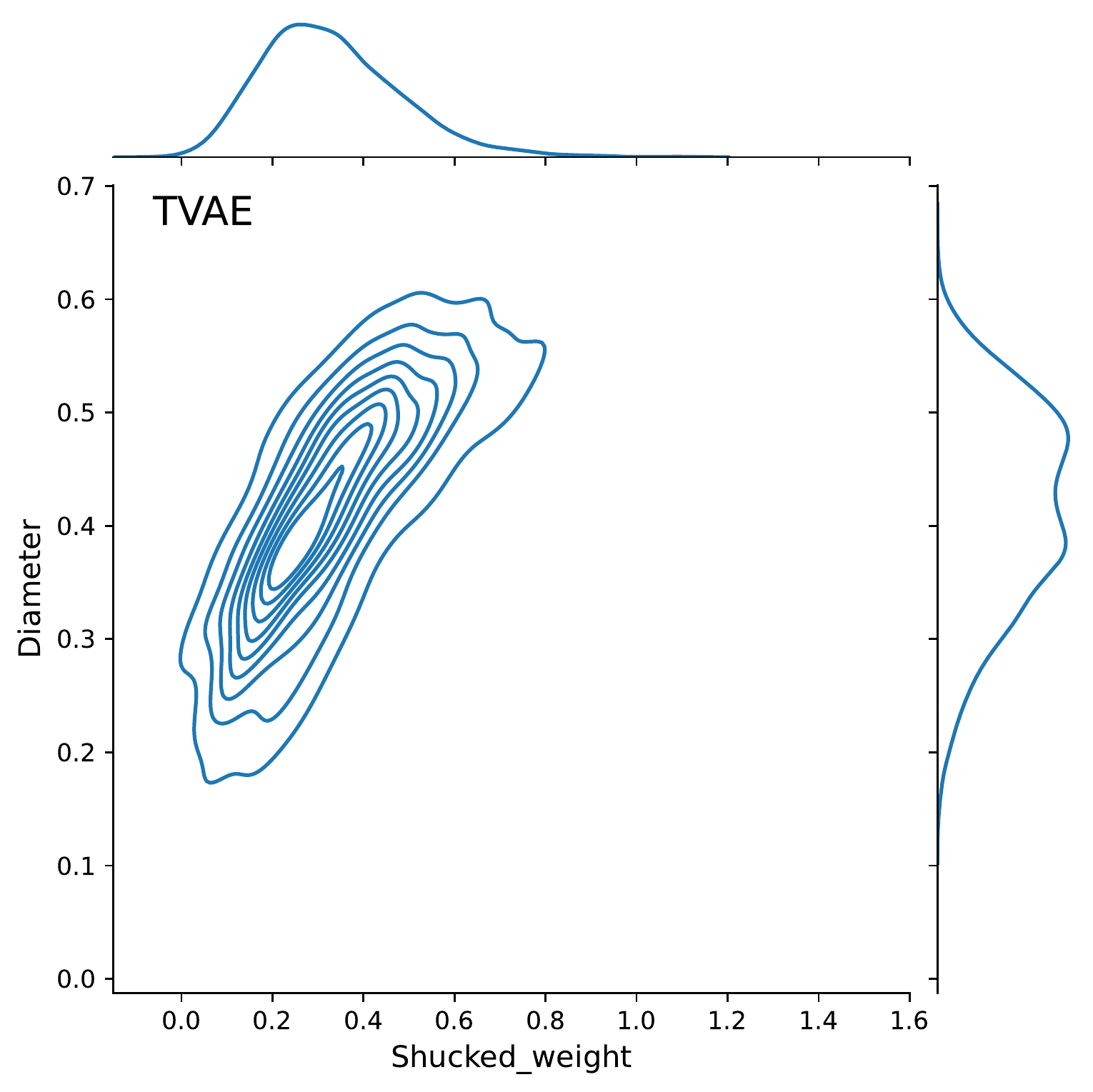}
    \includegraphics[width=\imgjointwidth\textwidth]{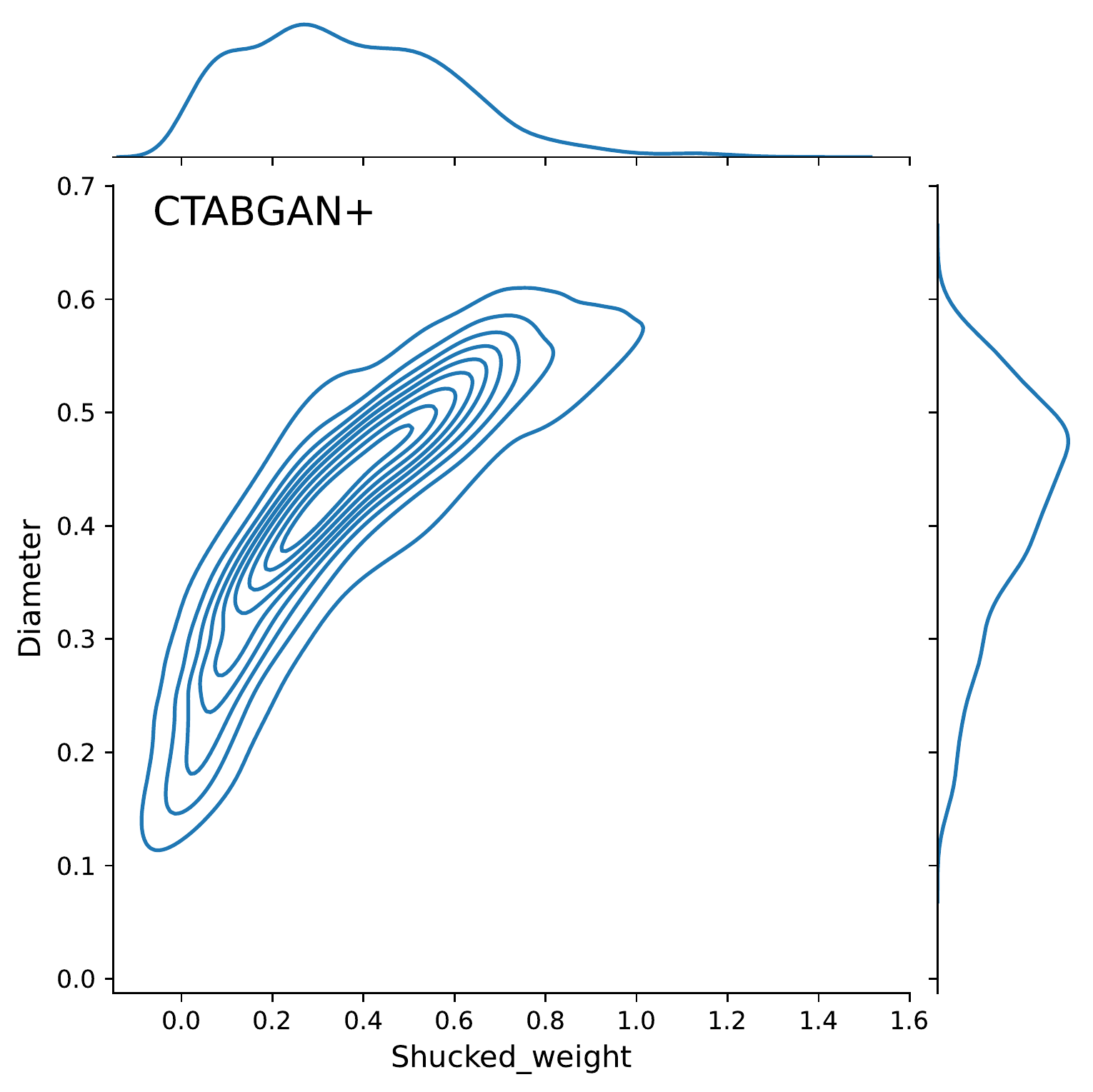}
    \includegraphics[width=\imgjointwidth\textwidth]{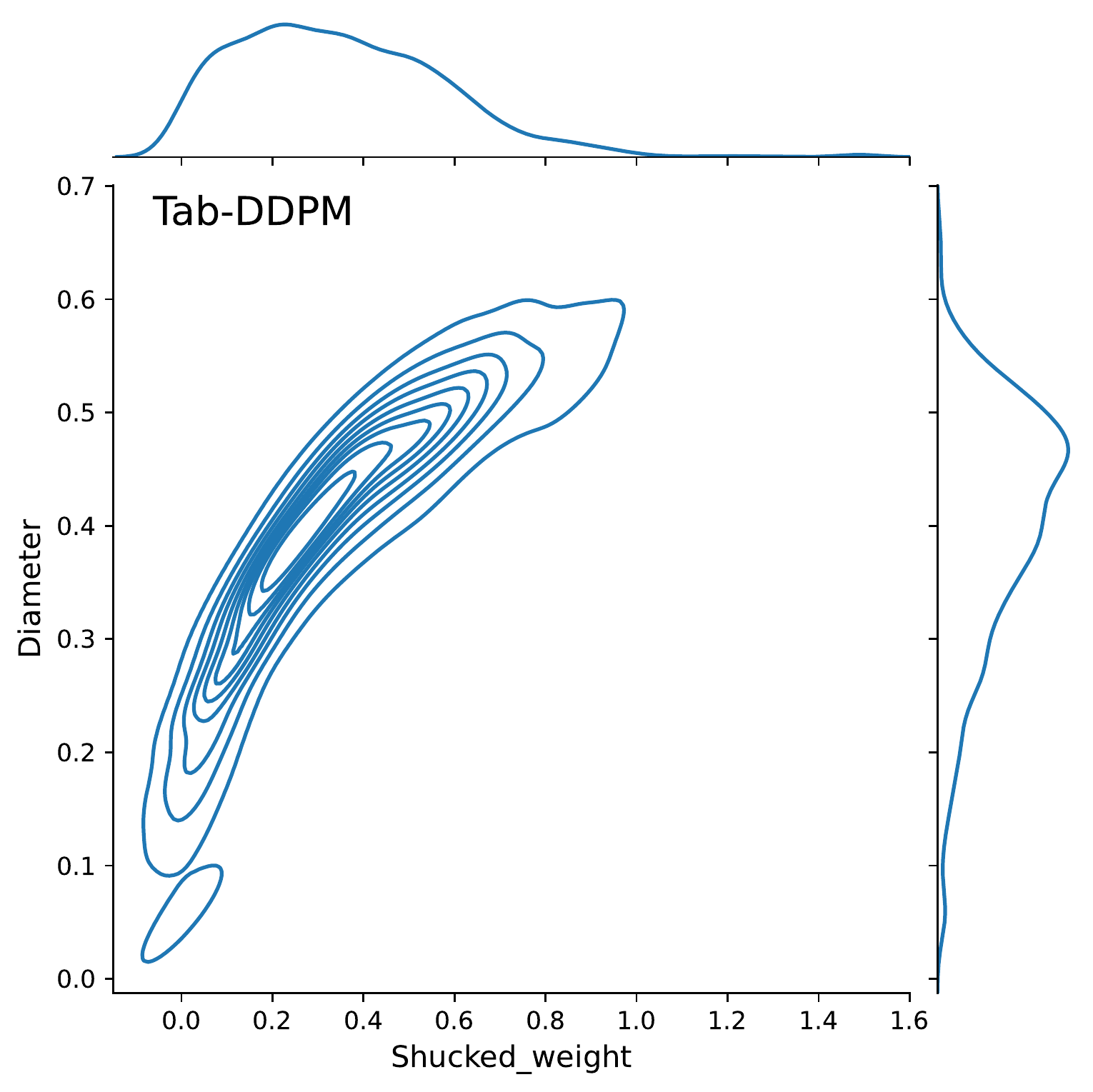}
    \includegraphics[width=\imgjointwidth\textwidth]{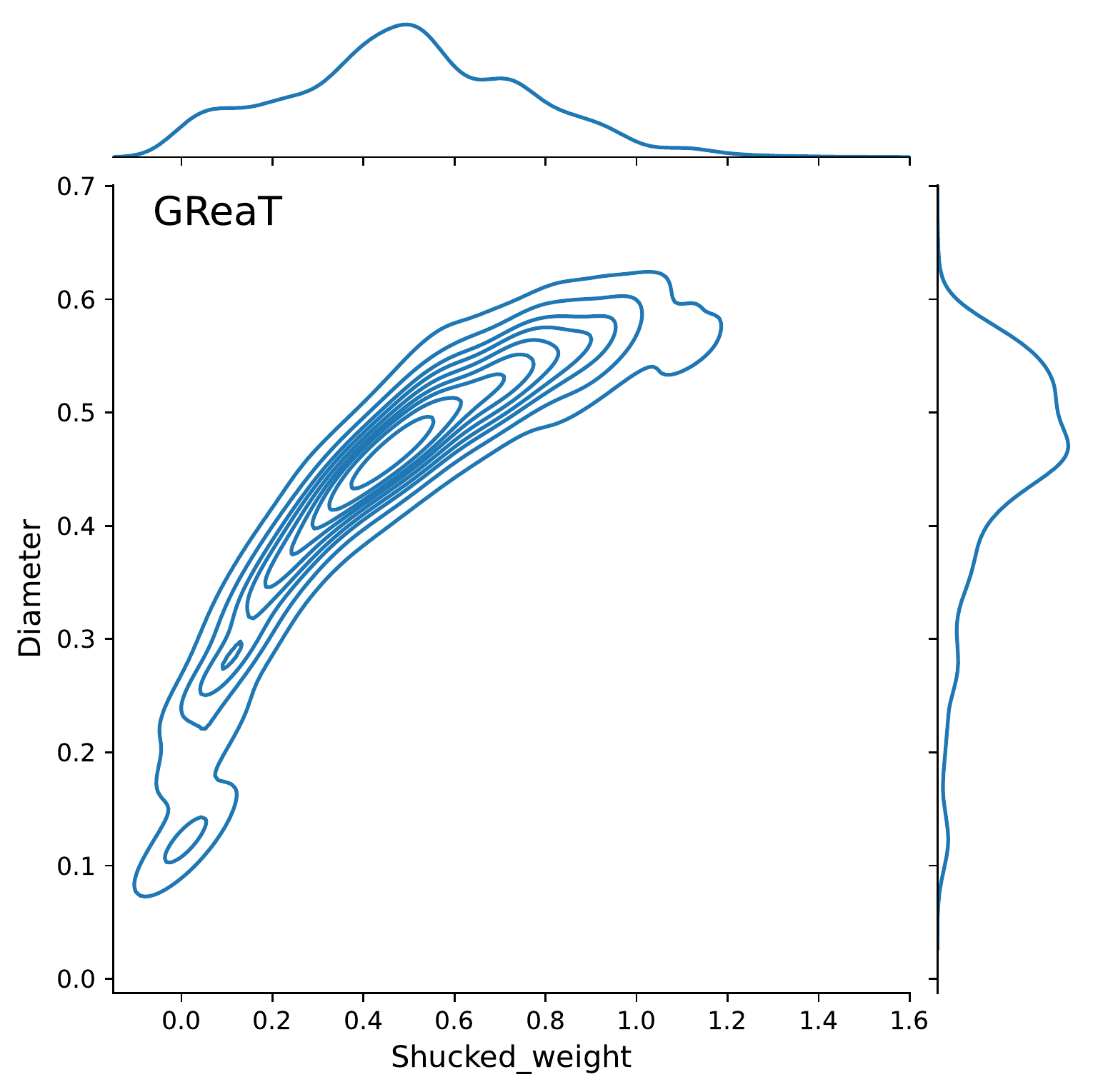}
    \includegraphics[width=\imgjointwidth\textwidth]{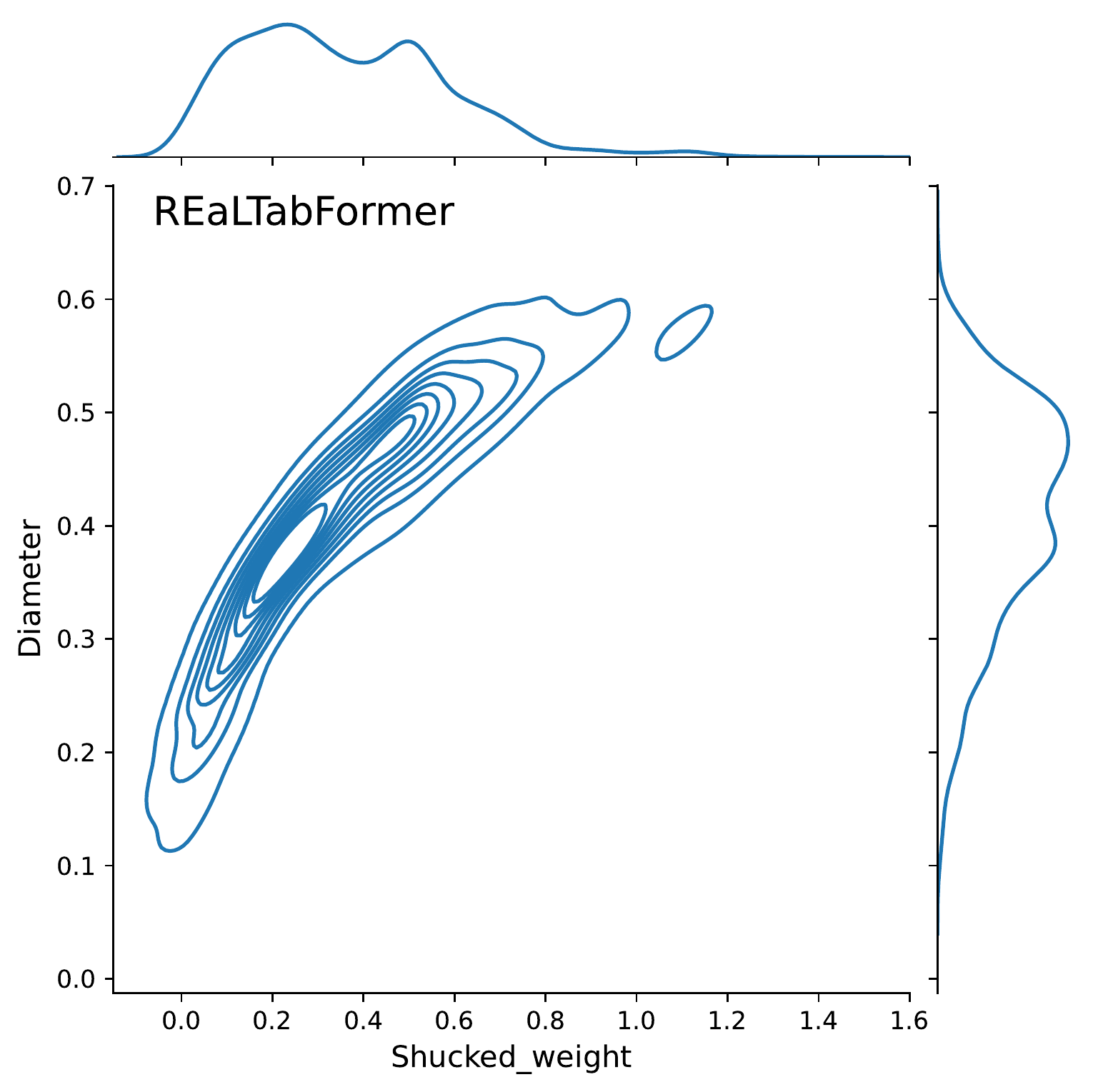}
    \caption{Joint plot of two numerical variables in the Abalone data compared across the samples generated by the different models.}
    \label{fig:joint-abalone}
\end{figure}

\begin{figure}[H]
    \centering
    \includegraphics[width=\imgjointwidth\textwidth]{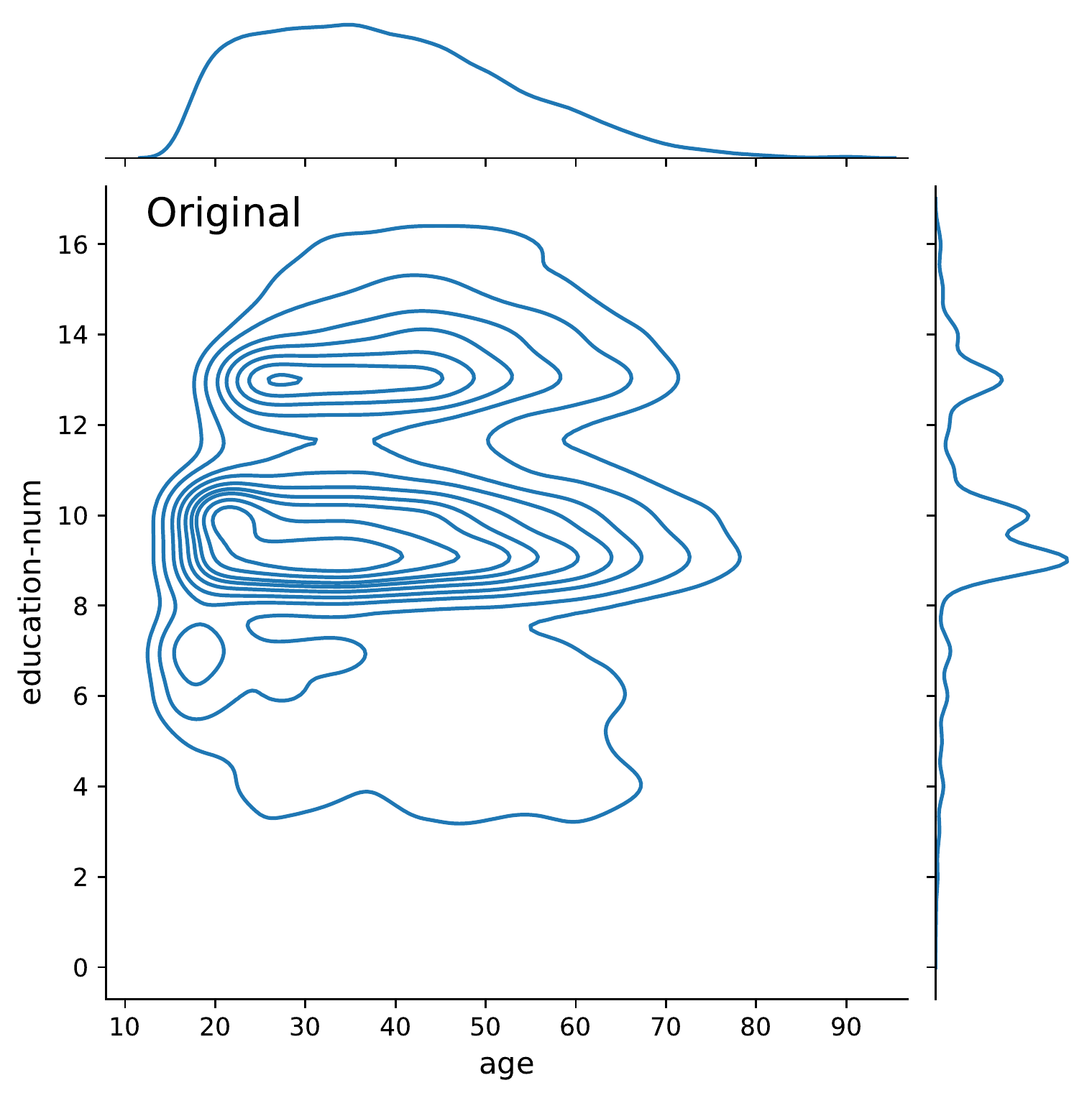}
    \includegraphics[width=\imgjointwidth\textwidth]{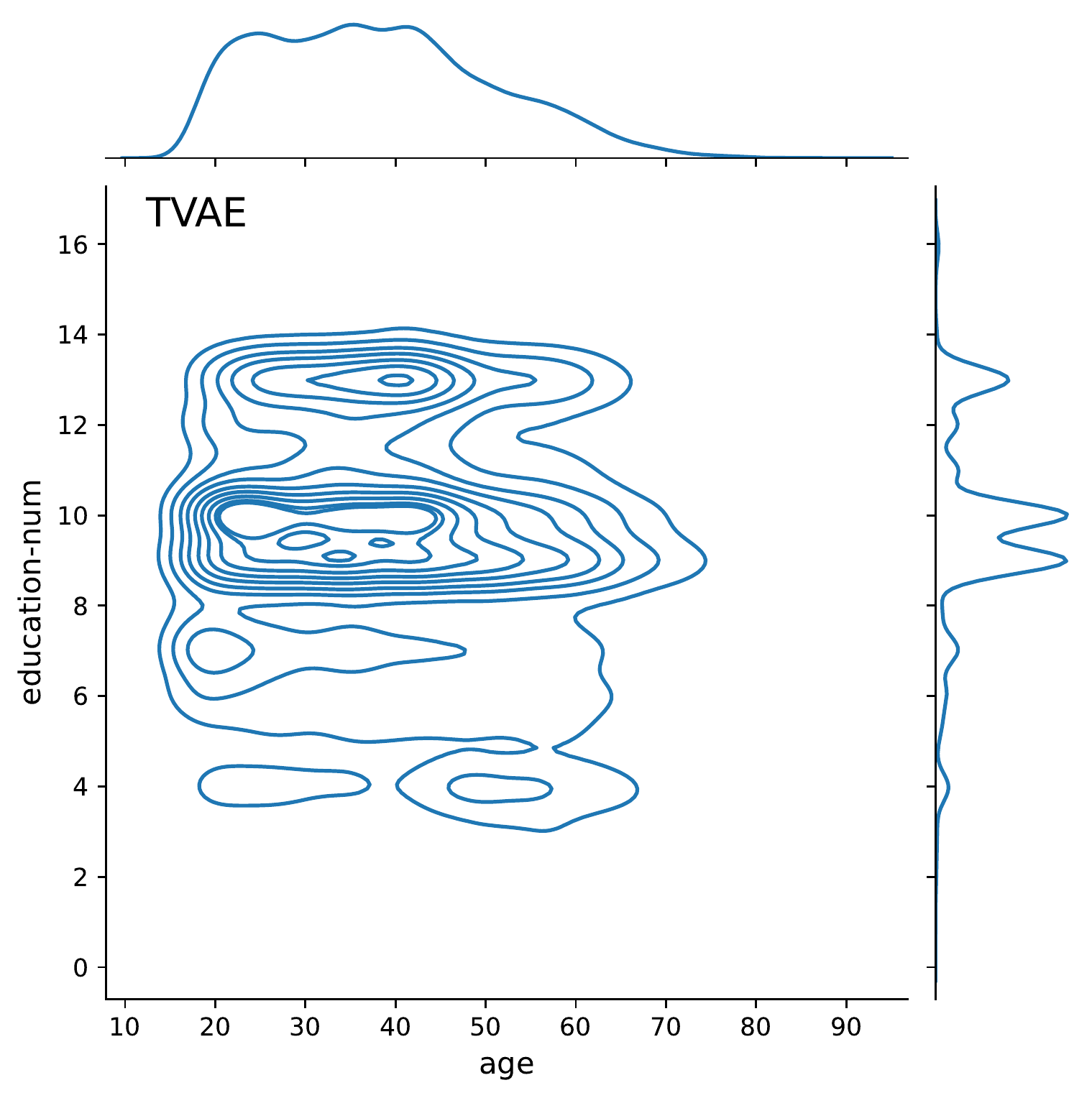}
    \includegraphics[width=\imgjointwidth\textwidth]{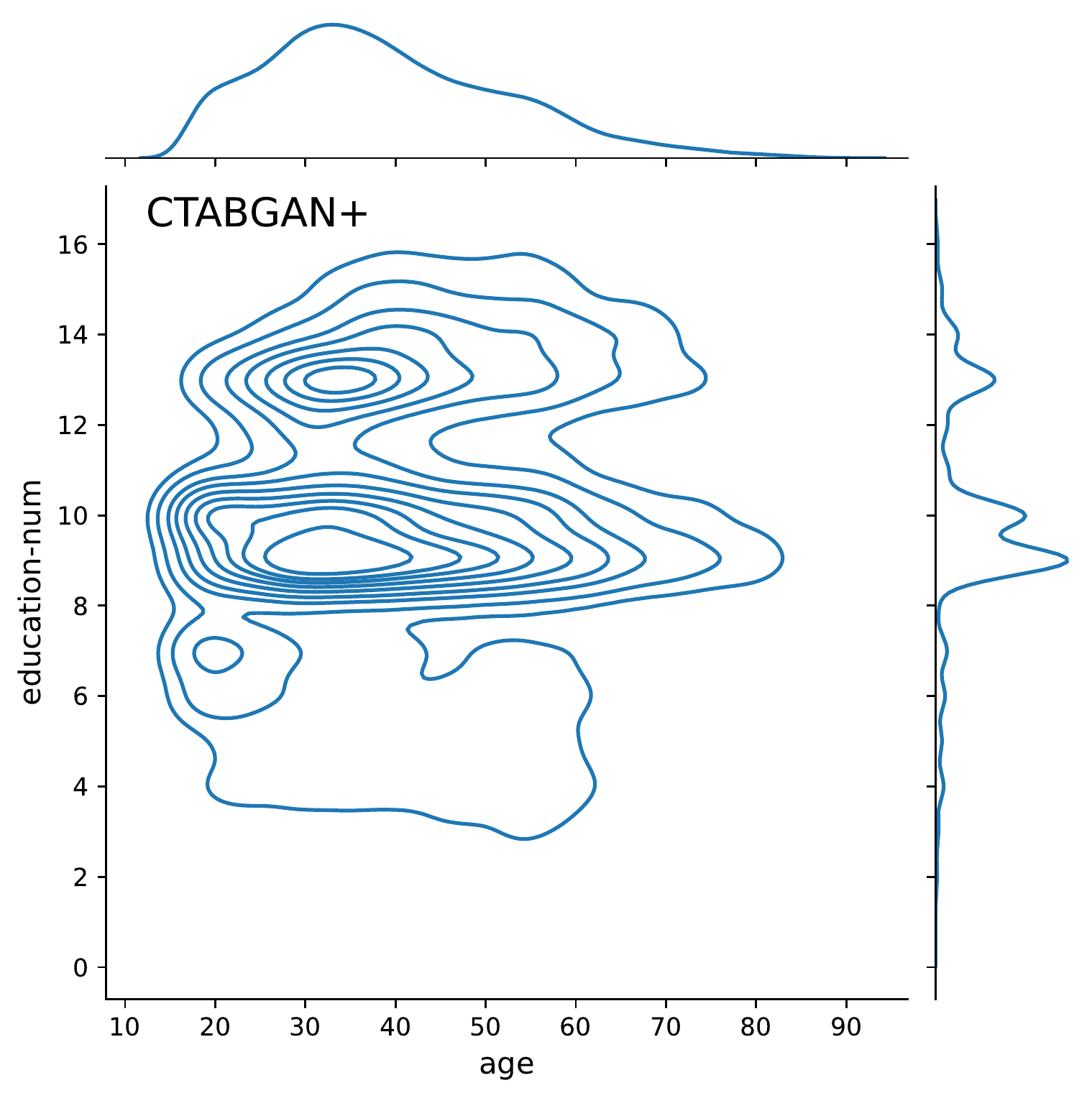}
    \includegraphics[width=\imgjointwidth\textwidth]{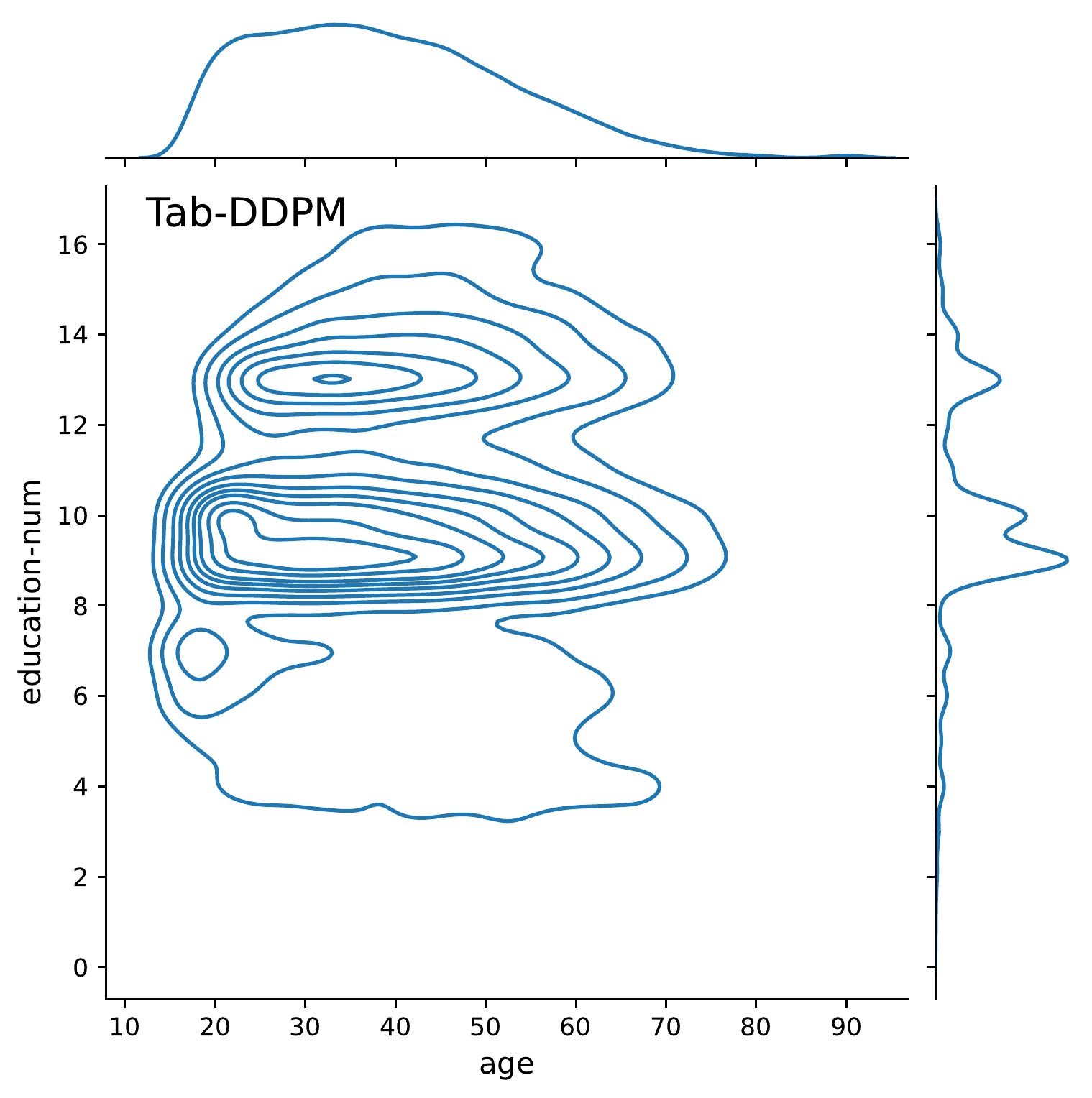}
    \includegraphics[width=\imgjointwidth\textwidth]{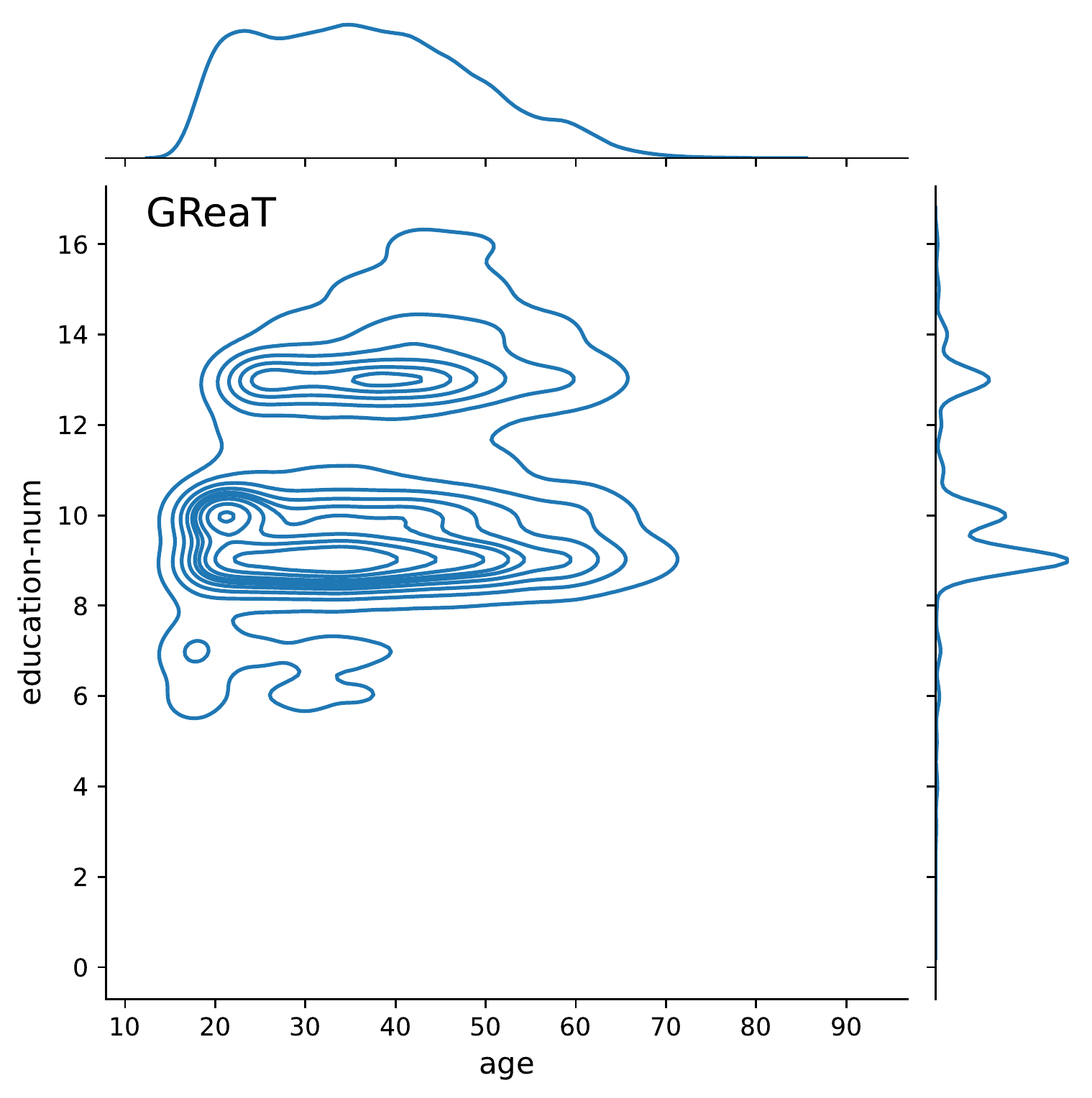}
    \includegraphics[width=\imgjointwidth\textwidth]{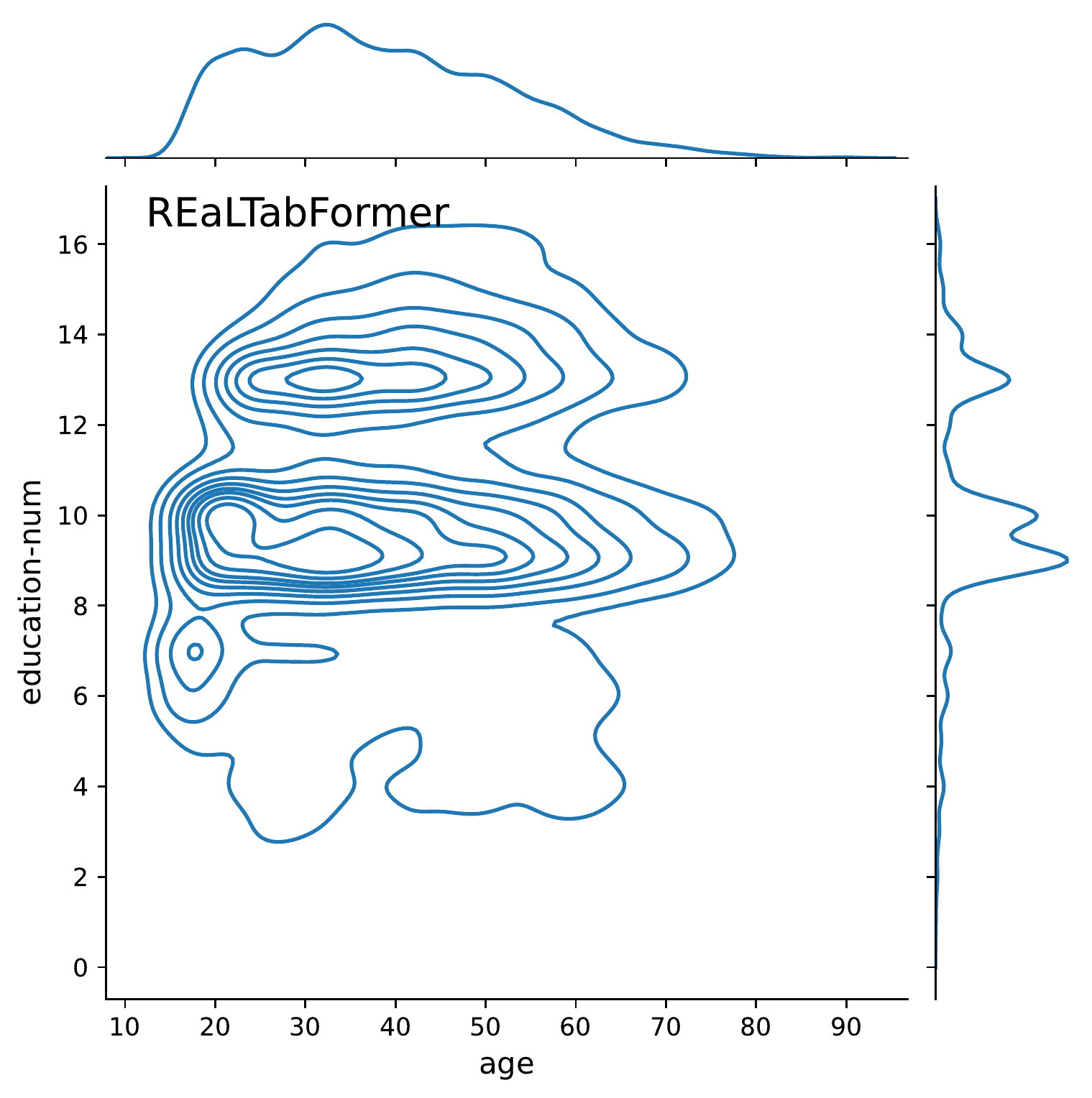}
    \caption{Joint plot of two numerical variables in the Adult data compared across the samples generated by the different models.}
    \label{fig:joint-adult}
\end{figure}

\begin{figure}[H]
    \centering
    \includegraphics[width=\imgjointwidth\textwidth]{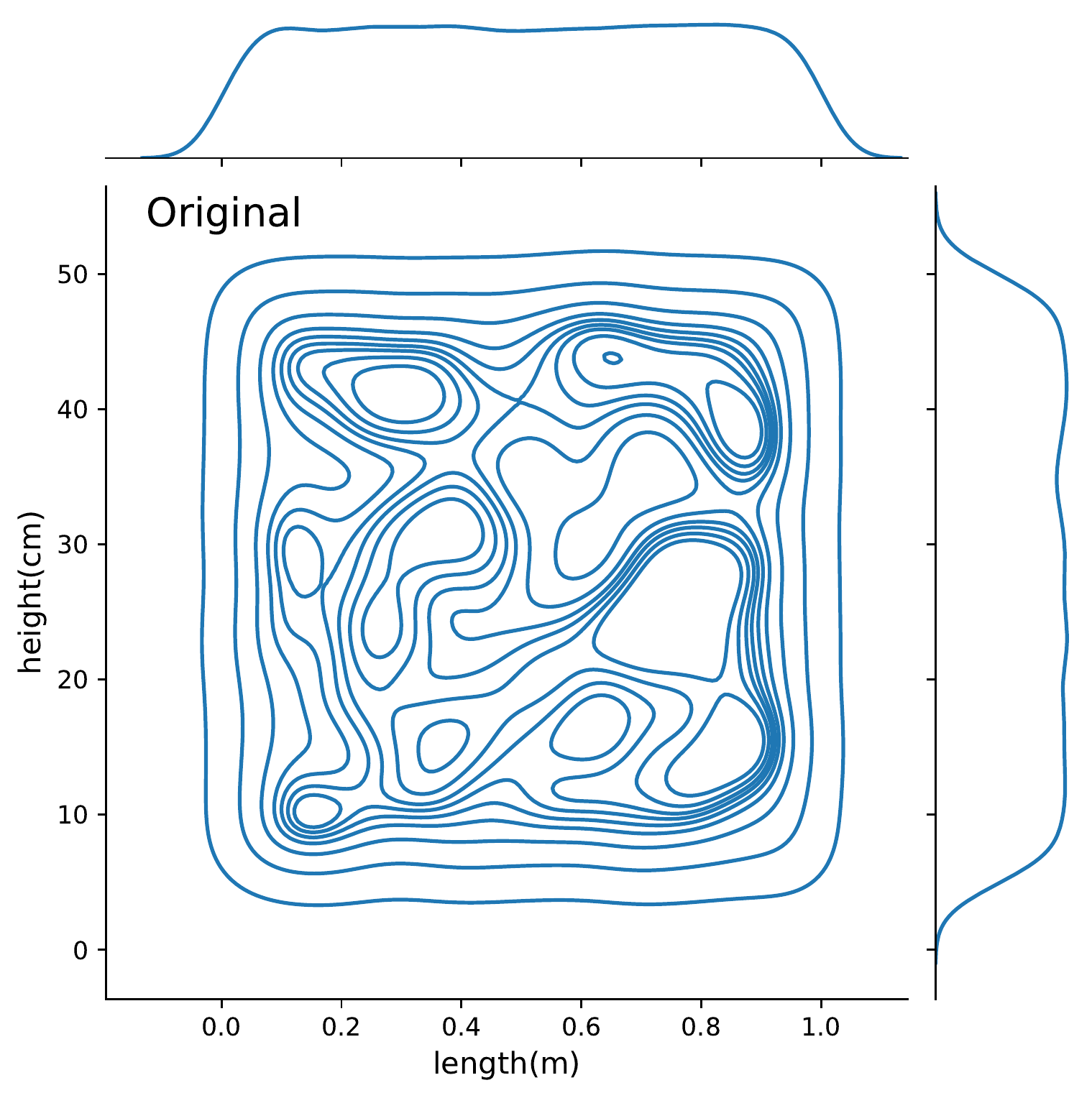}
    \includegraphics[width=\imgjointwidth\textwidth]{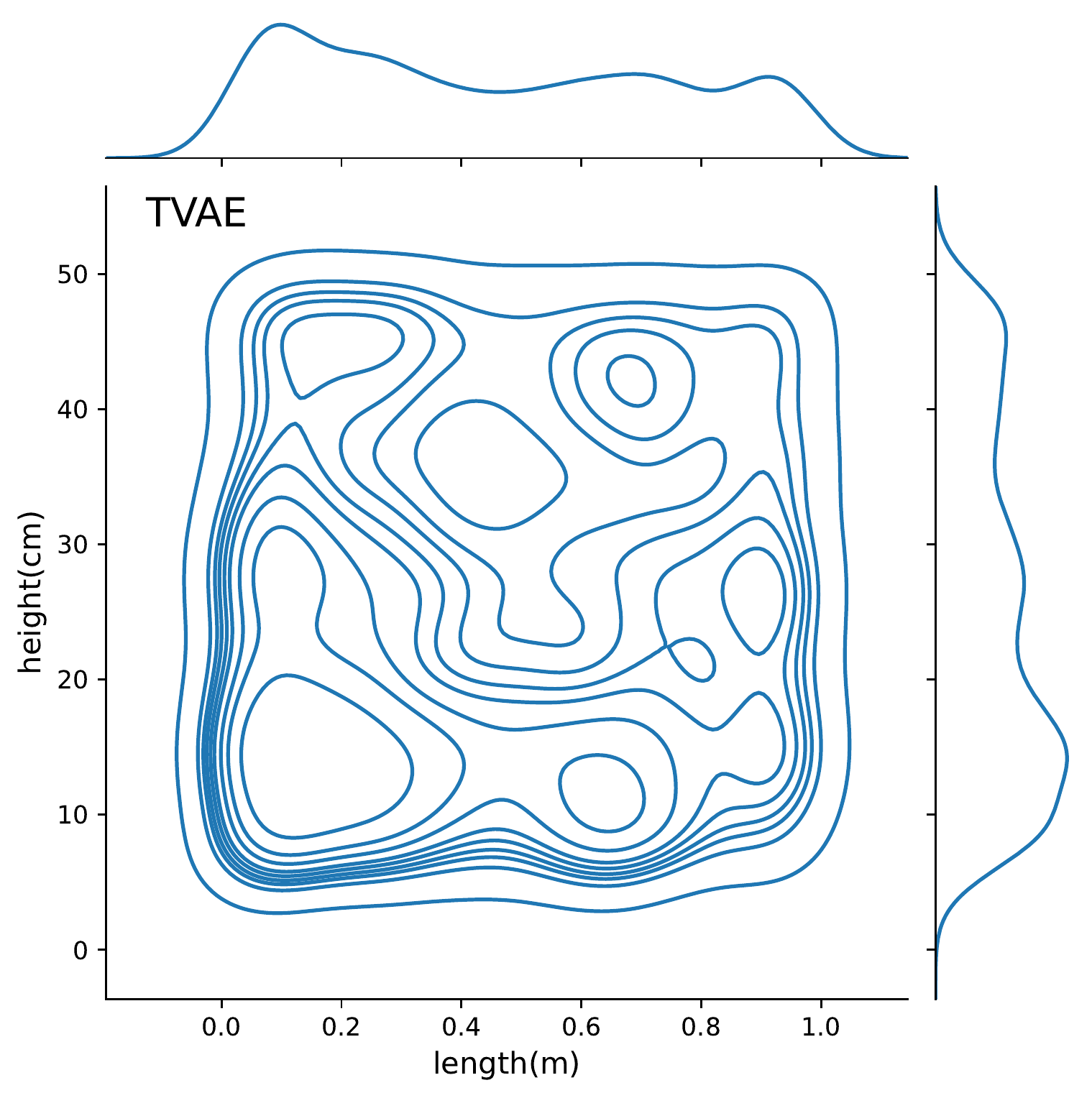}
    \includegraphics[width=\imgjointwidth\textwidth]{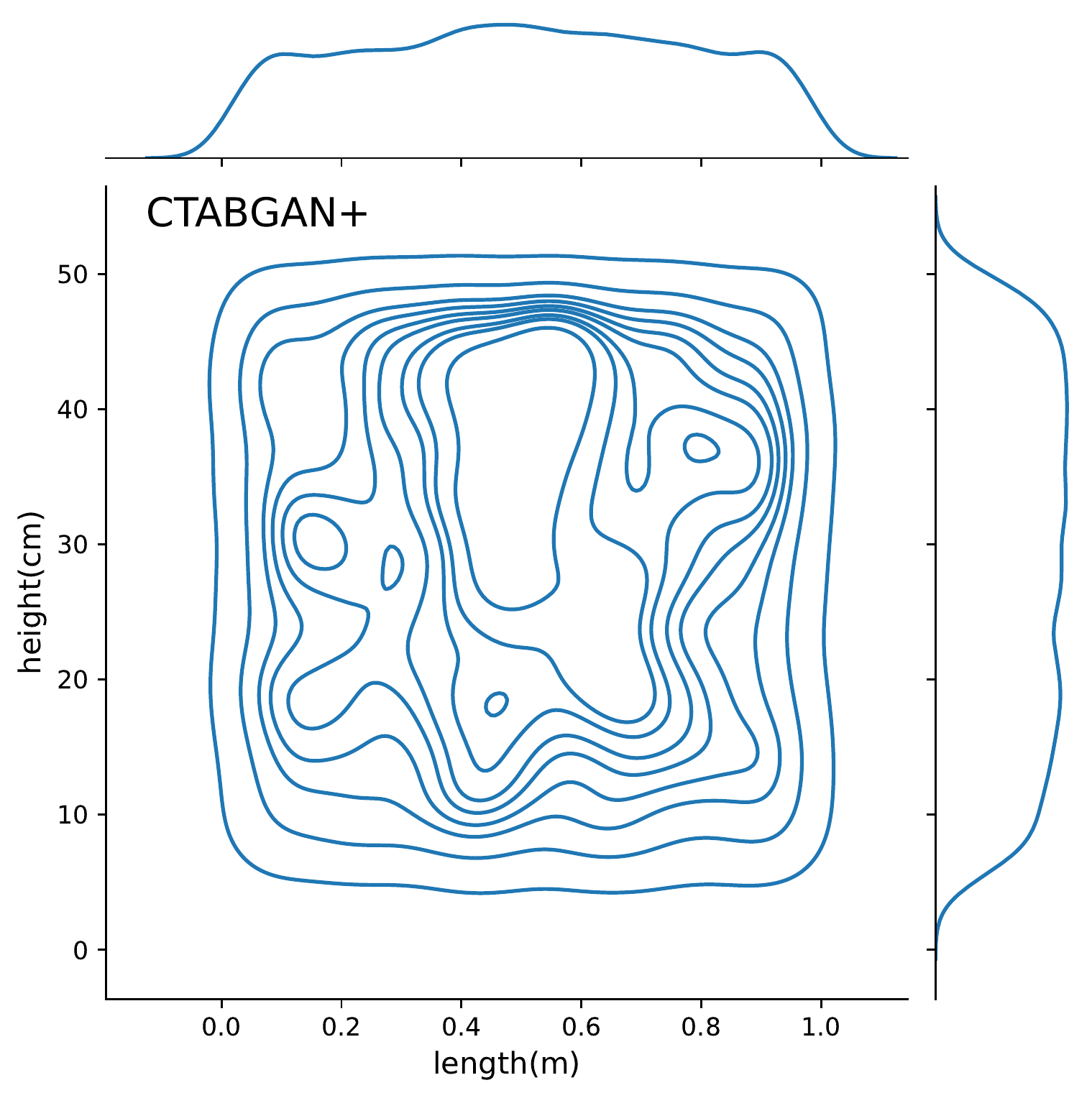}
    \includegraphics[width=\imgjointwidth\textwidth]{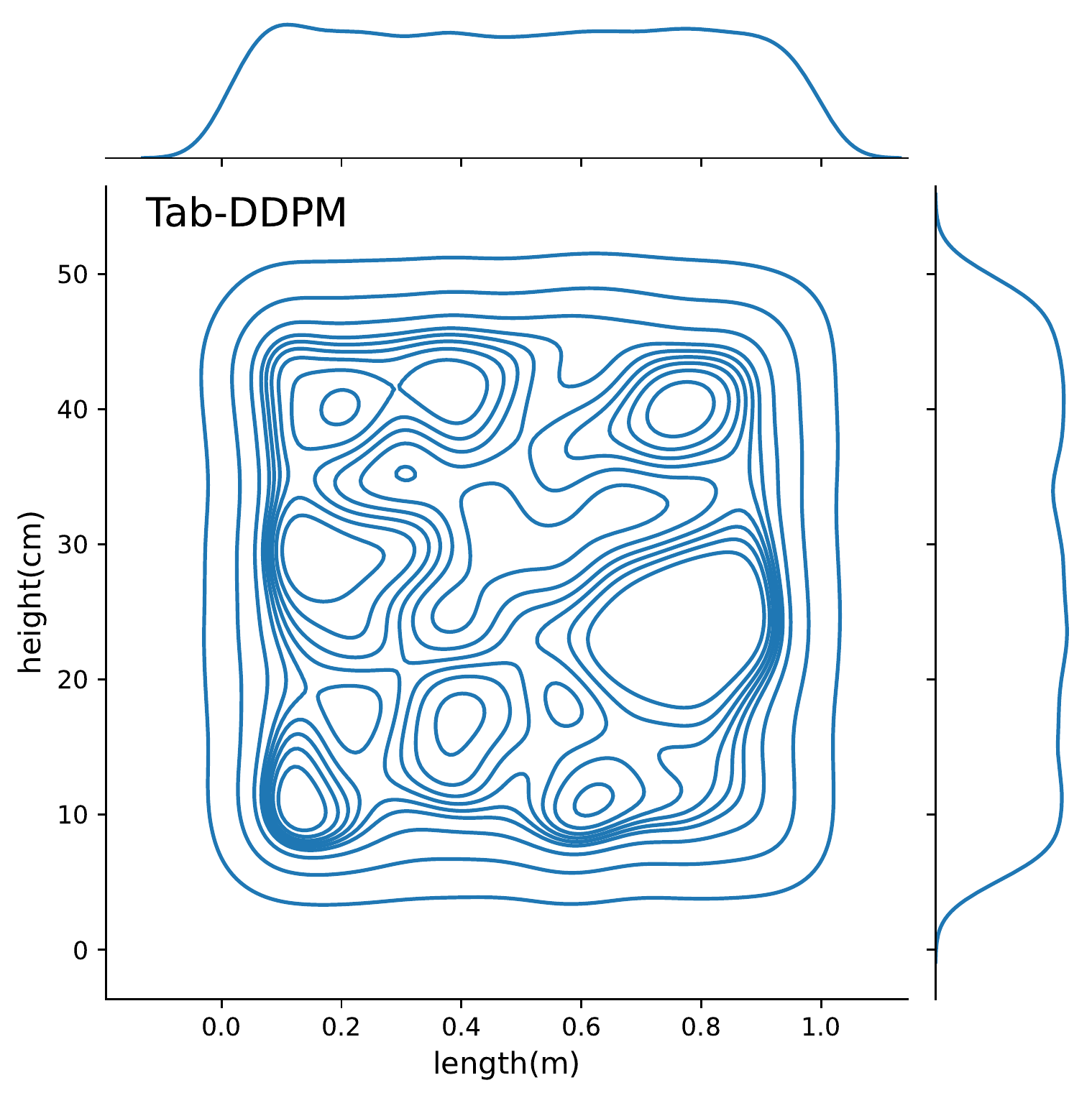}
    \includegraphics[width=\imgjointwidth\textwidth]{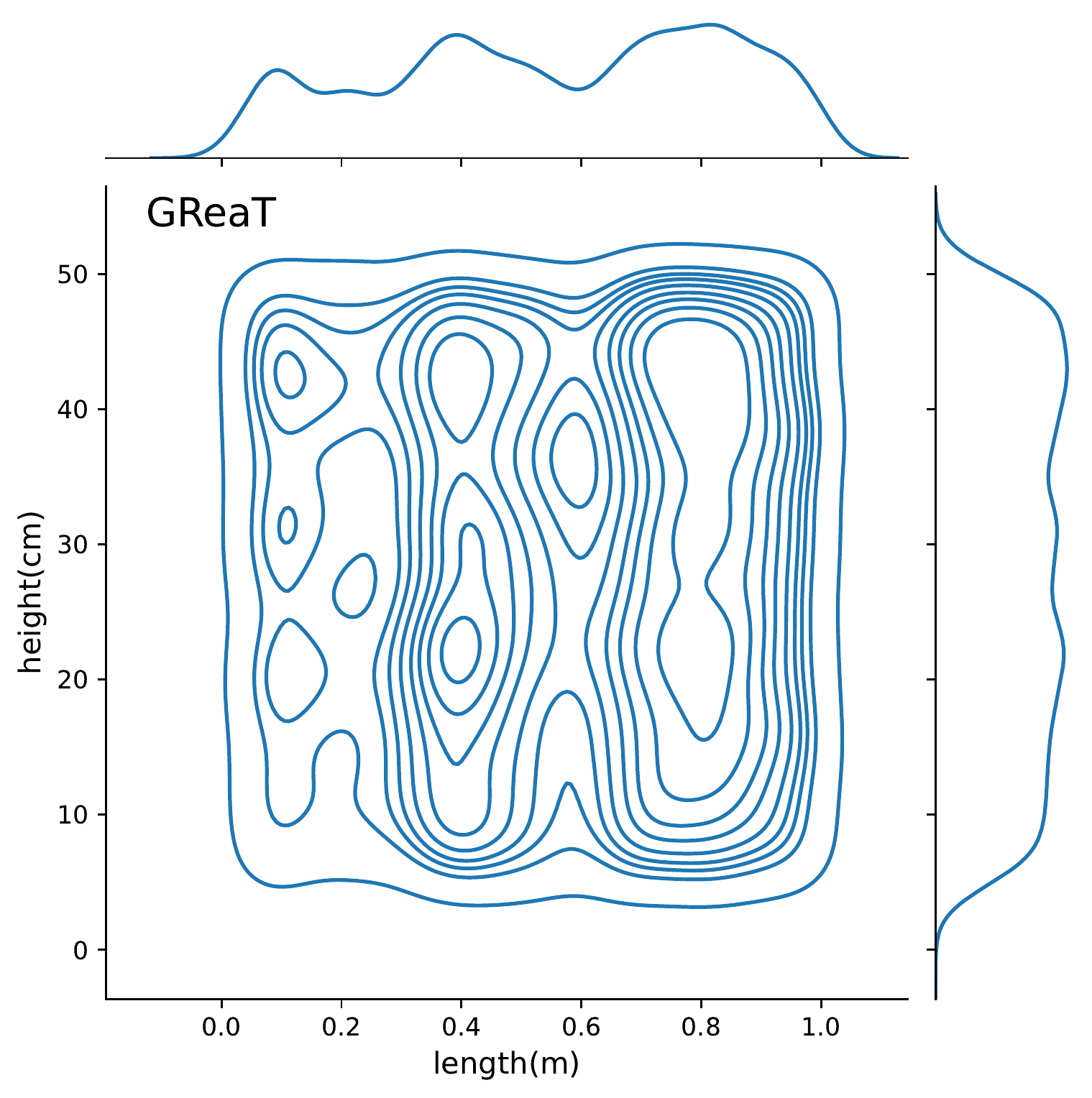}
    \includegraphics[width=\imgjointwidth\textwidth]{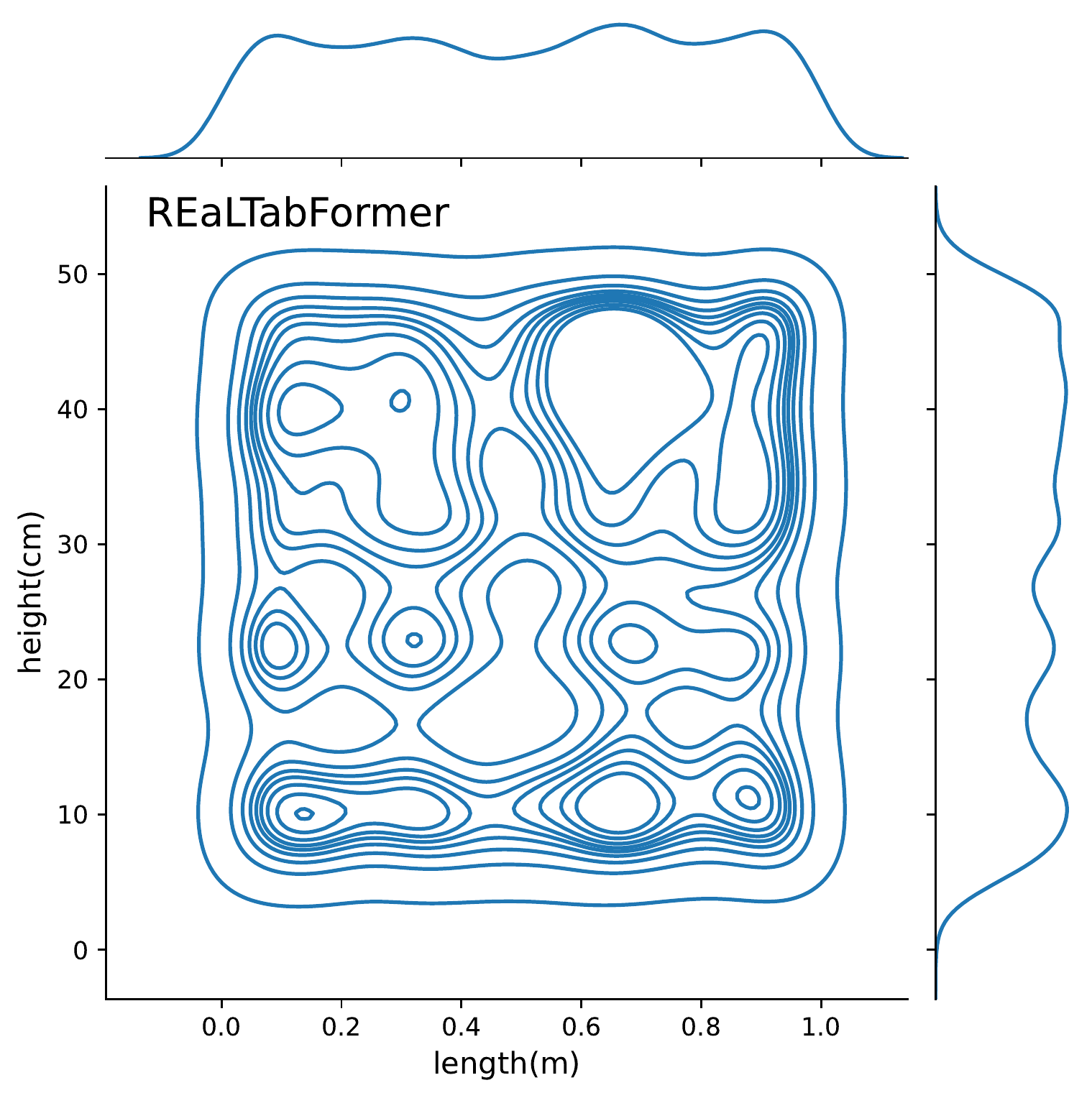}
    \caption{Joint plot of two numerical variables in the Buddy data compared across the samples generated by the different models.}
    \label{fig:joint-buddy}
\end{figure}

\begin{figure}[H]
    \centering
    \includegraphics[width=\imgjointwidth\textwidth]{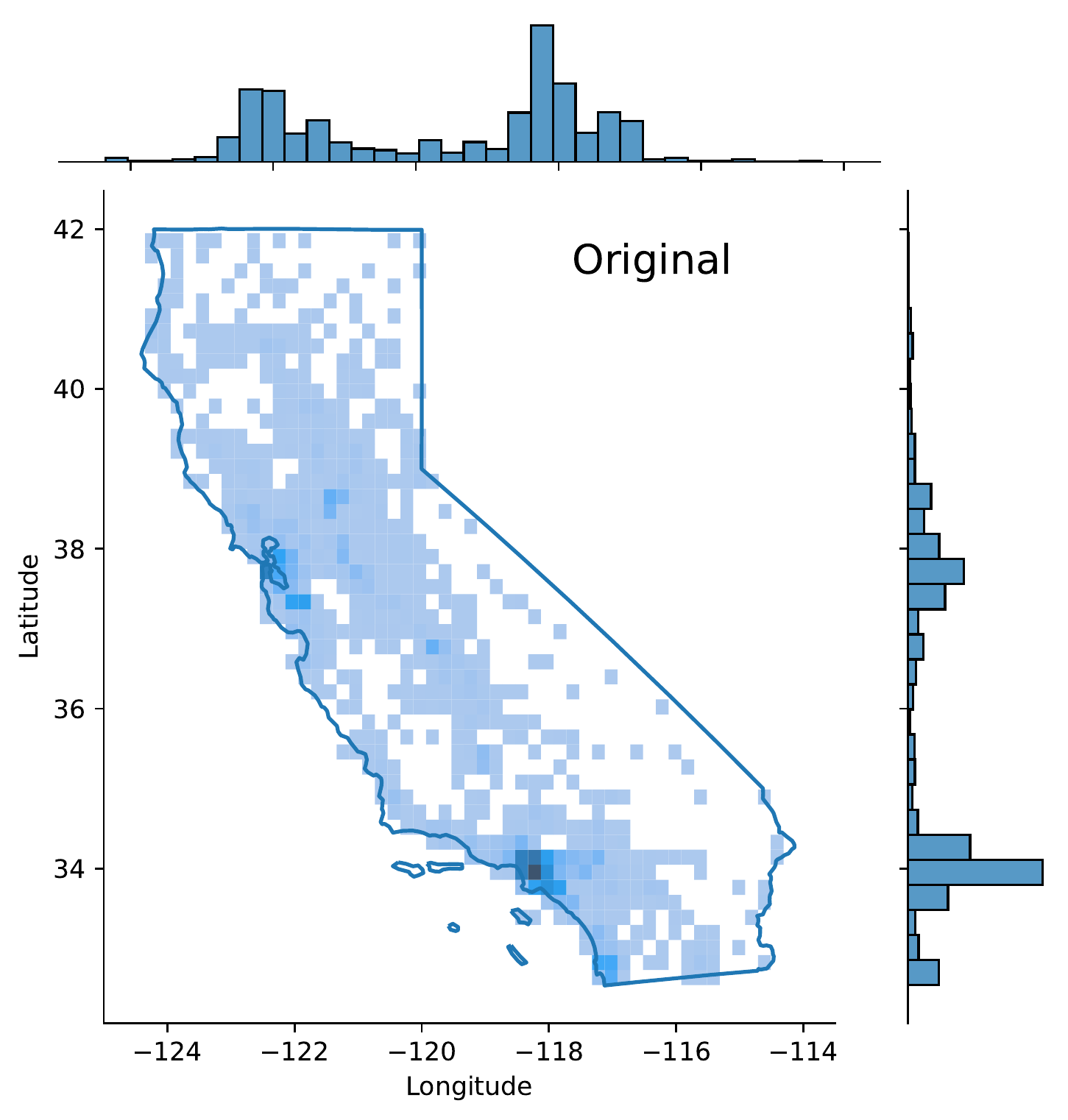}
    \includegraphics[width=\imgjointwidth\textwidth]{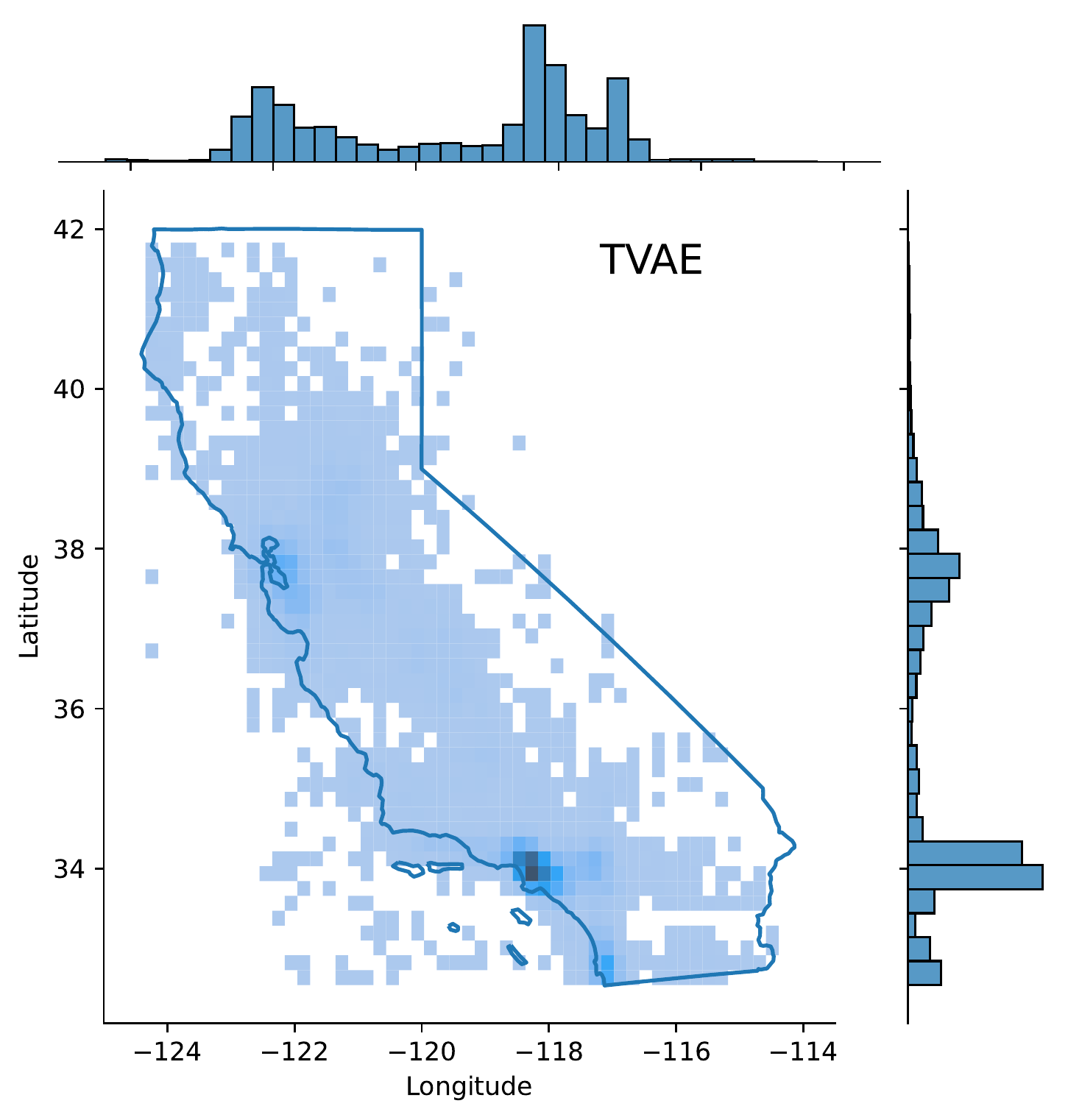}
    \includegraphics[width=\imgjointwidth\textwidth]{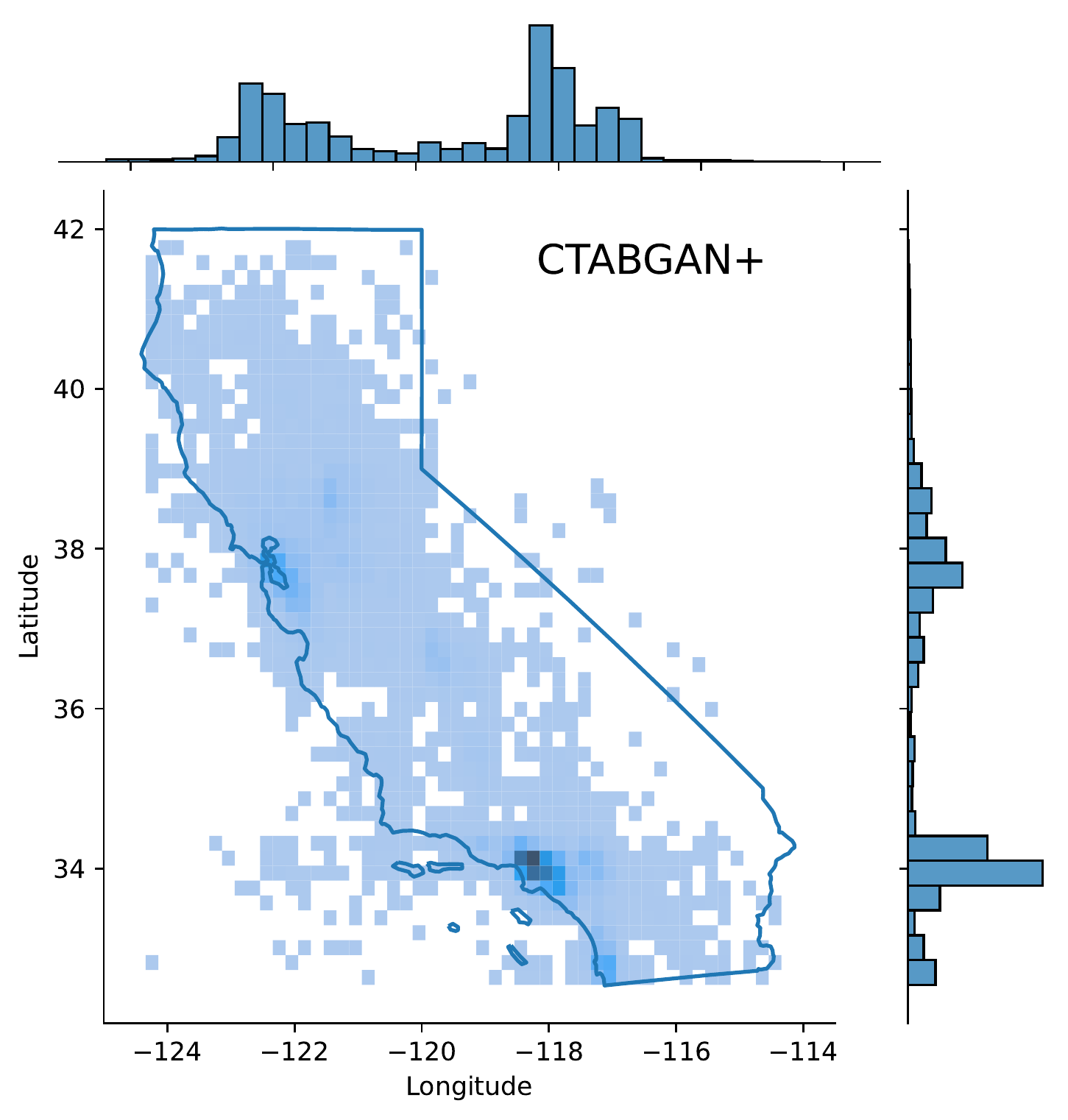}
    \includegraphics[width=\imgjointwidth\textwidth]{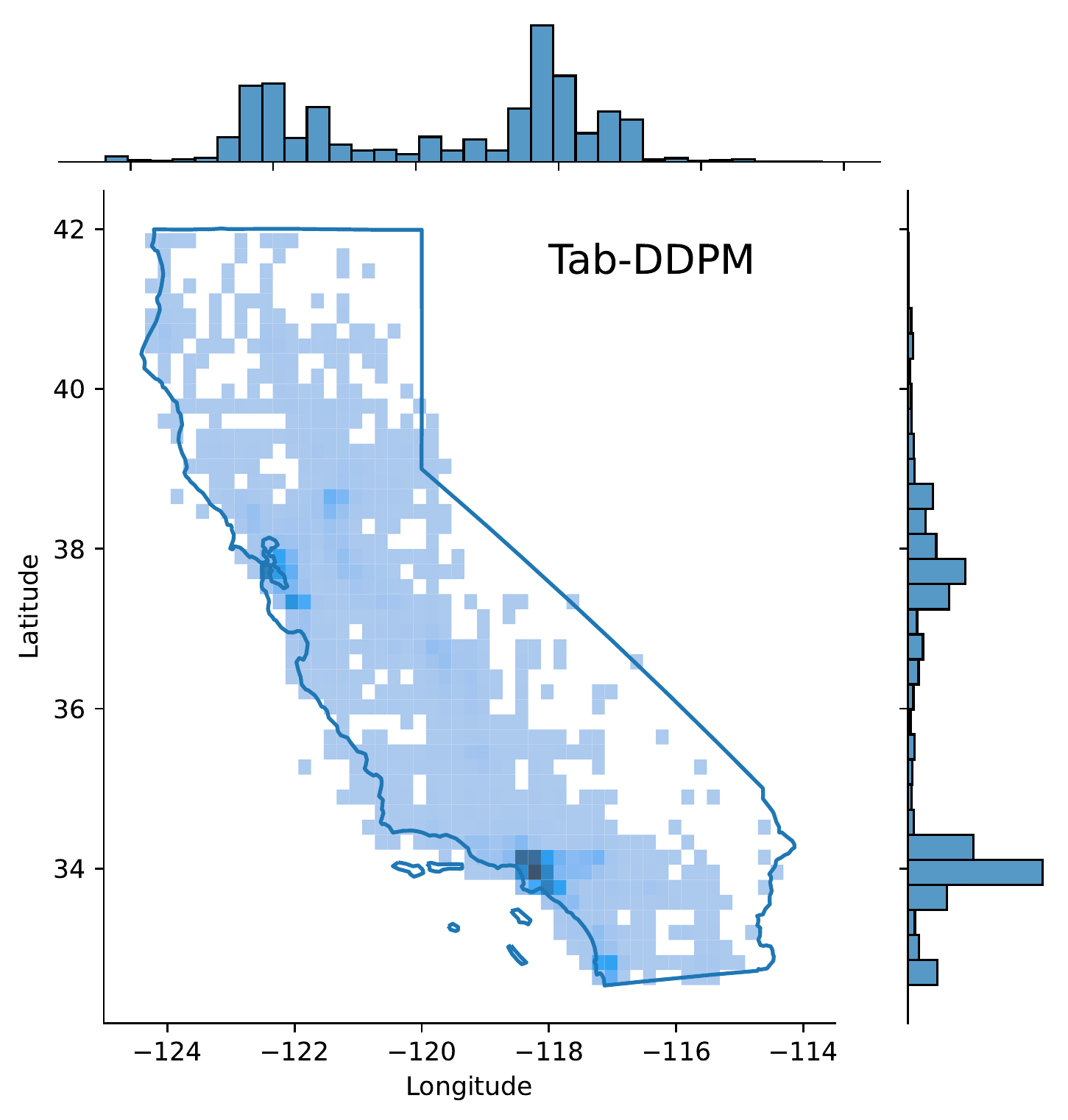}
    \includegraphics[width=\imgjointwidth\textwidth]{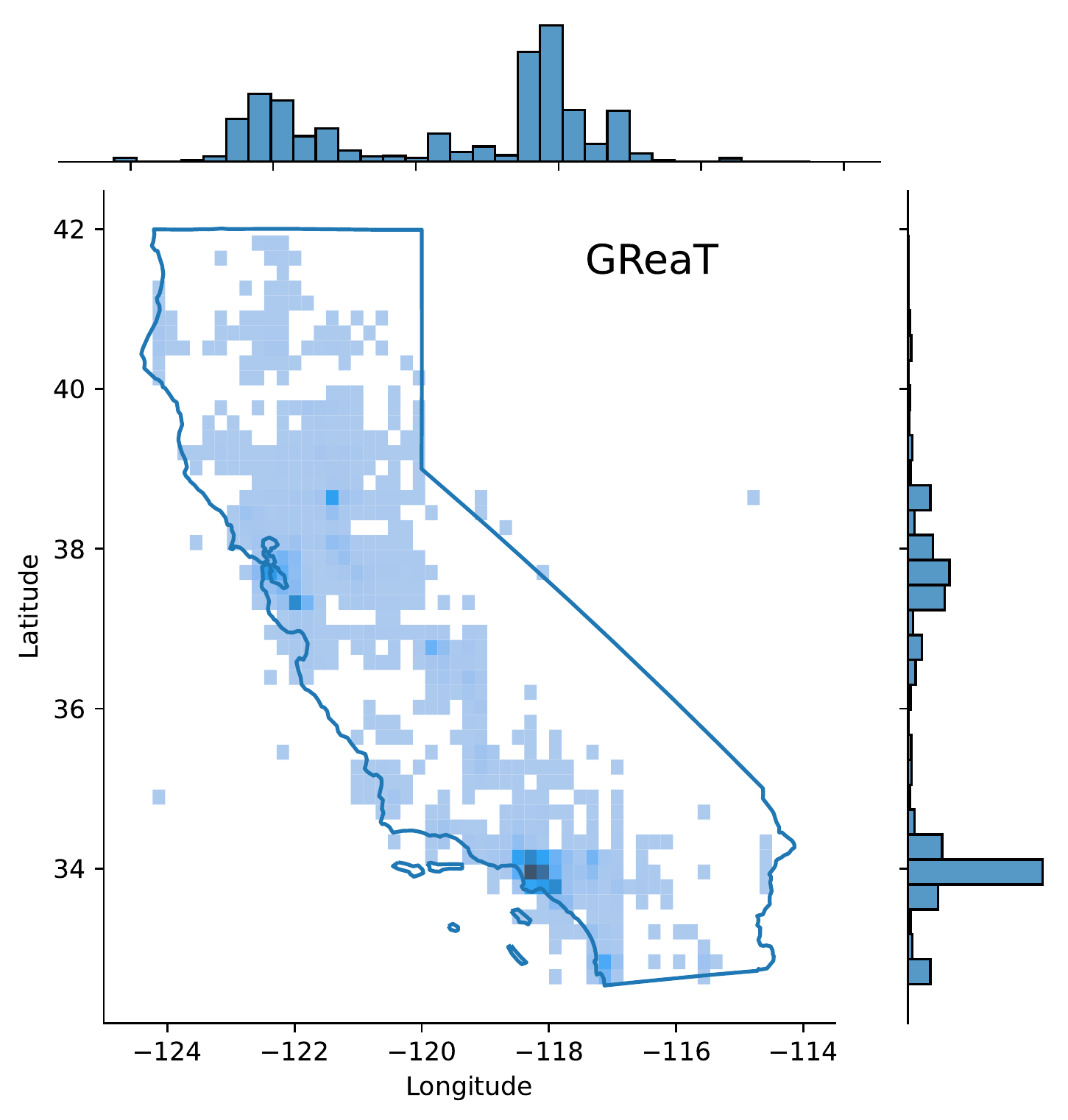}
    \includegraphics[width=\imgjointwidth\textwidth]{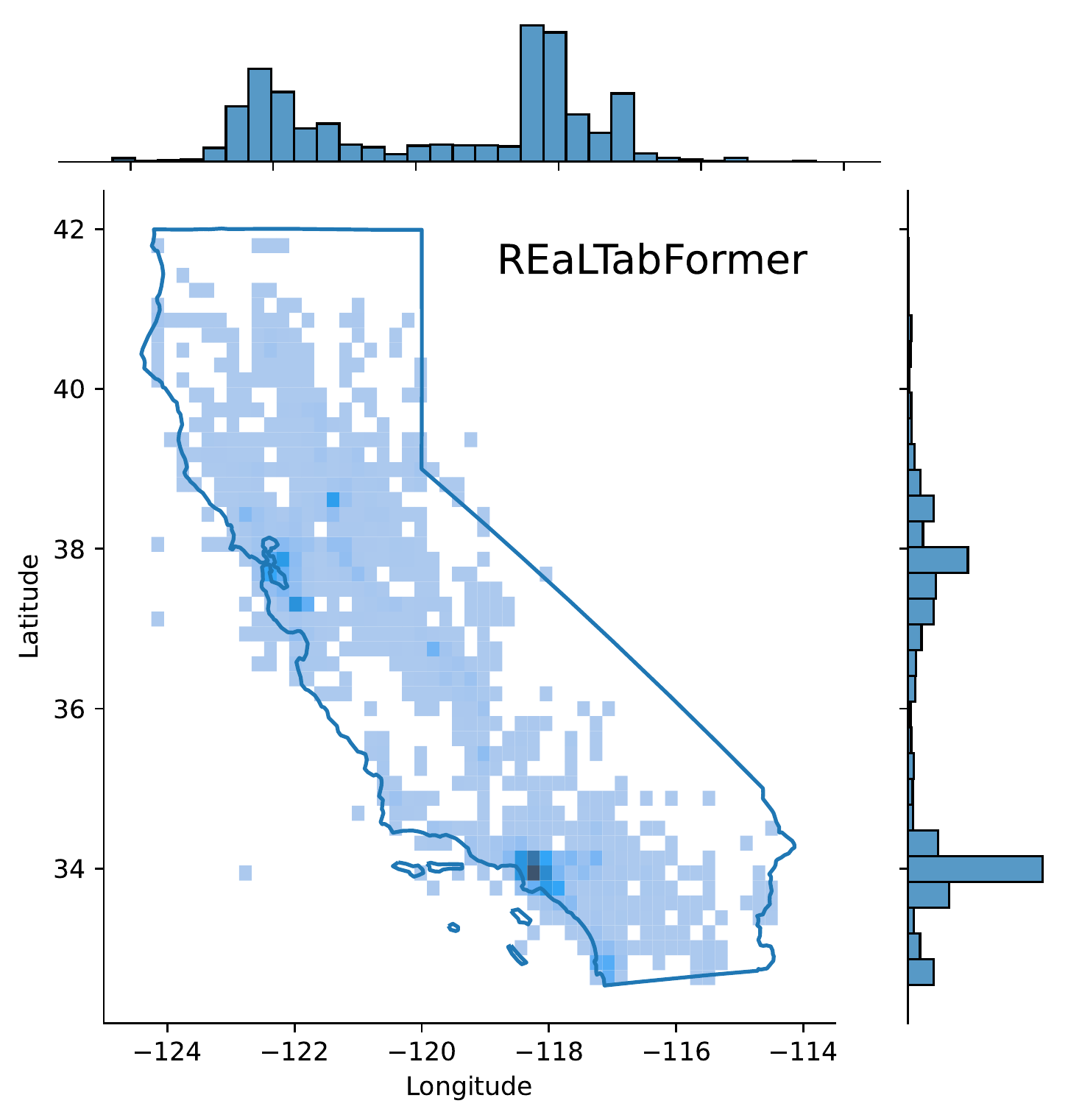}
    \caption{Joint plot of two numerical variables in the California housing data compared across the samples generated by the different models.}
    \label{fig:joint-california}
\end{figure}

\begin{figure}[H]
    \centering
    \includegraphics[width=\imgjointwidth\textwidth]{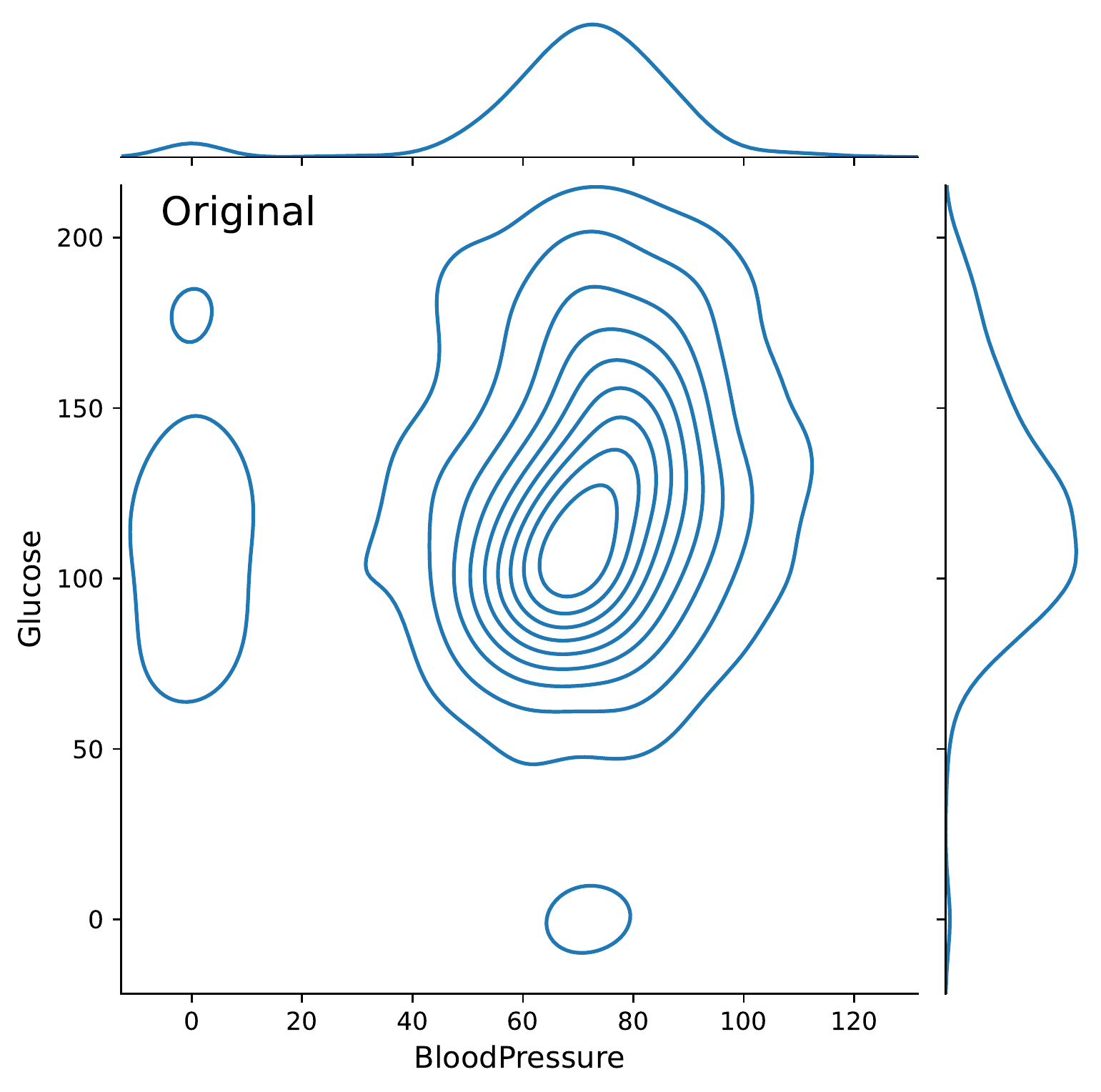}
    \includegraphics[width=\imgjointwidth\textwidth]{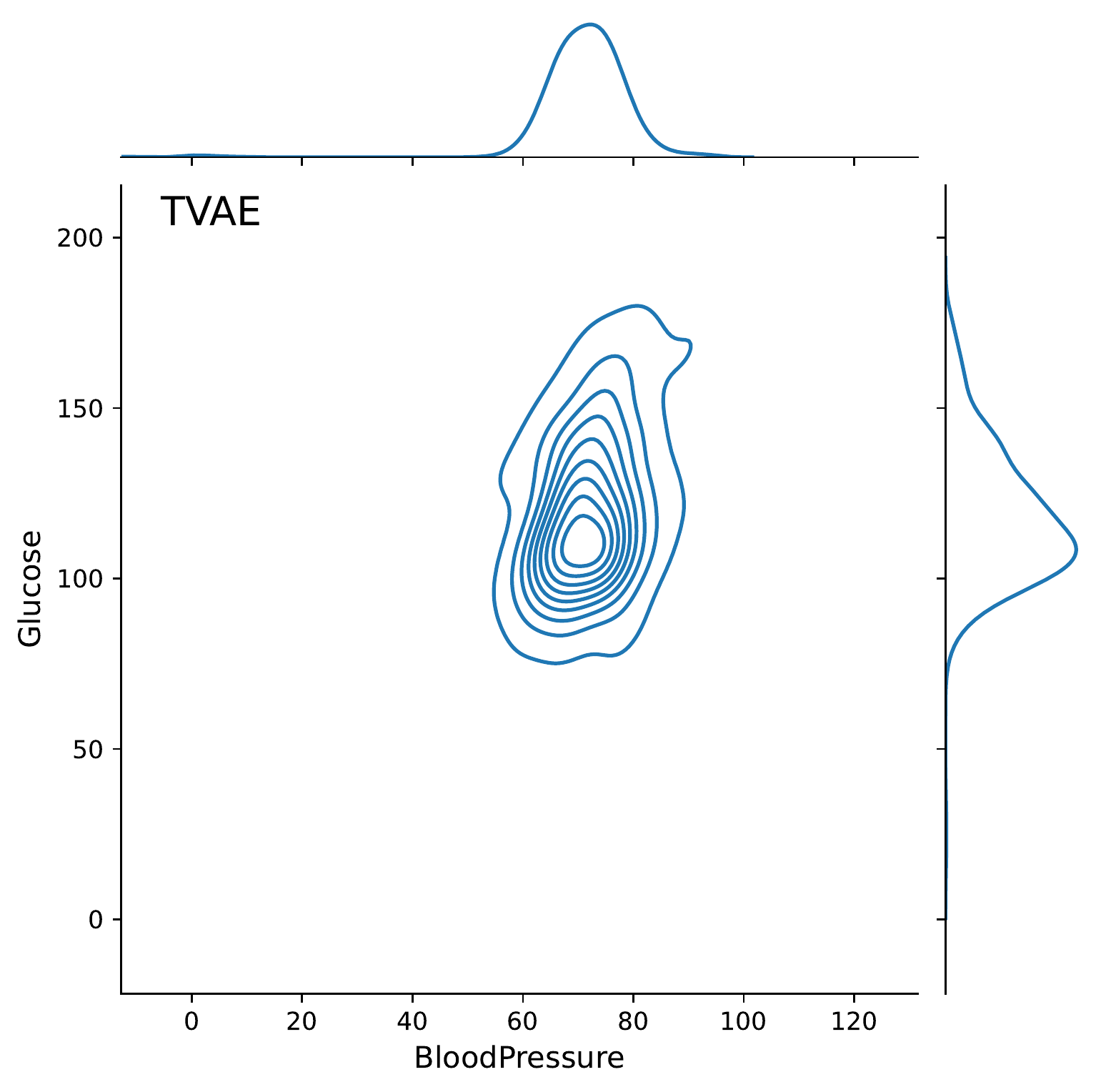}
    \includegraphics[width=\imgjointwidth\textwidth]{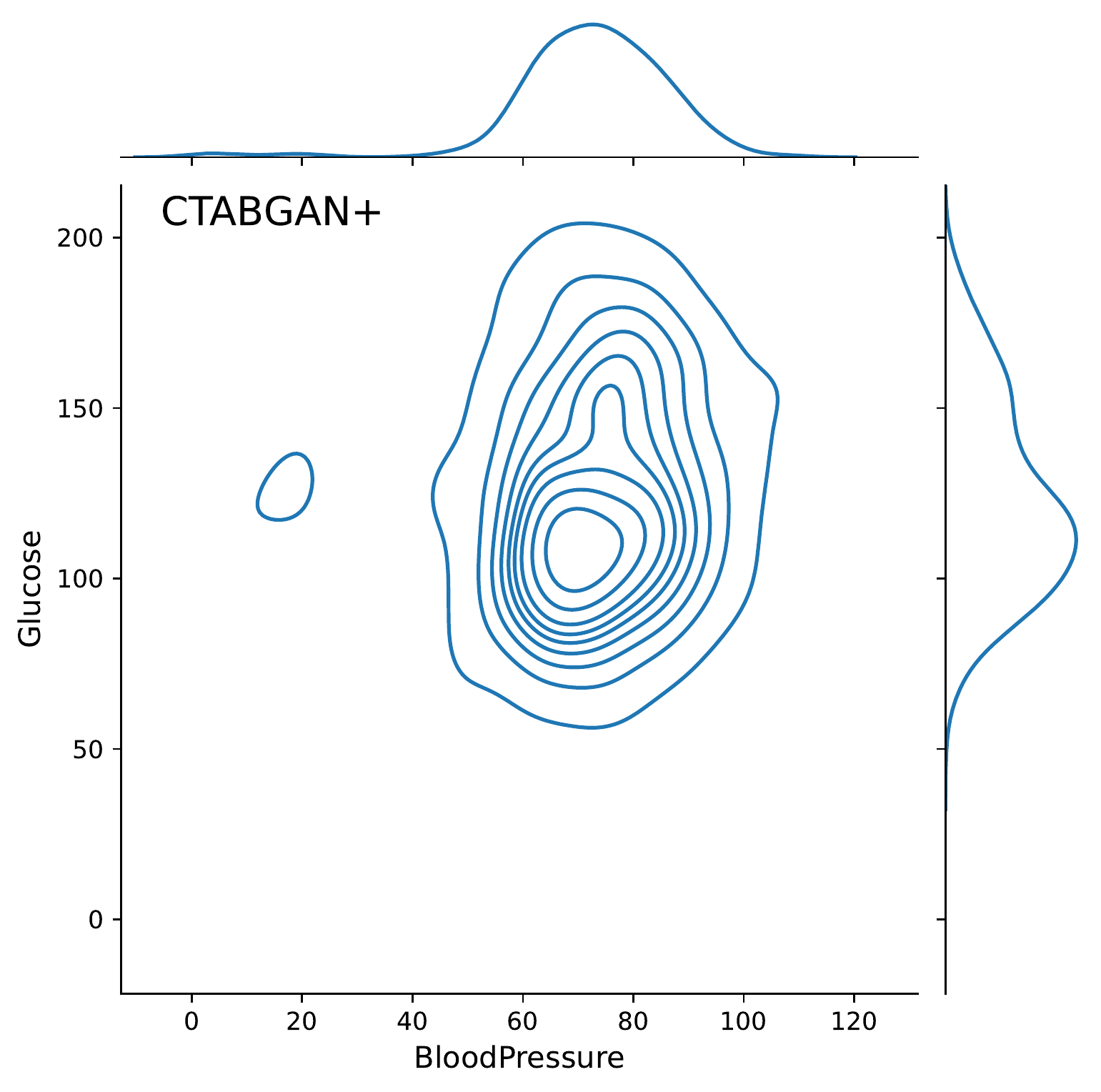}
    \includegraphics[width=\imgjointwidth\textwidth]{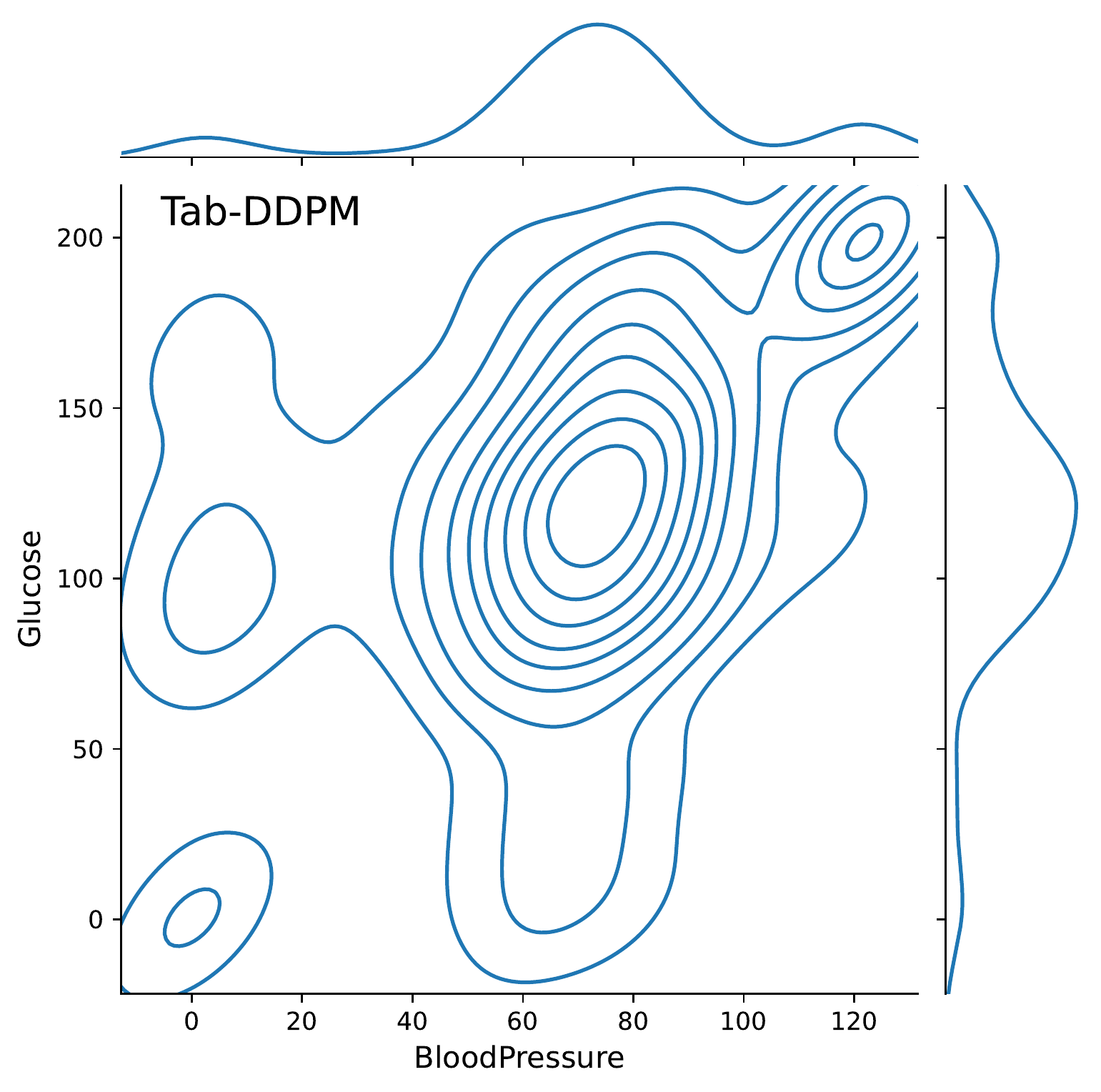}
    \includegraphics[width=\imgjointwidth\textwidth]{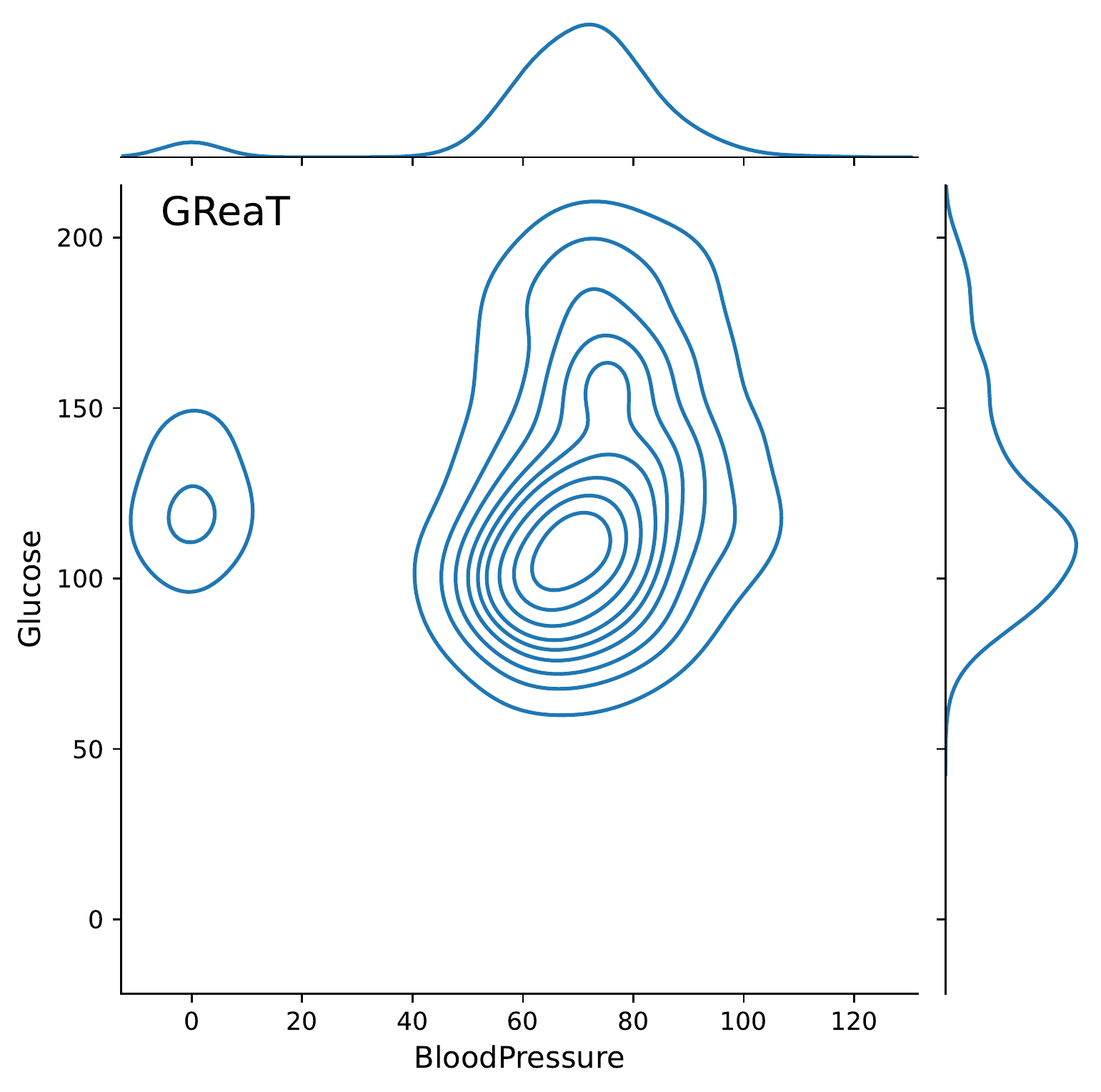}
    \includegraphics[width=\imgjointwidth\textwidth]{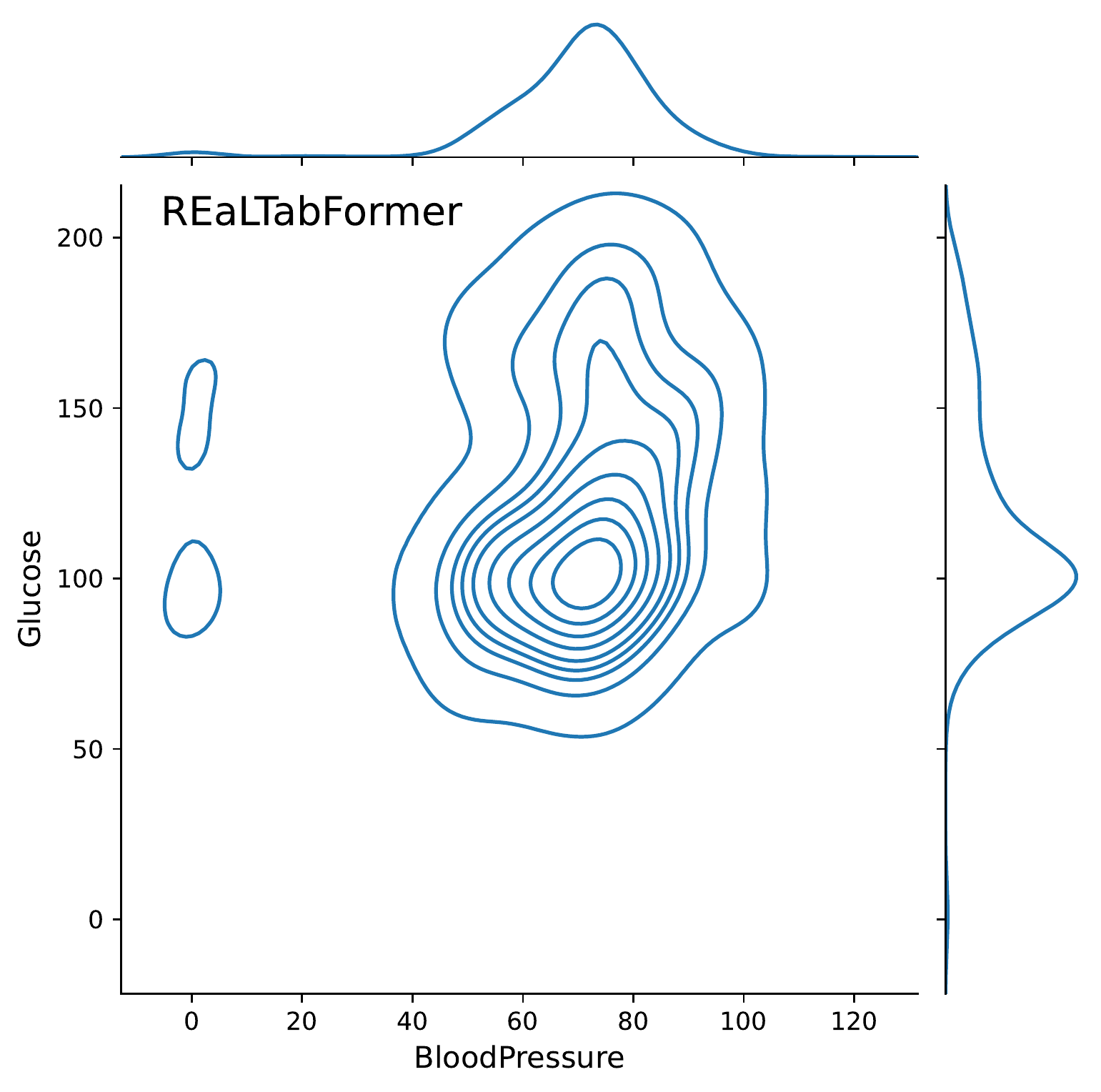}
    \caption{Joint plot of two numerical variables in the Diabetes data compared across the samples generated by the different models.}
    \label{fig:joint-diabetes}
\end{figure}

\subsection{Distance to closest record (DCR) distribution}

We present earlier the distance to closest record (DCR) distribution, Equation \ref{eq:dcr}, as part of our proposed strategy to detect overfitting in training the REaLTabFormer model. Here, we use the DCR distribution to visually assess whether the generative models create exact copies of observations from the training data. We also show the DCR distribution of the real test data as a reference.

\newcommand\imgwidth{0.16}

\begin{figure}[H]
    \centering
    \includegraphics[width=\imgwidth\textwidth]{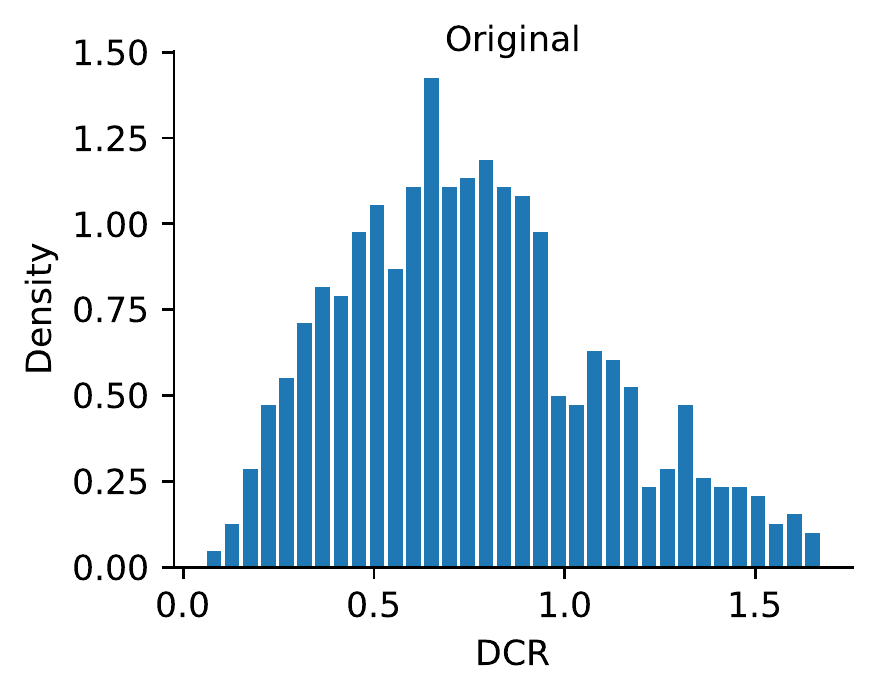}
    \includegraphics[width=\imgwidth\textwidth]{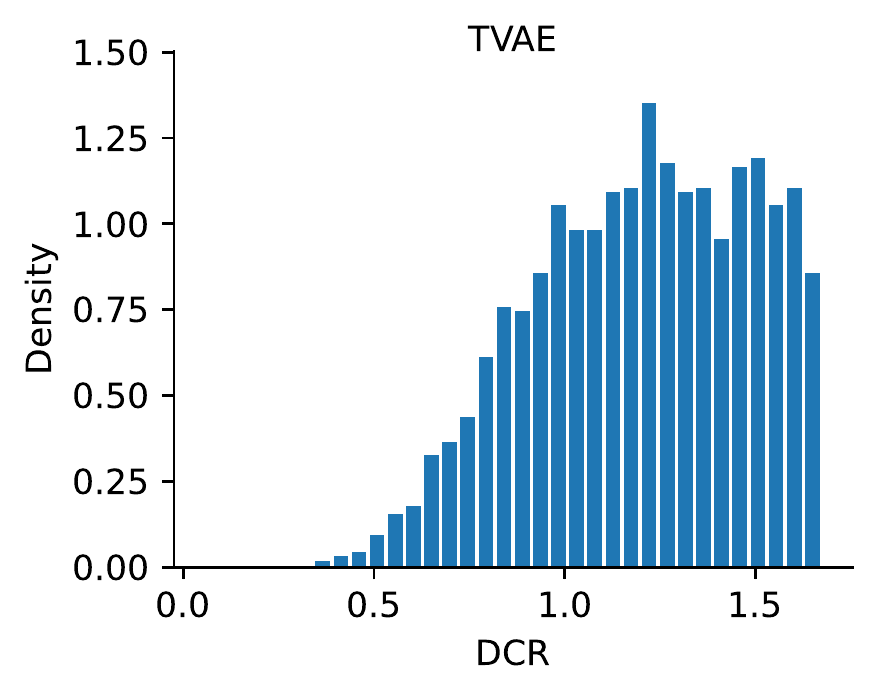}
    \includegraphics[width=\imgwidth\textwidth]{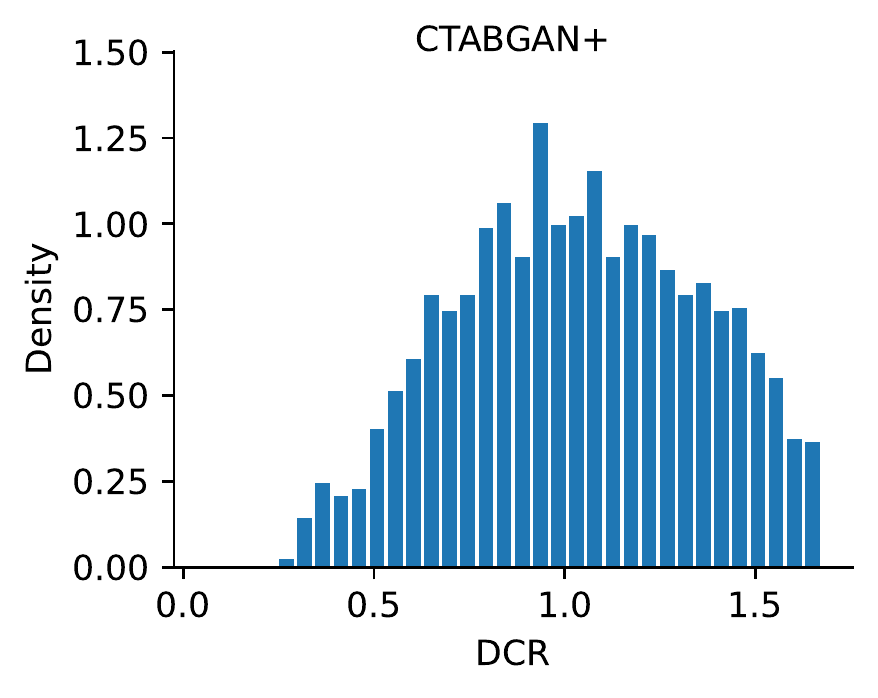}
    \includegraphics[width=\imgwidth\textwidth]{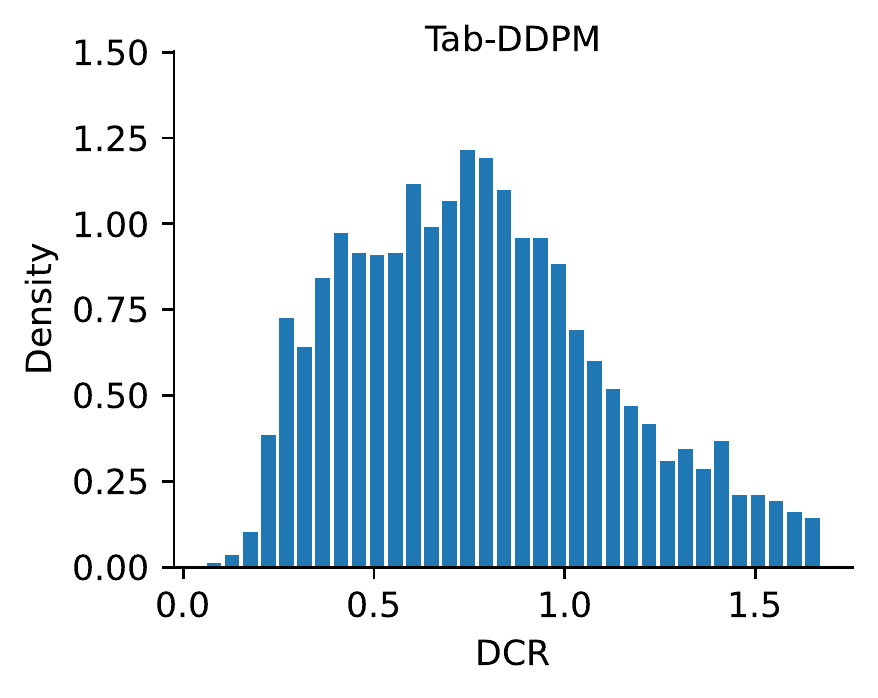}
    \includegraphics[width=\imgwidth\textwidth]{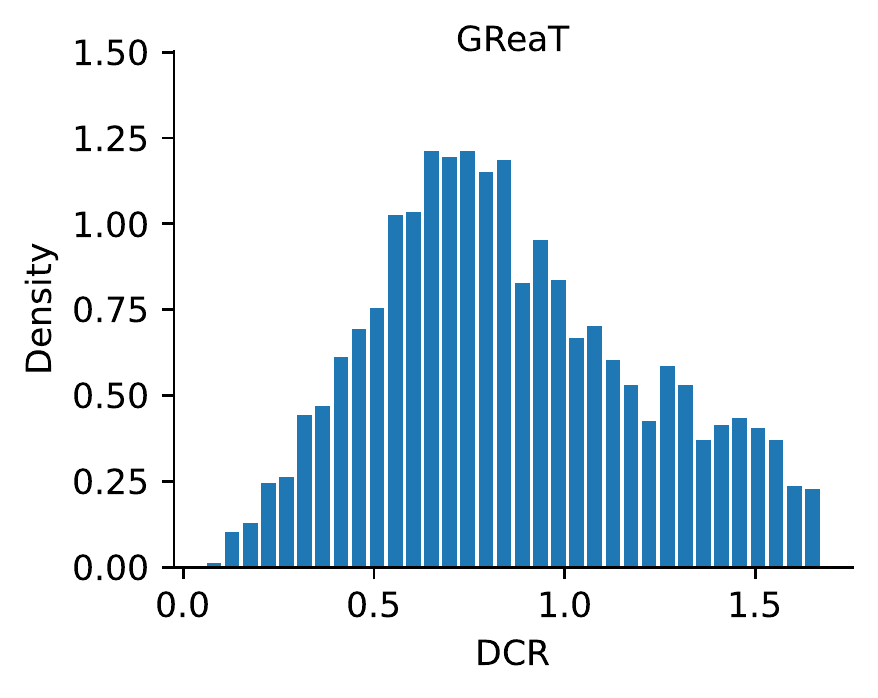}
    \includegraphics[width=\imgwidth\textwidth]{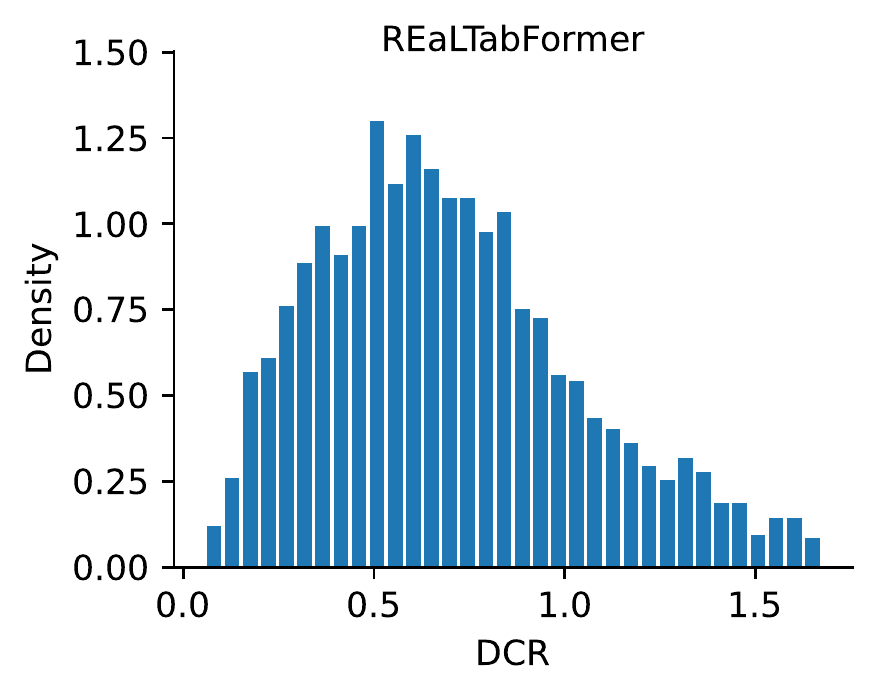}
    \caption{Distance to closest record (DCR) distributions of the different models for the Abalone data.}
    \label{fig:dcr-abalone}
\end{figure}

\begin{figure}[H]
    \centering
    \includegraphics[width=\imgwidth\textwidth]{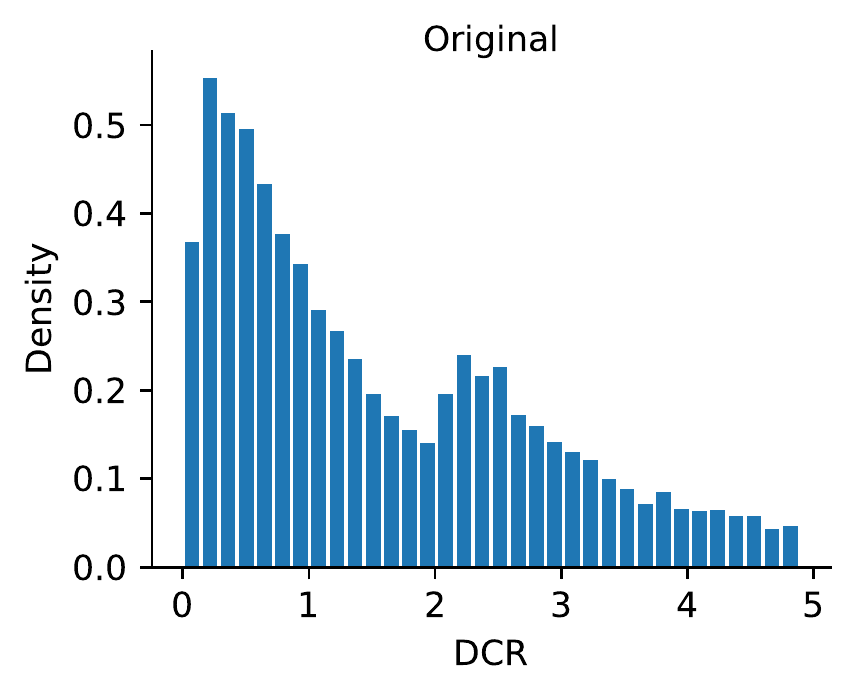}
    \includegraphics[width=\imgwidth\textwidth]{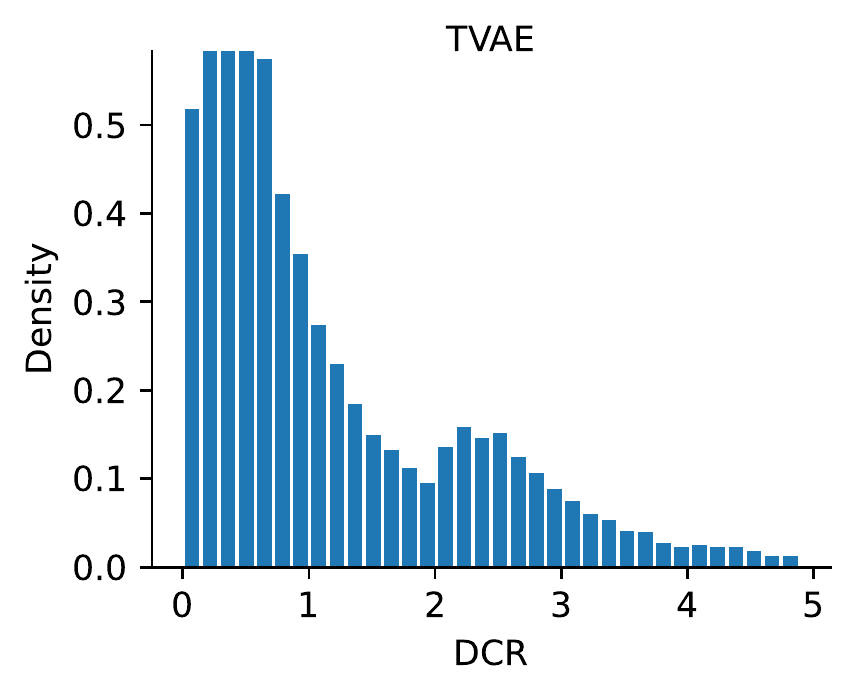}
    \includegraphics[width=\imgwidth\textwidth]{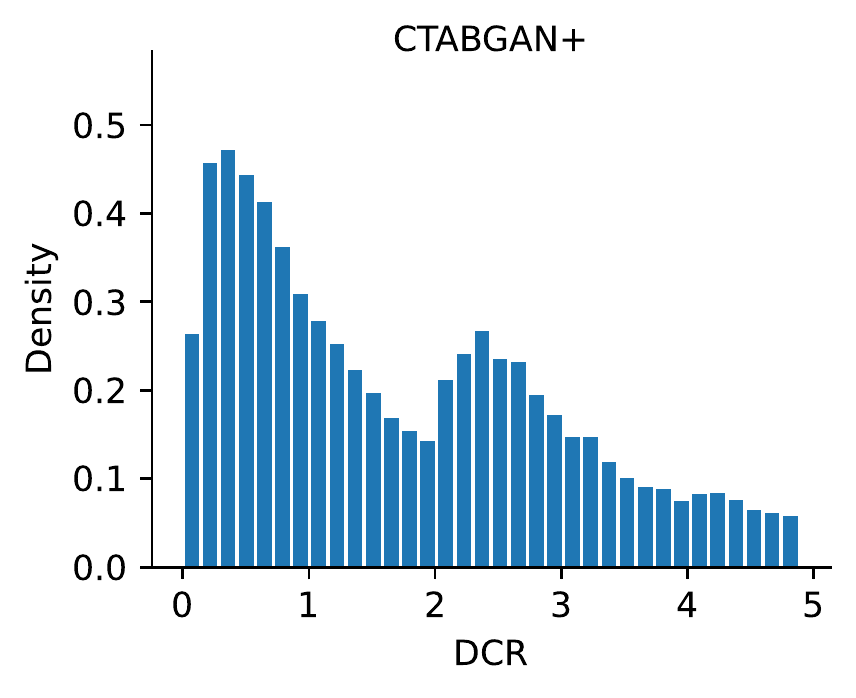}
    \includegraphics[width=\imgwidth\textwidth]{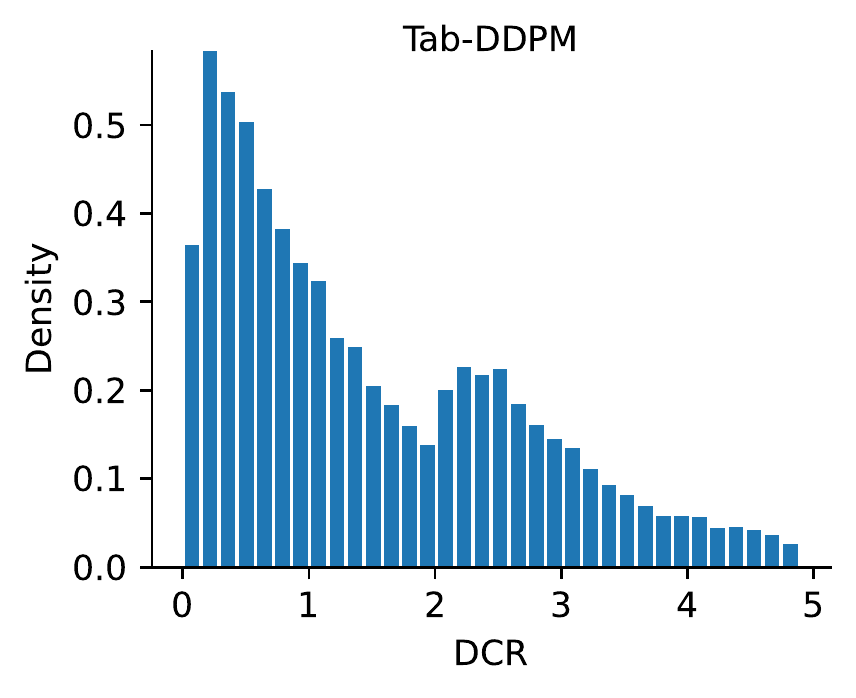}
    \includegraphics[width=\imgwidth\textwidth]{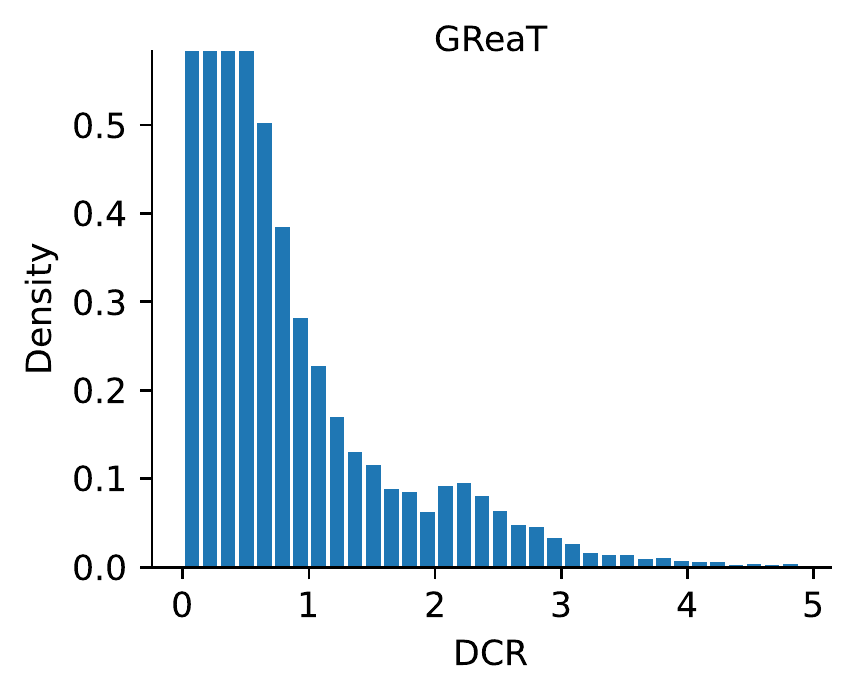}
    \includegraphics[width=\imgwidth\textwidth]{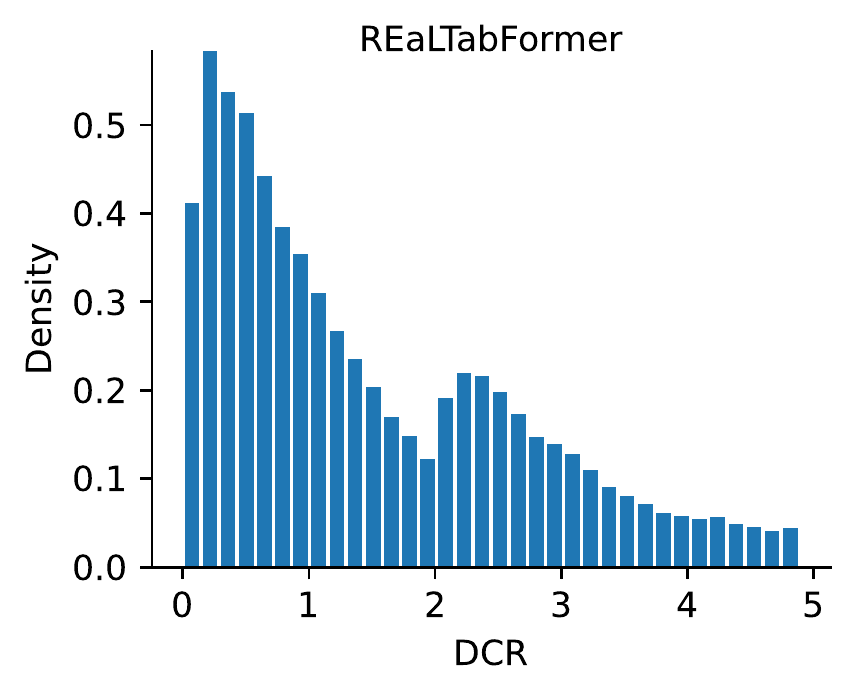}
    \caption{Distance to closest record (DCR) distributions of the different models for the Adult data.}
    \label{fig:dcr-adult}
\end{figure}

\begin{figure}[H]
    \centering
    \includegraphics[width=\imgwidth\textwidth]{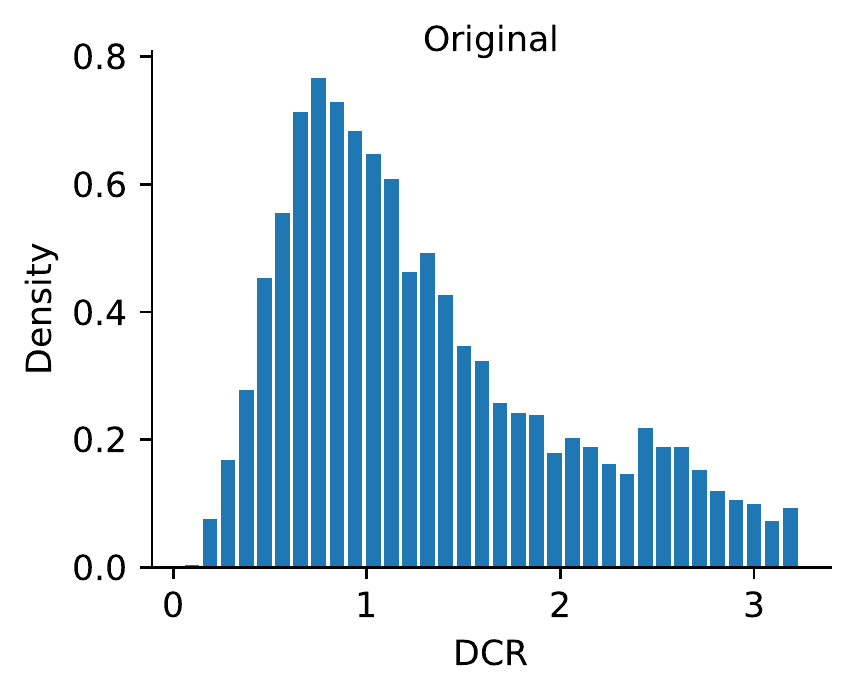}
    \includegraphics[width=\imgwidth\textwidth]{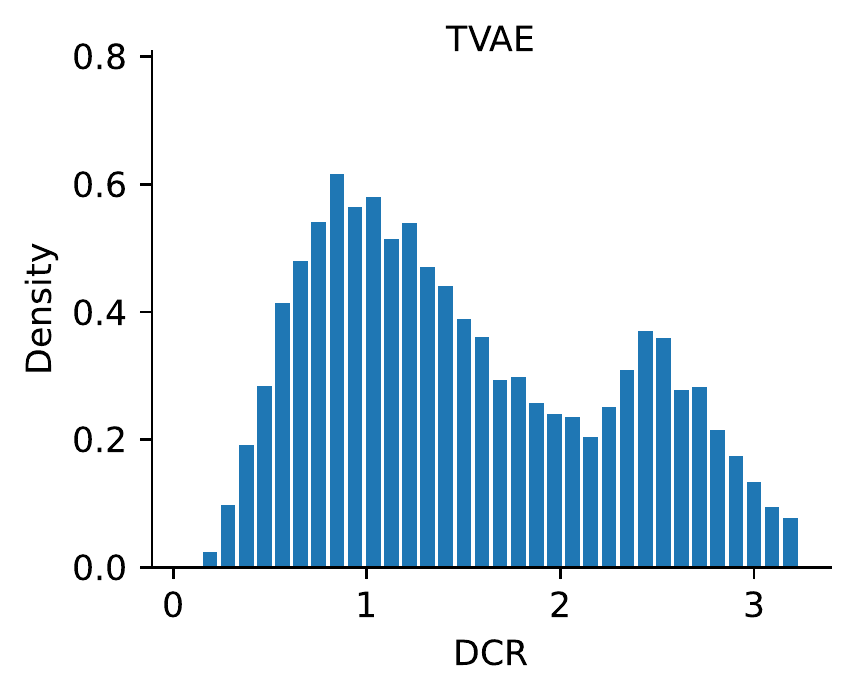}
    \includegraphics[width=\imgwidth\textwidth]{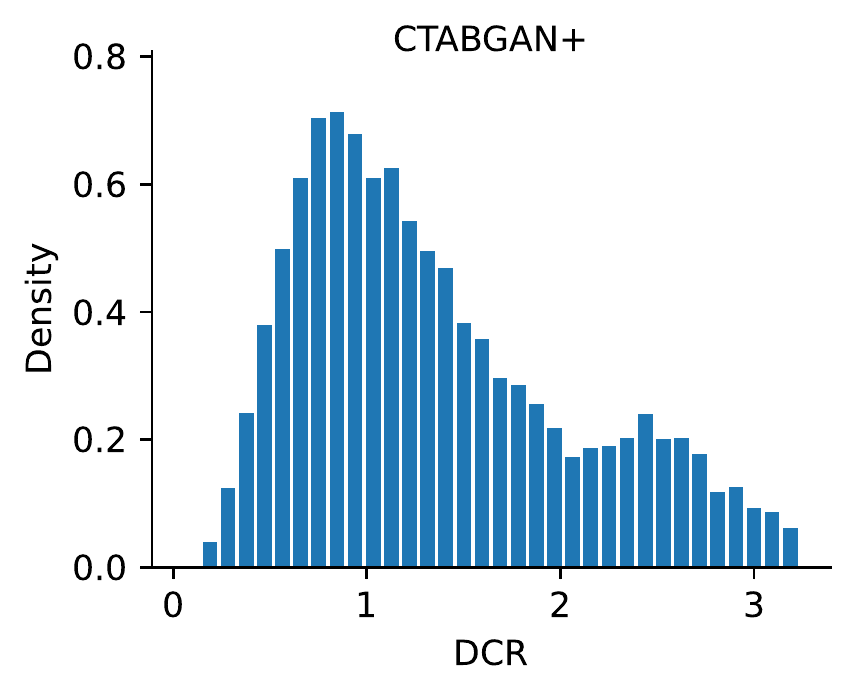}
    \includegraphics[width=\imgwidth\textwidth]{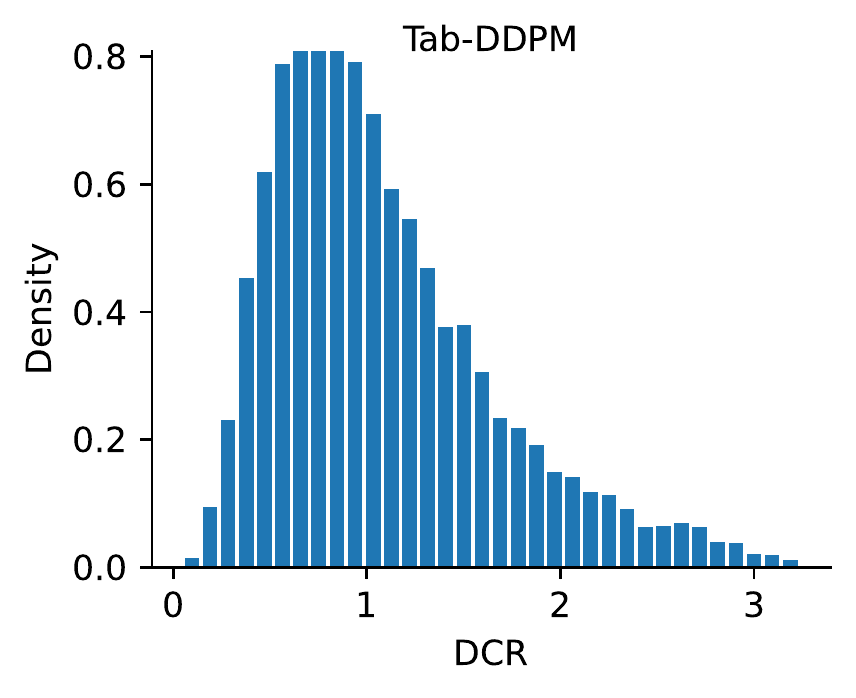}
    \includegraphics[width=\imgwidth\textwidth]{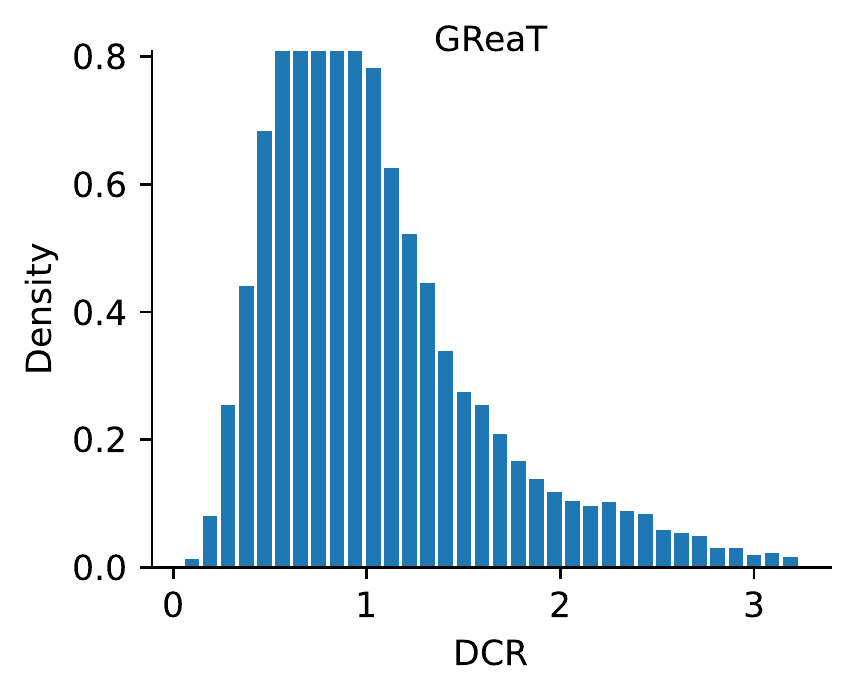}
    \includegraphics[width=\imgwidth\textwidth]{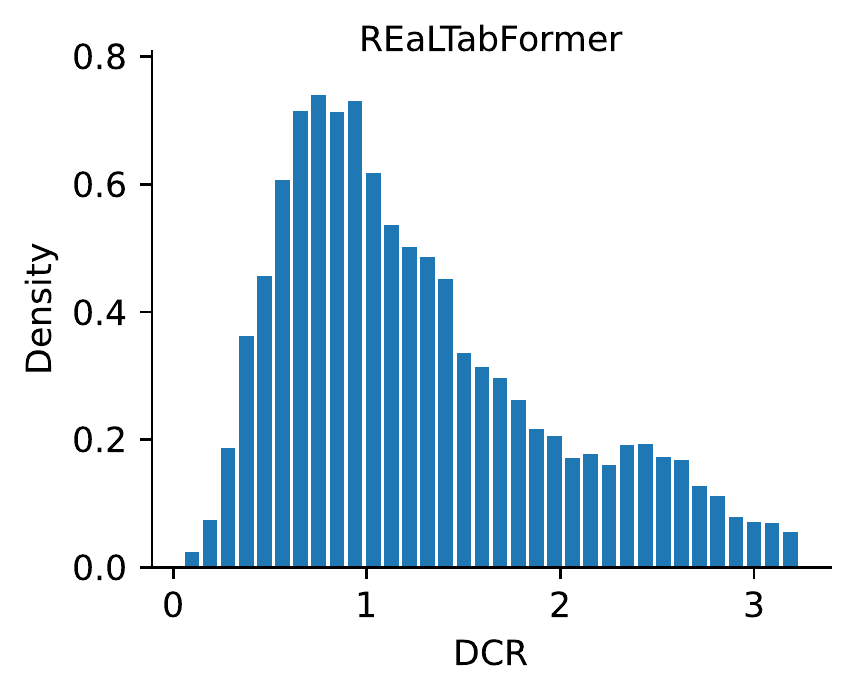}
    \caption{Distance to closest record (DCR) distributions of the different models for the Buddy data.}
    \label{fig:dcr-buddy}
\end{figure}

\begin{figure}[H]
    \centering
    \includegraphics[width=\imgwidth\textwidth]{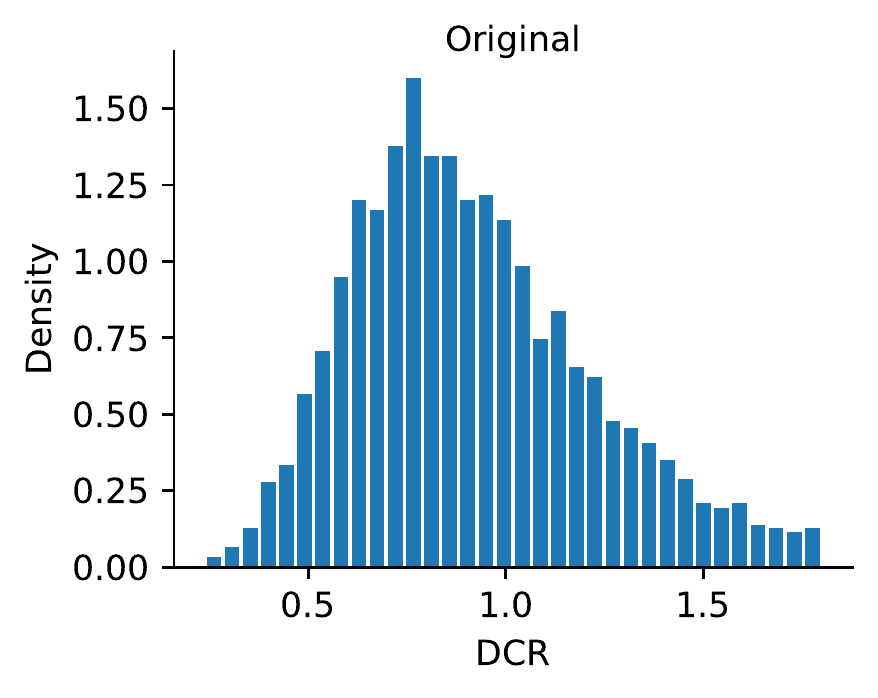}
    \includegraphics[width=\imgwidth\textwidth]{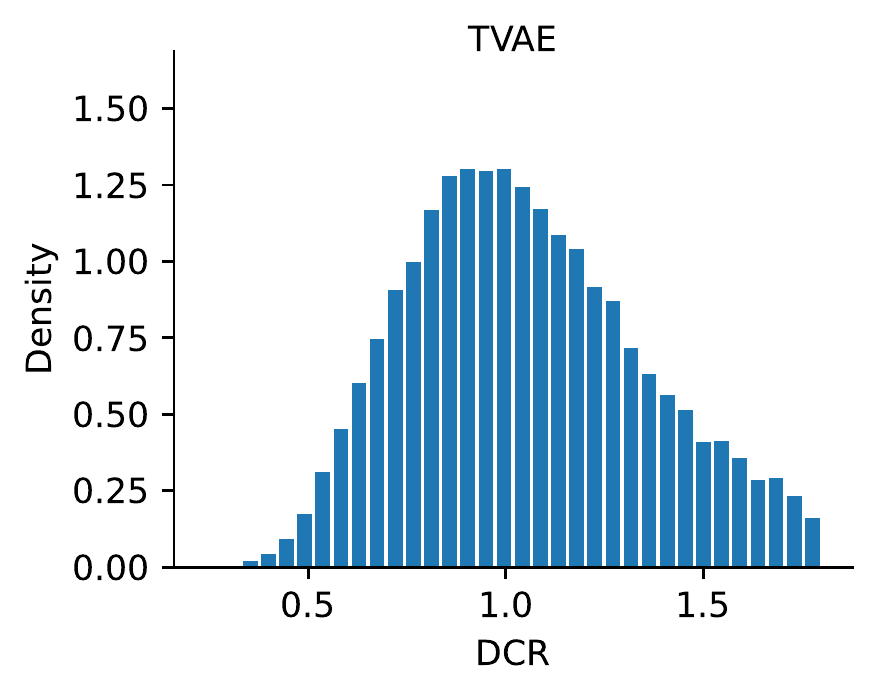}
    \includegraphics[width=\imgwidth\textwidth]{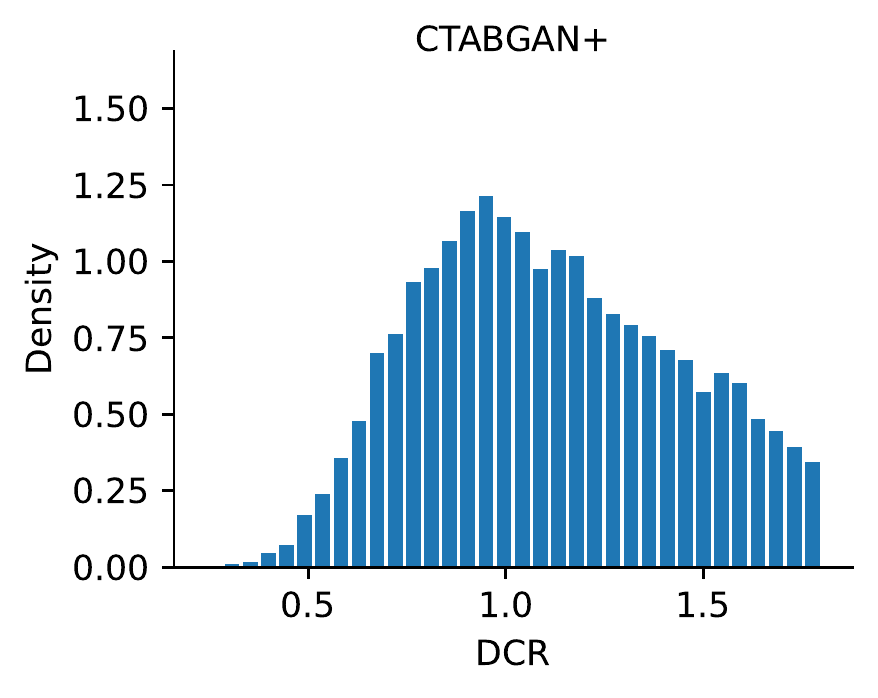}
    \includegraphics[width=\imgwidth\textwidth]{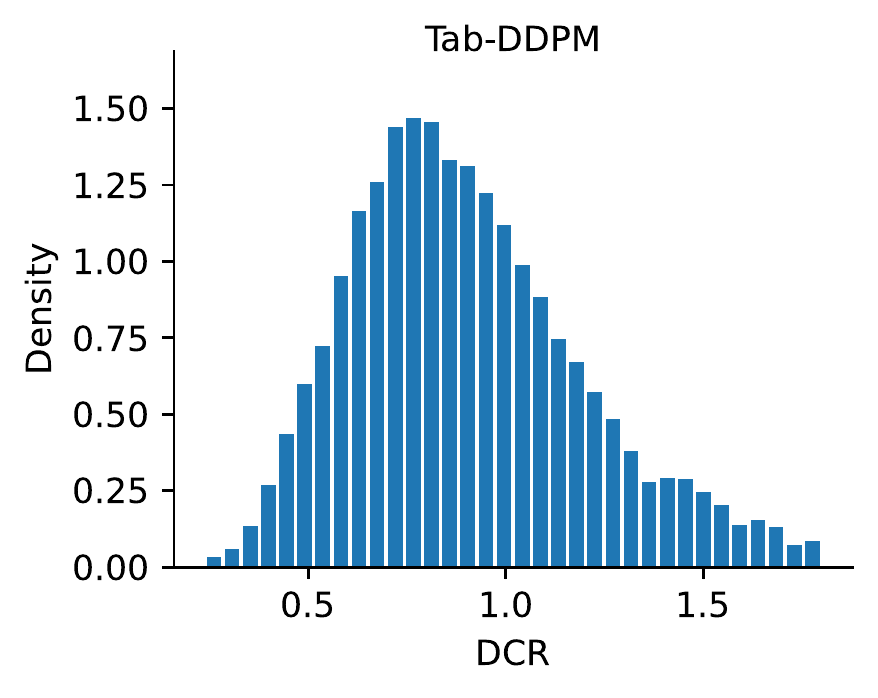}
    \includegraphics[width=\imgwidth\textwidth]{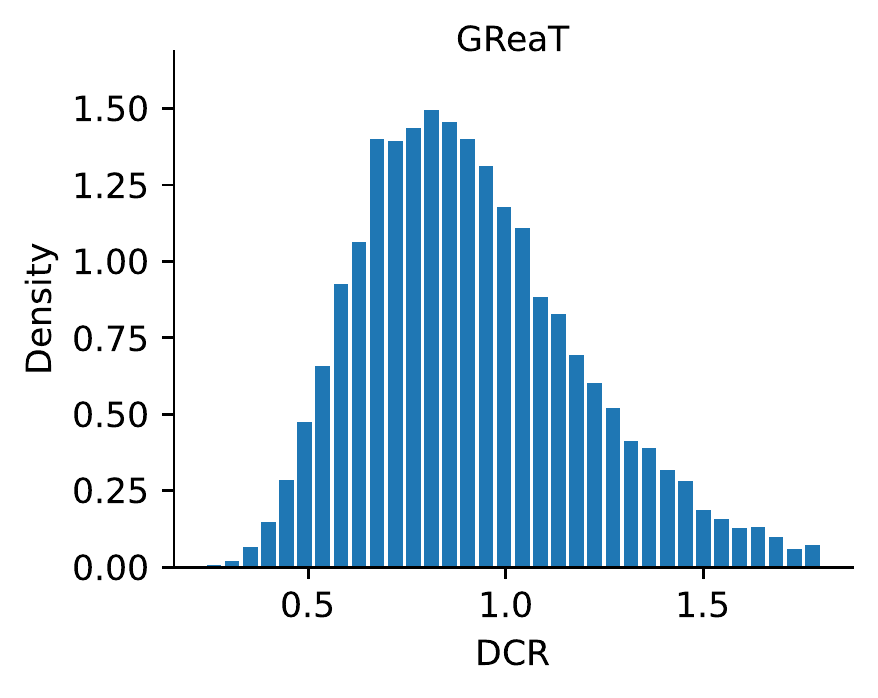}
    \includegraphics[width=\imgwidth\textwidth]{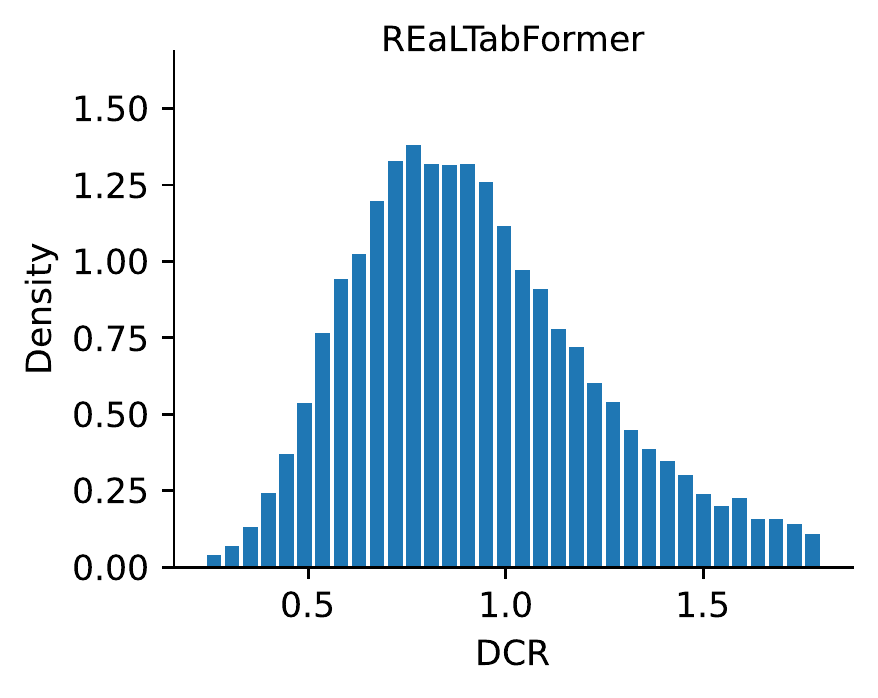}
    \caption{Distance to closest record (DCR) distributions of the different models for the California housing data.}
    \label{fig:dcr-california}
\end{figure}

\begin{figure}[H]
    \centering
    \includegraphics[width=\imgwidth\textwidth]{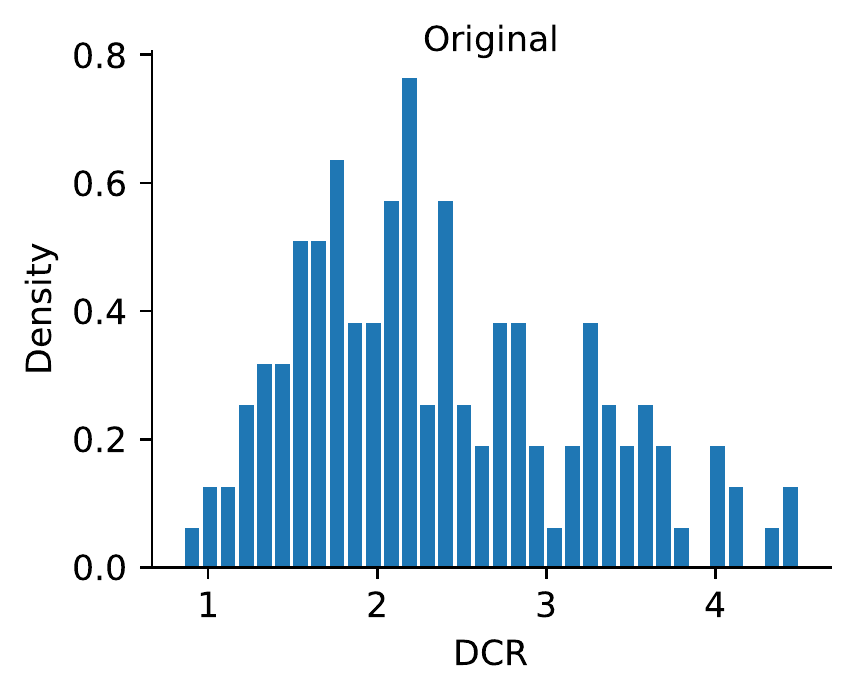}
    \includegraphics[width=\imgwidth\textwidth]{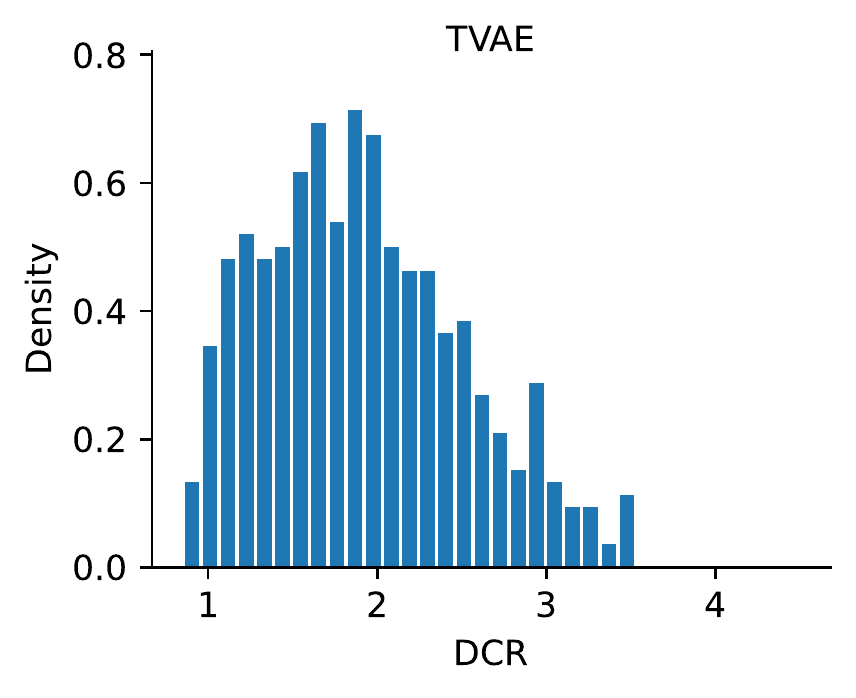}
    \includegraphics[width=\imgwidth\textwidth]{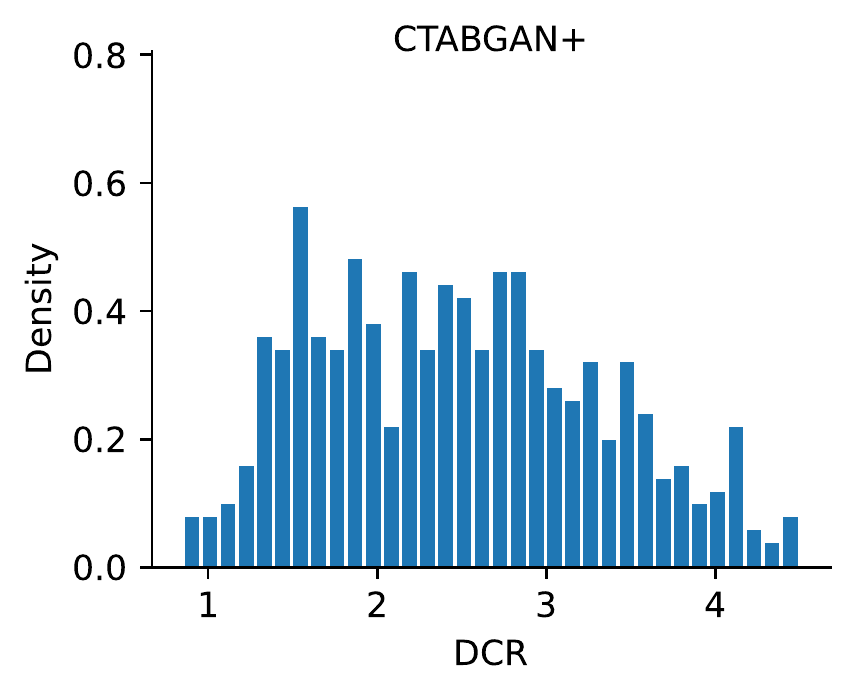}
    \includegraphics[width=\imgwidth\textwidth]{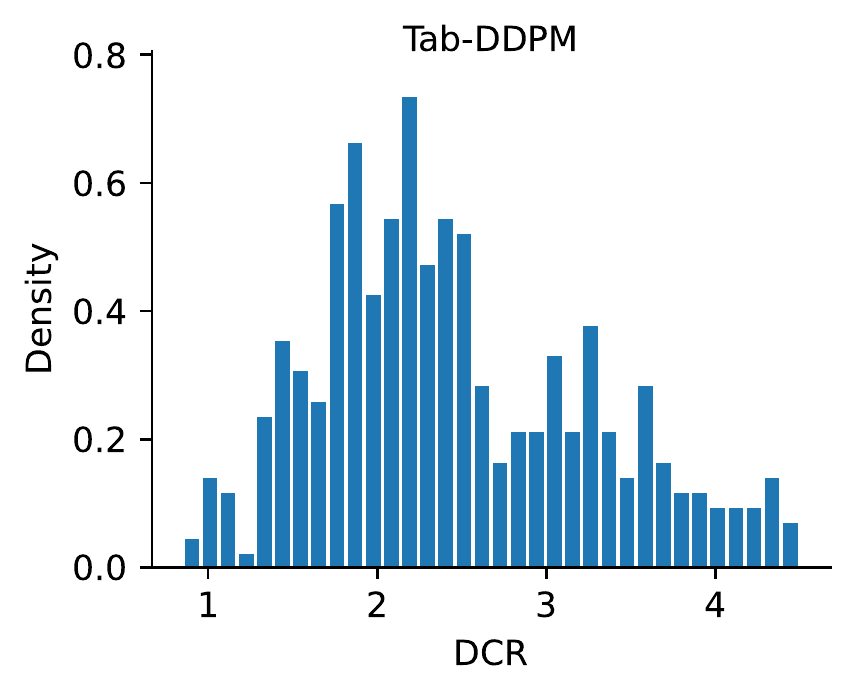}
    \includegraphics[width=\imgwidth\textwidth]{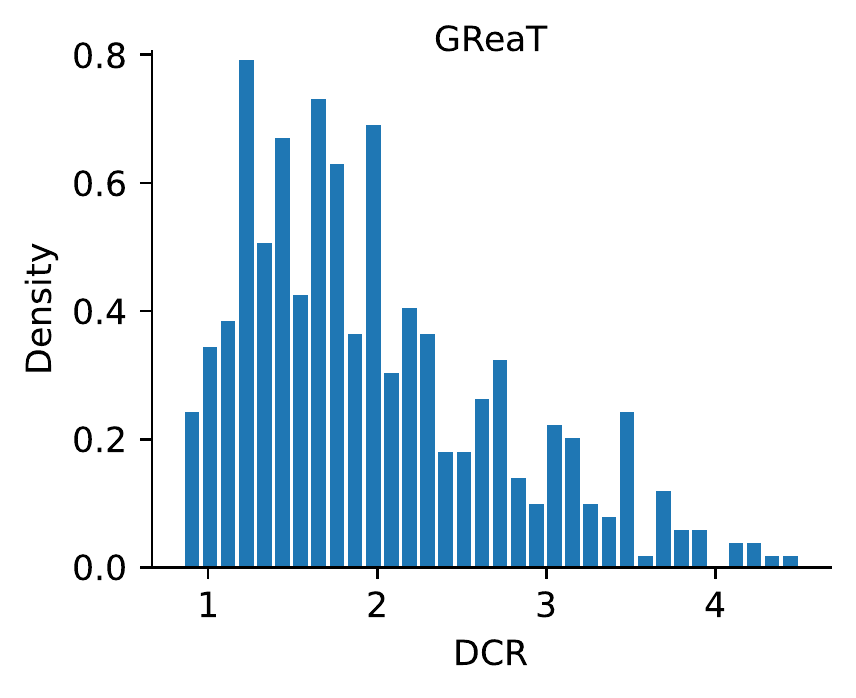}
    \includegraphics[width=\imgwidth\textwidth]{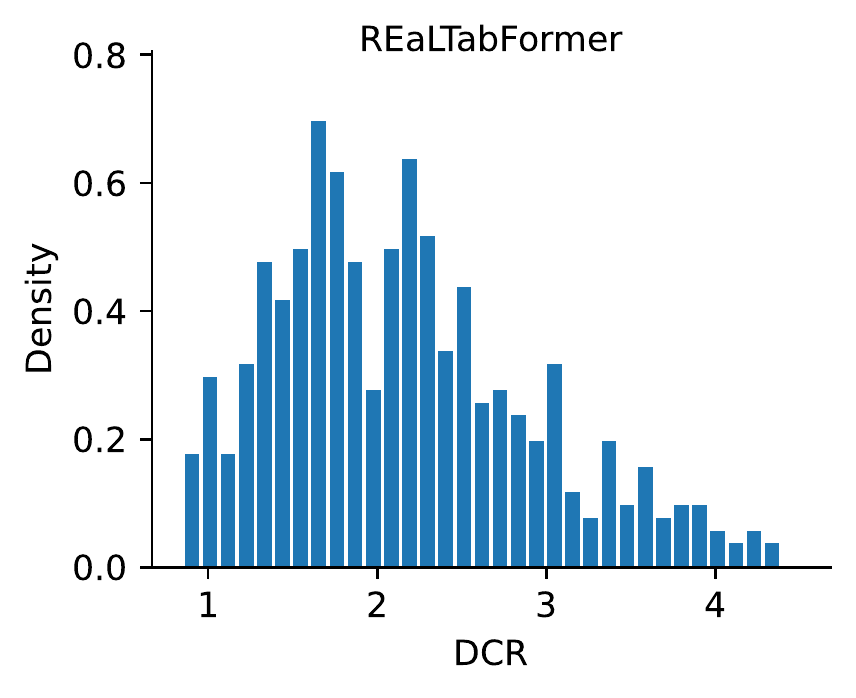}
    \caption{Distance to closest record (DCR) distributions of the different models for the Diabetes data.}
    \label{fig:dcr-diabetes}
\end{figure}


\end{document}

%% file: tables/results_ml_efficiency.tex
\newcommand{\wstd}[2]{#1\small{$\scriptscriptstyle \pm \scriptstyle #2$}}
\newcommand{\bwstd}[2]{\textbf{#1}\small{$\scriptscriptstyle \pm \scriptstyle \mathbf{#2}$}}
\newcommand{\uwstd}[2]{\underline{#1\small{$\scriptscriptstyle \pm \scriptstyle #2$}}}

\begin{table*}[ht!]
\caption{Machine learning efficacy (MLE) and discriminator measure (DM) evaluated on the synthetic data generated by the models (columns) trained on six real-world datasets (rows): Abalone (AB), Adult income (AD), Buddy (BU), California housing (CA), Diabetes (DI), and Facebook Comments (FB). The MLE is measured by the $R^2$ for regression, while macro average $F_1$ is used for classification tasks; higher scores are better. A discriminator measure closer to 50\% is better. Best scores are highlighted for the MLE measure, considering standard deviation. No reported results for GReaT on the FB dataset due to impractical training time.}

\label{tab:mle-dm-results}
\vskip 0.1in
\begin{center}
\begin{small}
    \begin{tabular}{llcccccc}
        \toprule
         & & Original & TVAE & CTABGAN+ & Tab-DDPM & GReaT & \textbf{REaLTabFormer} \\
         \midrule
        
        \multirow{2}{*}{
        AB ($R^2$)} & 
        
        MLE ($\uparrow$) &
        \wstd{0.5562}{0.004} & 
        \wstd{0.3943}{0.012} & 
        \wstd{0.4697}{0.014} & 
        \textbf{ \wstd{0.5248}{0.011} } & 
        \wstd{0.3530}{0.031} & 
        \textbf{ \wstd{0.5035}{0.011} } \\ & 
        
        
        
        
        DM \ \ ($\downarrow$) &
        - &  
        \wstd{82.96}{2.42} & 
        \wstd{75.64}{1.20} & 
        \wstd{59.88}{2.22} & 
        \wstd{70.46}{0.92} & 
        \wstd{63.08}{1.18} \\ 
        
        \midrule
        
        \multirow{2}{*}{
        AD ($F1$)} &
        
        MLE &
        \wstd{0.8155}{0.001} & 
        \wstd{0.7695}{0.004} & 
        \wstd{0.7778}{0.003} & 
        \wstd{0.7922}{0.002} & 
        \wstd{0.7997}{0.002} & 
        \textbf{\wstd{0.8113}{0.002}}\\ & 
        
        DM &
        - &  
        \wstd{95.48}{1.34} & 
        \wstd{61.17}{0.40} & 
        \wstd{53.73}{0.22} & 
        \wstd{68.04}{0.26} & 
        \wstd{55.78}{0.20} \\ 
        
        \midrule
        
        \multirow{2}{*}{
        BU ($F1$)} &
        
        MLE &
        \wstd{0.9303}{0.002} & 
        \textbf{ \wstd{0.9233}{0.002} } & 
        \textbf{ \wstd{0.9267}{0.002} } & 
        \wstd{0.9057}{0.003} & 
        \textbf{ \wstd{0.9279}{0.003} } & 
        \textbf{ \wstd{0.9278}{0.003} } \\ & 
        
        DM &
        - &  
        \wstd{66.56}{0.56} & 
        \wstd{58.33}{0.49} & 
        \wstd{54.43}{0.47} & 
        \wstd{62.18}{0.45} & 
        \wstd{55.86}{0.47} \\ 
        
        \midrule
        
        \multirow{2}{*}{
        CA ($R^2$)} &
        
        MLE &
        \wstd{0.8568}{0.001} & 
        \wstd{0.7373}{0.004} & 
        \wstd{0.5231}{0.006} & 
        \textbf{ \wstd{0.8252}{0.002} } & 
        \wstd{0.7189}{0.004} & 
        \wstd{0.8076}{0.003} \\ & 
        
        DM &
        - &  
        
        \wstd{62.06}{0.60} & 
        \wstd{90.14}{1.03} & 
        \wstd{54.30}{0.89} & 
        \wstd{66.78}{0.47} & 
        \wstd{57.29}{0.56} \\ 
        
        \midrule
        
        \multirow{2}{*}{
        DI ($F1$)} &
        
        MLE &
        \wstd{0.7759}{0.014} & 
        \wstd{0.7395}{0.035} & 
        \wstd{0.7339}{0.024} & 
        \wstd{0.7448}{0.031} & 
        \wstd{0.7419}{0.03} & 
        \wstd{0.7315}{0.027} \\ & 
        
        DM &
        - &  
        
        \wstd{90.16}{1.31} & 
        \wstd{70.94}{1.99} & 
        \wstd{69.00}{1.56} & 
        \wstd{74.88}{1.79} & 
        \wstd{75.56}{2.84} \\ 
        
        \midrule
        
        \multirow{2}{*}{
        FB ($R^2$)} &
        
        MLE &
        \wstd{0.8371}{0.001} & 
        \wstd{0.6374}{0.007} & 
        \wstd{0.4722}{0.053} & 
        \wstd{0.6850}{0.006} & 
        - & 
        \textbf{\wstd{0.7702}{0.004} }\\ & 
        
        DM &
        - &  
        \wstd{97.72}{0.80} & 
        \wstd{93.60}{0.28} & 
        \wstd{66.07}{0.23} & 
        - & 
        \wstd{65.46}{0.83} \\ 
        
        
        
        
        
        \bottomrule
        
    \end{tabular}
\end{small}
\end{center}
\vskip -0.1in
\end{table*}

%% file: tables/relational_ld_results.tex
\begin{table}[t]
\caption{Logistic detection (LD) measure using random forest model for the generated parent, child, and merged tables by the Hierarchical Modeling Algorithm (HMA) from SDV and the REaLTabFormer (RTF) models. Our model consistently beats the SDV model on this metric.}
\label{tab:relational-ld-results}
\vskip 0.1in
\begin{center}
\begin{small}
\begin{sc}
\begin{tabular}{lccc}
\toprule
Dataset & Table & SDV & RTF \\
\midrule
\multirow{3}{*}{ Rossmann } &

Parent &
\wstd{31.77}{3.41} & 
\textbf{ \wstd{81.04}{4.54} } \\ & 

Child &
\wstd{6.53}{0.39} & 
\textbf{ \wstd{52.08}{0.89} } \\ & 

Merged &
\wstd{2.80}{0.25} & 
\textbf{ \wstd{28.33}{2.31} } \\ 

\midrule

\multirow{3}{*}{ Airbnb } &

Parent &
\wstd{7.37}{0.72} & 
\textbf{ \wstd{89.65}{1.92} } \\ & 

Child &
\wstd{0.00}{0.00} & 
\textbf{ \wstd{30.48}{0.79} } \\ & 

Merged &
\wstd{0.00}{0.00} & 
\textbf{ \wstd{21.43}{1.10} } \\ 






\bottomrule
\end{tabular}
\end{sc}
\end{small}
\end{center}
\vskip -0.1in
\end{table}

%% file: tables/data_summary.tex
\begin{table*}[t]
\caption{Summary of the datasets used in the experiments for non-relational tabular data.}
\label{tab:data-summary}
\vskip 0.1in
\begin{center}
\begin{small}
    \begin{tabular}{lcccccccc}
        \toprule
        Abbr & Name & \# Train & \# Validation & \# Test & \# Num & \# Cat & Task type \\
        \midrule
        AB & Abalone & $2672$ & $669$ & $836$ & $7$ & $1$ & Regression \\
        AD & Adult ROC & $26048$ & $6513$ & $16281$ & $6$ & $8$ & Binclass \\
        BU & Buddy & $12053$ & $3014$ & $3767$ & $4$ & $5$ & Multiclass \\
        CA & California Housing & $13209$ & $3303$ & $4128$ & $8$ & $0$ & Regression \\
        DI & Diabetes & $491$ & $123$ & $154$ & $8$ & $0$ & Binclass \\
        FB & Facebook Comments Volume & $157638$ & $19722$ & $19720$ & $50$ & $1$ & Regression  \\
        \bottomrule
    \end{tabular}
\end{small}
\end{center}
\vskip -0.1in
\end{table*}

%% file: tables/relational_ld_logistic_results.tex
\begin{table}[t]
\caption{Logistic detection measure for the generated parent, child, and merged tables by the Hierarchical Modeling Algorithm (HMA) from SDV and the REaLTabFormer (RTF) models. This uses the logistic regression model as the detector.}
\label{tab:relational-ld-logistic-results}
\vskip 0.1in
\begin{center}
\begin{small}
\begin{sc}
\begin{tabular}{lccc}
\toprule
Dataset & Table & SDV & RTF \\
\midrule

\multirow{3}{*}{ Rossmann } &

Parent &
\wstd{78.67}{6.79} & 
\wstd{92.75}{4.28} \\ & 

Child &
\wstd{16.62}{0.86} & 
\wstd{59.00}{2.92} \\ & 

Merged &
\wstd{12.00}{0.73} & 
\wstd{50.69}{2.41} \\ 

\midrule

\multirow{3}{*}{ Airbnb } &

Parent &
\wstd{98.66}{1.34} & 
\wstd{99.68}{0.38} \\ & 

Child &
\wstd{0.00}{0.00} & 
\wstd{26.33}{0.78} \\ & 

Merged &
\wstd{96.71}{2.79} & 
\wstd{98.93}{0.82} \\ 






\bottomrule
\end{tabular}
\end{sc}
\end{small}
\end{center}
\vskip -0.1in
\end{table}